\documentclass[Afour,sageh,times]{sagej}

\usepackage{moreverb,url}
\usepackage{subfigure}
\usepackage{hhline}

\usepackage{graphics} %

\usepackage{amsmath} %
\usepackage{amssymb}  %
\usepackage{array}

\usepackage{pgf}

\usepackage{afterpage}

\usepackage{subeqnarray}

\usepackage{algorithm}
\usepackage{algpseudocode}

\usepackage{placeins}
\usepackage{multirow}

\usepackage{tikz}
\usetikzlibrary{positioning}

\usepackage[colorlinks,bookmarksopen,bookmarksnumbered,citecolor=red,urlcolor=red]{hyperref}

\newcommand{\TwoRows}[2]{\begin{tabular}{@{}c@{}}#1\\#2\end{tabular}}
\newcommand{\TwoRowsR}[2]{\begin{tabular}{@{}r@{}}#1\\#2\end{tabular}}
\newcommand{\TwoRowsBf}[2]{\begin{tabular}{@{}c@{}}\textbf{#1}\\\textbf{#2}\end{tabular}}
\newcommand{\ThreeRows}[3]{\begin{tabular}{@{}r@{}r@{}}#1\\#2\\#3\end{tabular}}
\newcommand{\ThreeRowsC}[3]{\begin{tabular}{@{}c@{}c@{}}#1\\#2\\#3\end{tabular}}

\newcommand{\Figure}[1]{Figure~\ref{#1}}
\newcommand{\Eq}[1]{Eq.~(\ref{#1})}

\newcommand{\MyEps}{{\boldsymbol{\xi}}}

\algnewcommand{\LineComment}[1]{\State \(\triangleright\) #1}
\algnewcommand\algorithmicforeach{\textbf{for each}}
\algdef{S}[FOR]{ForEach}[1]{\algorithmicforeach\ #1\ \algorithmicdo}

\newcommand{\REVIEW}[1]{#1}

\usepackage{rigidnotation}
\NewRigidNotation{Vel}{v} %
\NewRigidNotation{AngVel}{\omega} %
\NewRigidNotation{Trans}{t} %
\SetConciseNotation{\BooleanTrue}
    
\newif\ifJLBCPREPRINT

\JLBCPREPRINTtrue %

\def\journalname{Preprint}
\def\volumeyear{}
\def\volumenumber{}
\def\issuenumber{2024}

\begin{document}

\title{A flexible framework for accurate LiDAR odometry, map manipulation, and localization}

\author{Jose Luis Blanco-Claraco\affilnum{1}}

\affiliation{\affilnum{1}Engineering Department, University of Almería, Spain.
}

\corrauth{José Luis Blanco-Claraco}

\email{jlblanco@ual.es}

\begin{abstract}
Light Detection and Ranging (LiDAR)-based simultaneous localization and mapping (SLAM) 
is a core technology for
autonomous vehicles and robots.
One key contribution of this work to 3D LiDAR SLAM and localization is a fierce defense of view-based maps (pose graphs with time-stamped sensor readings)
as the fundamental representation of maps. 
As will be shown, they allow for the greatest flexibility, enabling the posterior generation of arbitrary metric maps optimized for particular tasks, e.g. obstacle avoidance, real-time localization.
Moreover, this work introduces a new framework in which mapping pipelines can be defined without coding, 
defining the connections of
a network of reusable blocks much like deep-learning networks are designed by connecting layers of standardized elements.
We also introduce tightly-coupled estimation of linear and angular velocity vectors within the Iterative Closest Point (ICP)-like optimizer,
leading to superior robustness against aggressive motion profiles
without the need for an IMU.
Extensive experimental validation reveals that the proposal compares well to, or improves, former state-of-the-art (SOTA) LiDAR odometry systems,
while also successfully mapping some hard sequences where others diverge.
A proposed self-adaptive configuration has been used, without parameter changes, 
for all 3D LiDAR datasets with sensors between 16 and 128 rings, 
and has been extensively tested on 83 sequences over more than 250~km of automotive, hand-held, airborne, and quadruped LiDAR datasets, both indoors and outdoors.
The system flexibility is demonstrated with
additional configurations for 2D LiDARs and
for building 3D NDT-like maps.
The framework is open-sourced online: \url{https://github.com/MOLAorg/mola}.
\end{abstract}


\makeatletter
\let\@oldmaketitle\@maketitle%
\renewcommand{\@maketitle}{\@oldmaketitle%
\begin{center}
\includegraphics[width=0.8\linewidth]{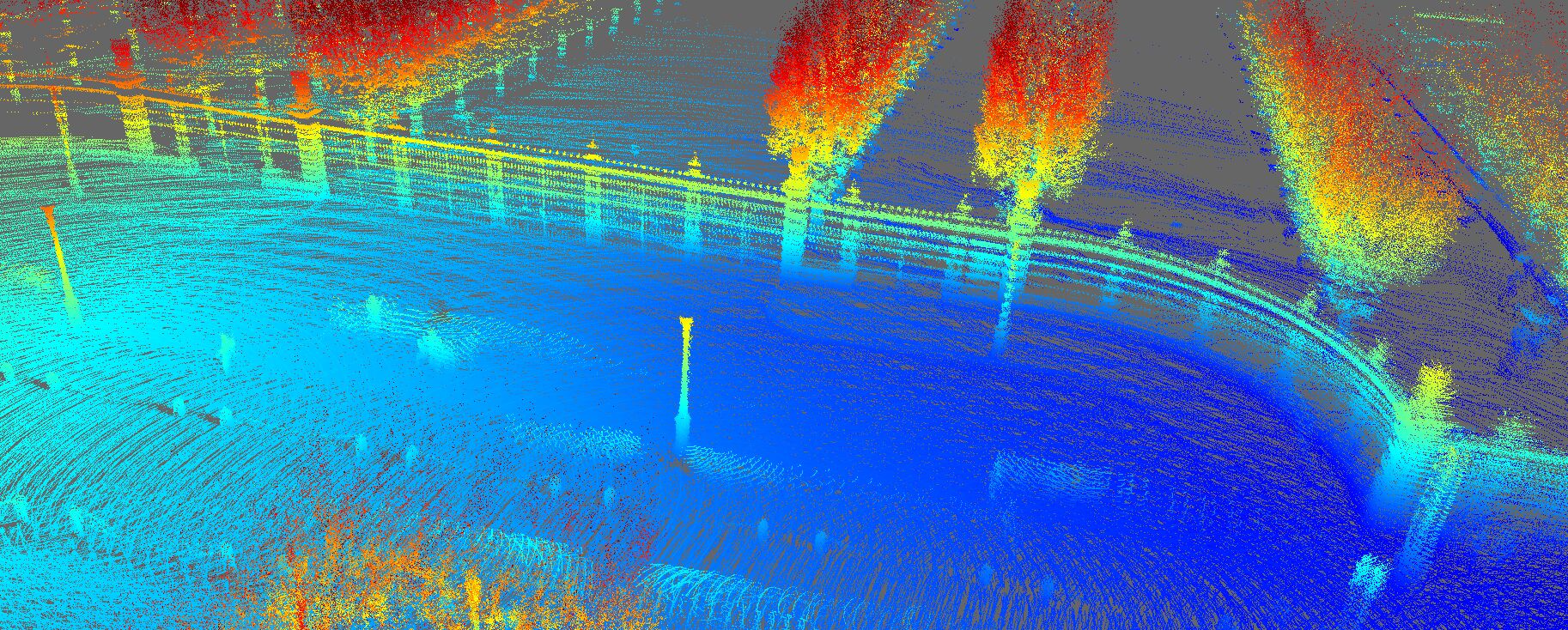}
\end{center}
\textbf{Figure 1.} Example point cloud built from the presented LiDAR odometry system: 
a view of \emph{Esplanade Gaston Monnerville} (Paris)
from the Luxembourg Garden automotive dataset in \cite{dellenbach2022ct}.
This is part of a multi-layer metric map, illustrated in \Figure{fig:sm2mm.example.paris}.
\addtocounter{figure}{+1}

  \bigskip}%
\makeatother

\maketitle

\clearpage

\tableofcontents

\newpage

\section{Introduction}

3D LiDAR is among the most widely-used sensors
in contemporary mobile robots, autonomous
vehicles, and Unmanned Aerial Vehicles (UAV) \citep{elhousni2020survey,lee2023lidar}.
While vision and RADAR are also relevant technologies
and may become essential for many tasks,
LiDAR will likely remain a core sensor for 
years to come \citep{roriz_automotive_2022} 
due to a number of intrinsic 
advantages: direct and accurate 3D sensing
with wide field of view,
robustness against varying sunlight conditions, 
reduced computation demand to obtain 
dense point clouds, etc.

Among all the potential applications of LiDAR sensing, 
this work focuses on the problems of pose tracking
and mapping.
We call \emph{localization} to pose tracking on a prebuilt environment model that remains unmodified.
Positioning in unknown environments becomes
the well-known Simultaneous Localization and Mapping (\emph{SLAM})
problem \citep{cadena2016past}.
When only the most recently-observed parts of the environment
are kept, the term \emph{odometry} is used instead,
hence we will leave the category SLAM for those approaches that are aware of loop closures \citep{grisetti2010tutorial}.
That is, odometry methods provide local accurate estimation of relative motion in the short term, while SLAM must also ensure 
global consistency of both, the trajectory and the world model.

Therefore, when LiDAR is the unique sensor in an odometry system
it is called LiDAR odometry (LO). 
The main component presented in this work falls into that category. 
LO can be used alone or as part of a larger system, 
leading to full metric SLAM, e.g. Cartographer in \citet{hess2016real},
\REVIEW{or multi-robot SLAM systems}, e.g. multi S-Graphs in \citet{fernandez2024multi}.
In the literature, we can find how additional sensors have been 
successfully integrated with LO, leading to 
LiDAR Inertial odometry (LIO) \citep{shan2020lio,Xu2022TRO},
Visual Inertial Odometry (VIO) \citep{leutenegger2015keyframe,forster2016manifold}, 
or Visual Lidar-Inertial Odometry (VLIO) \citep{shan2021lvi}.
However, we will show that, for most conditions found in practice, 
our system is able to achieve good performance and accuracy relying on LiDAR alone, 
notwithstanding that even better results might arise from
multimodal sensor fusion.

The present work defends that LO and SLAM systems 
can be built in a way that maximizes the reusability
of their components, making them more flexible and easier
to modify by end users, 
while also allowing resulting world models (``maps'')
to be built minimizing the loss of information 
with respect to raw sensor data.
Two core ideas arise from these goals:
(i) odometry and SLAM pipelines, together with their associated data structures, should
be refactored into their smallest reusable components, 
allowing building whole systems from scratch by wiring
them together and setting their parameters much like
Deep Neural Networks (DNN) are designed by stacking
prebuilt elements; 
(ii) disregarding the specific map type or features
used by a particular LO or SLAM system, the most 
versatile map data structure is a \emph{view-based map} 
(dubbed ``simple maps'' in our framework),
where a sparse set of key-frames are retained together
with their fundamental information, e.g. raw sensor observations,
kinematic state of the vehicle, etc.
The first usage of the term view-based SLAM
can be found in the 2000s \citep{eustice2005visually,konolige2010view}, although the idea
can be traced back to the ground-breaking work 
\cite{lu1997globally} which served as the seed 
for an extremely successful approach to mapping 
dubbed Graph-SLAM \citep{grisetti2010tutorial}.
A more recent application of this idea of keeping the raw sensor information
in graph-based SLAM can be found within
Google's Cartographer \citep{hess2016real} open-source implementation (refer to ``assets writer''),
although not mentioned in the original publication.

Therefore, building upon the two ideas above, 
\REVIEW{this work introduces an open-source software ecosystem}
to the community.
Note that none of the presented algorithms are based on learning methods, although the modular nature of the framework allows
their integration as future works.

To sum up, the key contributions of this work are:
\begin{itemize}
\item A software ecosystem providing reusable components for building point cloud processing pipelines without coding, including generation of voxel maps, occupancy 2D or 3D grid maps, feature detection, etc.
\item A reference pipeline for LO with 3D LiDARs, well-tested against more than 80 public dataset sequences.
\item A method to easily integrate vehicle velocity as an unknown within ICP loops, allowing \REVIEW{better point} cloud undistortion, subject to a constant velocity assumption during each sensor sweep.
\item A simple method to let ICP take into account prior uncertainty, including that from a constant velocity motion model or other sources of odometry\REVIEW{, e.g. wheel encoders or an IMU}.
\item A loop-closure algorithm to generate globally consistent maps from the LO outcome, optionally including georeferencing.
\item A uniform API (Application Programming Interface) to access a large variety of robotics datasets, simplifying and easing benchmarking SLAM methods.
\end{itemize}

The rest of the article is organized as follows. 
Section~\ref{sect:related} reviews previous works closely related to the presented contributions.
Next, the proposed architecture and their individual parts are introduced in detail in 
Section~\ref{sect:proposed}.
Experimental validation is provided in three parts: 
Section~\ref{sect:results.quant} presents extensive quantitative performance metrics
of the proposed SLAM system, section~\ref{sect:results.qualitative} illustrates
in a qualitative way some use cases of our system, including situations and datasets where other
SOTA systems diverge, and section~\ref{sect:results.localiz} shows localization results
with previously-built maps.
\REVIEW{Furthermore, ablation studies are presented in section~\ref{sect:ablation}
and we} wrap up with conclusions in section~\ref{sect:conclusions}.

\section{Related work}
\label{sect:related}

This section analyzes many of the past research directly related to
the main topic of this paper: LiDAR odometry or SLAM.
LiDAR works have been divided into odometry, loop-closure, and 
localization, with an additional brief subsection on graph SLAM.
We will put a focus on those ideas that have been recurrently proposed
in the literature due to their good results, especially those in which
our system is based on.
An extensive review of these fields is out of the scope of the present work; 
interested readers may refer to existing surveys such as
\cite{cadena2016past}, \cite{bresson2017simultaneous},
or \cite{xu2022review}; or to \cite{durrant2006slam1} and \cite{bailey2006slam} for a 
summary of foundational ideas on which modern SLAM methods were built upon.

\subsection{LiDAR odometry (LO)}
\label{sect:related.lo}

The goal of a LO system is taking successive LiDAR scans and estimating the vehicle pose changes in between them.
At the core of most proposed systems there exist modified versions of the Iterative Closest Point (ICP)
algorithm \citep{besl1992method}, which solves that nonlinear problem in an iterative fashion,
finding successive approximations for the sough relative pose between two input point clouds,
looking for potential point-to-point pairings at each iteration. 
That process is often called scan matching or \REVIEW{point} cloud registration.

We can extract four key findings from past related works. 
First of all, existing approaches can be classified according to whether they try to match the latest observation against 
other past individual observations (the scan-to-scan approach), or against an incrementally-built local map (the scan-to-model or scan-to-map approach).
It has been experimentally found that scan-to-scan works extremely well for 2D LiDARs,
whereas the uneven distribution of measured points in 3D LiDARs favors the scan-to-model approach, 
as demonstrated with recent successful works 
like Cartographer \citep{hess2016real}, 
IMLS-SLAM \citep{deschaud2018imls}, 
or KISS-ICP \citep{vizzo2023kiss}.
Following this line, 
the system proposed in this work hence also adheres to the scan-to-model design (see Section~\ref{sect:lidar.odometry}).

Secondly, it is well known that trying to use ICP to register the raw points from 
a 3D LiDAR leads to extremely poor results. This can be explained by the uneven distribution of sampled points in the
environment, with a higher density of near points that then dominate the cost function of the ICP optimization stage.
This fact, together with the common use of LiDARs with a circular scanning pattern, leads to degenerate ICP solutions
where nearby rings are aligned with nearby rings, disregarding the contribution from points sampled from distant parts of the environment.
To overcome this problem, downsampling has been extensively used in the literature.
One of the first successful LO systems\REVIEW{, LOAM, introduced in} \cite{zhang2014loam}, splits the input clouds into two distinct subsets (corners and planar patches),
performing a subsampling to keep only a maximum of 2 edge or 4 planar points per planar scan.
\REVIEW{This seminal work led to many variations, 
such as Lego-LOAM in \cite{Shan2018IROS} or PaGO-SLAM in \cite{Seo2022UR} which exploited an explicit ground plane segmentation.
One of the most prominent derived works is Fast-LIO2 \citep{Xu2022TRO}, which 
leverages tightly-coupled IMU-LiDAR integration with efficient data structures for 
fast and accurate mapping without the need to extract features}.
Other recent works \citep{deschaud2018imls,vizzo2023kiss} employ voxel-based sampling,
where raw points are first classified into 3D voxel data structures and then
a single representative point is extracted from each occupied voxel following a given 
heuristic criterion (e.g. the average, the first point, etc.).
LOCUS 2.0, introduced in \cite{reinke2022iros}, proposes an adaptive multi-level
sampling of point clouds to try to maintain a constant number of points per scan.
Following the same goal, another recent work dubbed SIMPLE \citep{bhandari2024minimal} 
also performs spatial subsampling, but implemented via KD-trees instead of voxels. 
Efficient sampling of point clouds for registration is a research field on its own.
The \texttt{libpointmatcher} project, started with \cite{Pomerleau12comp}, 
implements different strategies, such as those 
proposed in \cite{rusinkiewicz2001efficient} or in \cite{labussiere2020geometry}.
Latest works tend to use simpler sampling schemes, but we have experimentally verified
that even tiny implementation differences in the way points are sampled (e.g. using different hash functions in unordered voxel containers for successive downsampling stages, which determines which is the first point to fall into a given voxel)
have a relevant impact in the accuracy of the obtained trajectories. 
Therefore, our system offers simple voxel-based sampling schemes as the default choice,
but it also offers other possibilities for the community to experiment and benchmark them,
while also being easily extensible with new sampling algorithms (see Section~\ref{sect:pipelines}), 
including those based on deep learning, e.g. following \cite{lang2019pointpillars}.

Thirdly, all modern approaches to LO have replaced the original 
closed-form least-squares formulation in the original ICP paper \citep{besl1992method}
with less efficient but more robust alternatives.
In particular, instead of leaving the cost function to uniquely depend on 
the L2 distances between pointwise pairings, 
two enhancements are nowadays common:
(i) employing robustified least squares, that is, robust kernel or loss functions 
as in \citep{deschaud2018imls,chebrolu2021adaptive,vizzo2023kiss}, 
and (ii) using more geometric constraints apart of point-to-point pairings. 
On this latter point, the most common alternative is using point-to-plane pairings, 
as used in the original LOAM \citep{zhang2014loam} or, \REVIEW{more recently, in Fast-LIO2 \citep{Xu2022TRO}} and in MAD-SLAM \citep{ferrari2024mad}.
LOCUS 2.0 \citep{reinke2022iros} introduces the estimated normals in the cost function, 
while another prominent work in this sense, MULLS \citep{pan2021mulls}, 
includes several
potential geometric pairings simultaneously.
A word is in order about SIMPLE \citep{bhandari2024minimal}, which 
employs a different approach: its cost function avoids the 
determination of point-to-point pairings by just adding Gaussian-like ``rewards'' for each nearby point, 
which can be seen as a one-to-many point pairing policy.
Learning from all this body of past works, our framework provides a generic ICP-like optimization
algorithm at its core, with user-extensible cost functions, predefined cost functions for 
point-to-point, point-to-plane or point-to-line (among others), different robust kernels, 
and the possibility of multiple correspondences (see Section~\ref{sect:icp.pipelines}).

The fourth lesson learned from past works comes 
from KISS-ICP \citep{vizzo2023kiss}, which demonstrated that
dynamically adaptable parameters are beneficial for two reasons: 
(i) they can improve the overall accuracy since parameters adapt automatically as the
environment or the motion profile change over time, 
and (ii) they reduce the need for manually tuning parameters.
From that work we have adopted (and modified) their idea of using an 
adaptive threshold for determining pairings, and extended the idea to many other 
parameters along the entire processing pipeline (see Section~\ref{sect:dyn.vars}).

\subsection{LiDAR localization}

Localization with a predefined, static map and a 3D LiDAR is a topic with less research activity than LO and SLAM, 
despite the fact that one of the fundamental reasons one may want to build a map is to use it for navigation, which 
implies the ability to keep the vehicle localized in the map frame of reference.
As in any other estimation problem, we could split classic approaches (those not based on deep learning)
into those relying on \emph{parametric} or on \emph{non-parametric} distributions.
The latter typically involves using particle filtering \citep{thrun2001robust},
which are an excellent approach for vehicles constrained to trajectories on SE(2),
but whose performance may degrade due to the curse of dimensionality for SE(3) motion.
Still, it was shown in \cite{blanco2019benchmarking} how it is possible to keep a vehicle well-localized
in SE(3) using a \REVIEW{particle filter}, from raw 3D LiDAR data and wheels odometry, much faster than real time. 
This line of work was enhanced by detecting and tracking features (pole-like objects) in 
\cite{schaefer2021long}, still using \REVIEW{particle filtering}.
On the other hand, parametric methods for localization, such as \cite{hess2016real} or 
\cite{lee2024localization_no_hd_map}, typically rely on an ICP-like optimization loop to find the 
best vehicle pose at a given instant.

Since we believe that both approaches have their applicability niches, both solutions are provided
in the present framework (see Section~\ref{sect:results.localiz}).

\subsection{Graph optimization}

The problem of finding the global optimal for a set of
rigid transformations between a sequence of key-frames
is a very well studied one in the robotics community, 
either in its planar SE(2) or spatial SE(3) form.
It has been dubbed Graph-SLAM \citep{grisetti2010tutorial,Grisetti2012IROS},
pose-graph optimization \citep{tian2021distributed},
or synchronization if mostly interested in the rotational parts 
\citep{wang2013exact,carlone2015initialization}.

The goal of all these methods is to find the poses between
the key-frames that minimize the mismatch of (typically sparse) relative pose observations. 
These techniques have an obvious direct application to optimization-based \citep{strasdat2012visual} odometry
frameworks that follow the frame-to-frame paradigm.
Despite the present work follows the frame-to-map model instead,
we mention graph optimization here since it becomes essential when handling loop closures.

The most relevant graph optimization libraries are reviewed next, 
with a focus on those tailored to graph SLAM rather than bundle adjustment.
One of the earliest works in this field is probably 
TORO \citep{grisetti2007tree,grisetti2007efficient,grisetti2009nonlinear}, which implemented efficient gradient descent 
for pose networks of arbitrary topology (multiple loop closures).
Next, G2O was presented in \cite{kummerle2011g2o}, becoming
extremely popular, to the point that it is still used nowadays
in SOTA frameworks.
Three G2O features that explain this success are: (i) its modular
design that allowed for arbitrary graphs to be defined, including
pose-only graphs and those required for visual bundle adjustment
\citep{triggs2000bundle}, 
(ii) the popularization of 
perturbation-based optimization on Lie groups 
(the $\boxplus$ notation)
\citep{kummerle2011g2o,sola2018micro, blanco2021tutorial}, 
and (iii) the support for robust kernels \citep{chebrolu2021adaptive}, essential for outlier rejection.

Afterwards, two other popular frameworks were introduced: 
GTSAM \citep{dellaert2021factor}, and 
CERES \citep{Agarwal_Ceres_Solver_2022}.
Both are frameworks for solving large, non-linear optimization
problems, and both feature robust kernels and some degree of automated calculation of cost function derivatives.
In our loop closure subsystem (described in Section~\ref{sect:lc}) we chose GTSAM as the graph optimization back-end.

\begin{figure*}
\centering
\begin{tabular}{c}
\subfigure[]{\includegraphics[width=1.0\columnwidth]{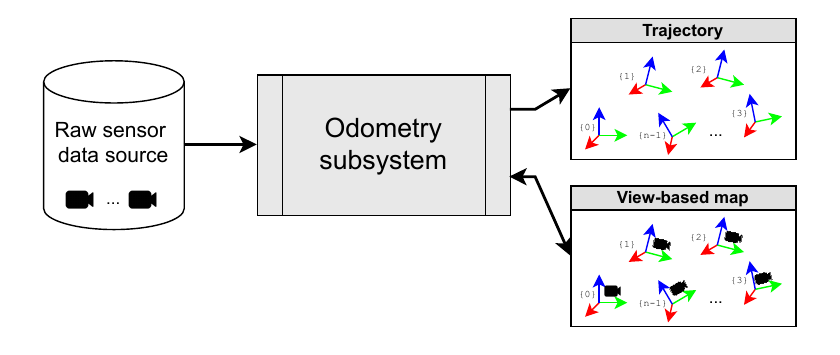}} 
\\
\subfigure[]{\includegraphics[width=1.0\columnwidth]{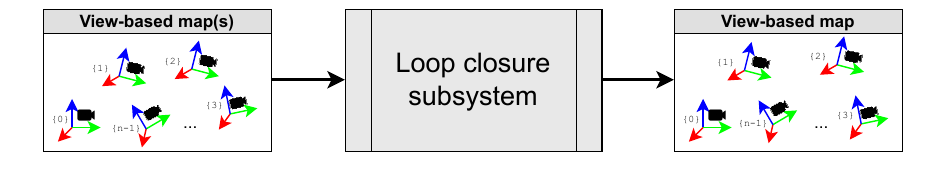}} 
\\
\subfigure[]{\includegraphics[width=1.0\columnwidth]{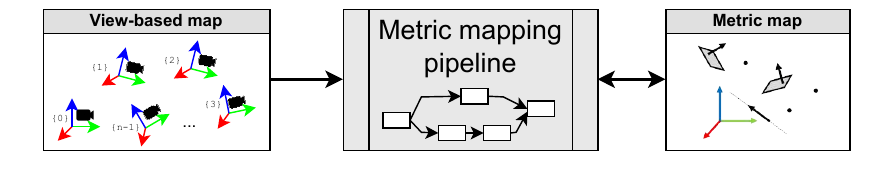}} 
\end{tabular}
\begin{tabular}{c}
\subfigure[]{\includegraphics[width=0.9\columnwidth]{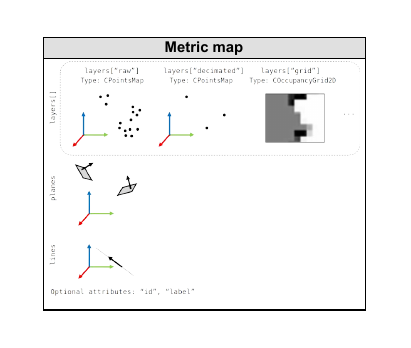}} 
\end{tabular}
\caption{Main top-level data structures and procedural components of the presented framework: 
(a) odometry and (b) loop-closure subsystems, 
(c) metric map generation or update pipelines, 
and (d) a scheme of the contents of
a metric map. Refer to discussion in Section~\ref{sect:overview}. 
The contents of each subsystem are detailed in subsequent figures.
}
\label{fig:overview}
\end{figure*}

\subsection{LiDAR-based loop closure and SLAM}

In the context of SLAM, loop closure can be defined as 
the detection of when the robot has revisited a placed that
is already known, often after traversing a path that resembles
a closed loop. \emph{Topological} loop closure just detects 
such events, while \emph{metrical} loop closure methods also
identify the relative pose between the old and the new observations.

What does actually mean a loop closure in practice, and the way the 
problem is approached, are both strongly influenced by the employed sensors. For visual SLAM with typical pin-hole cameras, moving a camera around a small room would lead to loop closures, while for a 3D LiDAR with 360$^\circ$ of horizontal field of view one might never consider a loop closure situation if moving in a large warehouse if there are no severe occlusions. 
This variety of situations makes this research field extremely active
for each sensor niche
\citep{huang2019visual,wang2019brief,arshad2021role,tsintotas2022revisiting}.

Regarding the topic of this work, LiDAR sensors, 
a recent work by \cite{gskim-2021-tro} introduced a new rotation-invariant descriptor 
for performing topological and (partial) metric localization.
Similar ideas are proposed in related works 
like LiDAR-IRIS \citep{wang2020lidar} which uses a polar representation of 3D point clouds, or in \cite{gupta2024effectively}.

The system presented in this work, however, does not rely on
topological loop closure detection (although it would be integrated as future works): instead, we rely on pure ICP-based alignment with
a special ICP pipeline different than the one used for regular LO,
fed by initial guesses from consumer-grade GNSS, if available.
This simple approach has demonstrated good enough for all the datasets
explored in this work.

\section{Proposed architecture}
\label{sect:proposed}

This section describes in detail the different parts of the proposed system.
The next subsection starts providing a top-level overview of the
key elements, and subsequently they are explained individually
from a scientific and algorithmic point of view.

\subsection{Overview}
\label{sect:overview}

Pursuing the goals defined in the Introduction,
we define the basic blocks illustrated in Figure~\ref{fig:overview}.
Among them, three essential types of procedural blocks or algorithms are defined. 
Firstly, we have the odometry and loop-closure subsystems. 
The idea of splitting SLAM into odometry (``local mapping'')
and loop-closure (``global mapping'') can be traced back
to the visual SLAM community with PTAM \citep{klein2007parallel}
and has been adopted by most subsequent successful solutions for 
visual SLAM (RTAB-Map \cite{labbe2014online}, ORB-SLAM \cite{campos2021orb})
or LiDAR SLAM (Cartographer \cite{hess2016real}).
These two blocks are discussed in 
Section~\ref{sect:lidar.odometry} and 
Section~\ref{sect:lc}, respectively.

Secondly, we define generic metric mapping pipelines as one of the core 
elements in our framework, since they are used internally as building
blocks for both, odometry and loop-closure, apart of having additional applications.
They can be thought of as arbitrary combinations of algorithm 
as processing nodes in a data-driven network, with data being metric maps.
Their final goal is taking raw sensory data (or a former metric map) as inputs,
and processing them to create a new metric map (or to modify the existing one).
They are introduced in Section~\ref{sect:pipelines}.

The input and output of the algorithms defined in Figure~\ref{fig:overview}
are data-related elements:

\subsubsection*{Raw sensor data source}:
This represents the source of sensor data, 
which can be either a real live device or an offline dataset. 
Real-world applications will normally use the former, 
while benchmarking and development predominantly use the latter
for convenience. Available options in our framework are enumerated in Section~\ref{sect:dataset.sources}

\subsubsection*{Metric maps}:
Most existing methods in the literature use one single metric map
representation (e.g. point clouds in \cite{deschaud2018imls} or \cite{vizzo2023kiss})
while others maintain two (e.g. corner and plane points in LOAM \citep{zhang2017low} and most derived works).
The present work proposes allowing the definition of an arbitrary set of 
metric map representations and data structure implementations.
Some of them will be more useful during real-time mapping, 
others during map postprocessing or loop closing, 
others for obstacle avoidance, etc.
More on metric maps follows up in Section~\ref{sect:metric.maps}

\subsubsection*{View-based maps}:
\label{sect:view.based.maps}
Dubbed ``simple-maps'' in our implementation,
they contain a subset of all incoming raw sensor data, 
synchronized and paired with the estimated pose according to an odometry or SLAM method to form key-frames.
More details are given in Section~\ref{sect:simple.maps}.

\subsubsection*{Trajectories}: A trajectory $\mathbf{t}$ is simply a 
collection of $n$ SE(3) timestamped poses, for time steps $t_i$
with $i=1,...,n$:
\begin{eqnarray}
\mathbf{t} = \left\{ 
  \left( t_1, \Pose{1} \right), ...,  
  \left( t_n, \Pose{n} \right)
\right\}
\end{eqnarray}
\noindent where, 
following the RIGID notation for poses \citep{nadeau2024standard},
$\Pose{i}{w}\equiv\Pose{i}$ stands for the $i$-th 
SE(3) pose of the vehicle local reference frame
with respect to the world ($w$), that is, in global coordinates.

\subsection{Metric maps}
\label{sect:metric.maps}

A metric map in the presented framework comprises a superposition
of multiple underlying individual maps called \emph{layers},
plus optional sets of planes and lines, as 
sketched in Figure~\ref{fig:overview}(d).

The idea is using such compound metric maps as the fundamental
data containers used as both inputs and outputs of metric map
pipelines (see Section~\ref{sect:pipelines}) and as inputs to the
ICP-like optimization pipeline (see Section~\ref{sect:icp.pipelines}).
The existence of different map layers is justified
by two facts:
\begin{itemize}
\item Layer map types are independent, e.g. some layers may be point clouds, others are occupancy voxel maps, each with a different resolution, etc.
\item Layers have diverse semantics, e.g. corners, planes, vehicles, pole-like objects, a horizontal slice of a 3D map useful for detecting walls, etc.
\end{itemize}

Such layer differentiation is useful during mapping, but also in posterior post-processing stages.
Our framework allows the definition of new metric map types by users by means of a virtual
base interface that abstracts all common operations of maps such as clearing, inserting sensor observations,
evaluating the likelihood of a given observation, 
searching for nearest neighbors, etc.
Having such abstract interface does not only allow for the extensibility and generic programming in our framework,
but also enables performing performance benchmarking to take optimal decisions about what map type to use
for each application.
Next we enumerate the metric map types that have been used in the experiments presented in this work:

\begin{enumerate}
\item \REVIEW{Point clouds} with contiguous storage layout in memory for each point component:
This group of types include maps for simple clouds where each point only has $(X,Y,Z)$ components,
with an intensity \REVIEW{($I$)} component $(X,Y,Z,I)$, and 
with LiDAR ring \REVIEW{($R$)} identifier and per-point timestamps \REVIEW{($T$), that is, $(X,Y,Z,I,R,T)$ components}.

\item \REVIEW{Point clouds} with a hashed voxels storage layout: this map type implements
3D voxels indexed by voxel coordinates using the optimal hash function introduced in \cite{teschner2003optimized}, 
with a hard-coded maximum number of points. This is the most common map layer type in most parts of the proposed system,
and it is based on similar successful implementations in past works \citep{deschaud2018imls,vizzo2023kiss}.

\item Occupancy 3D voxel map: this type of map employs the re-implementation 
in \cite{faconti2023} of Volumetric Dynamic B+Trees (VDB), originally introduced by \emph{DreamWorks Animation}
in \cite{museth2013vdb}.
It offers an efficient hierarchical data structure to cover large 3D volumes with high-resolution voxels.
VDB could be used to store any arbitrary data within voxels, but in our framework, we so far only offer
two map types: one to store the occupancy of each voxel, and another one that also includes its color.
This map type has been used to represent both, 2D and 3D occupancy grid maps, especially if they are sparse.

\item \REVIEW{3D-NDT (Normal Distribution Transform) maps, as proposed in \cite{magnusson2007scan},
where a 3D sparse voxel map stores fitted Gaussian distributions for points in each voxel.
In our implementation, voxels with a sufficient number of points whose spatial distribution clearly defines 
a plane store the mean and covariance matrix of such points, whereas voxels
with too few samples or which are not planar enough retain all individual points.
The idea is to enable our map to establish both, point-to-point pairings (for voxels which are not planar)
and point-to-plane pairings (for planar ones). In particular, the NDT representation for a voxel is
used if $\sigma_1/\sigma_2 < \tau$, with $\{\sigma_1,\sigma_2,\sigma_3\}$ the three eigenvalues of the
voxel point distribution sorted in ascending order and $\tau$ a threshold (0.05 in our experiments). As the map is populated with incoming observations, individual voxels may switch between the two representations.}

\item 2D grid maps with a plain contiguous memory layout: The most efficient data structure for 
2D occupancy grids that neither, are very sparse, nor need to grow the map limits often.
These maps have been popular in mobile robotics since \cite{elfes1987sonar}.
\end{enumerate}

\subsection{Metric map generation and update pipelines}
\label{sect:pipelines}

\begin{figure}
\centering
\subfigure[]{\includegraphics[width=0.75\columnwidth]{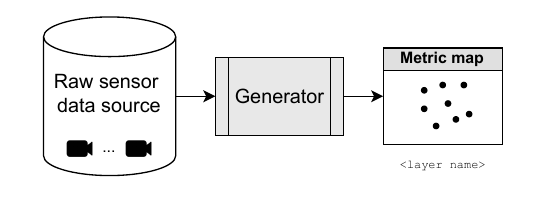}}
\subfigure[]{\includegraphics[width=0.55\columnwidth]{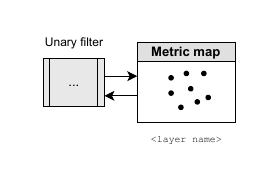}}
\subfigure[]{\includegraphics[width=0.75\columnwidth]{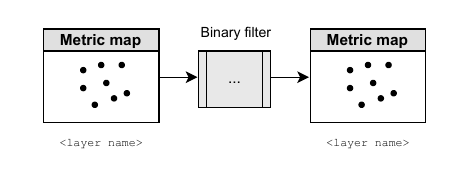}}
\subfigure[]{\includegraphics[width=0.75\columnwidth]{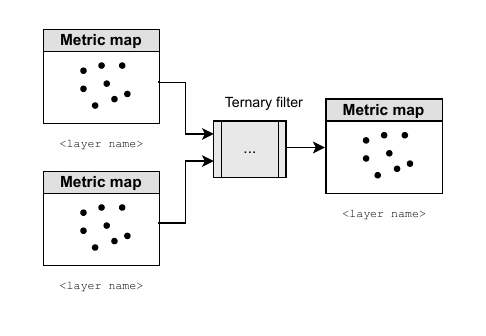}}
\caption{Metric map pipelines represent map processing networks as bipartite graphs with two kinds of nodes: data (map layers) and actions (generators or filters).
Different connection patterns exist: 
(a) generators converting raw observations into map layers, and filters of different arity such as 
(b) unary, (c) binary, or (d) ternary map filters.
See discussion in Section~\ref{sect:pipelines}.}
\label{fig:pipeline.graph.definitions}
\end{figure}

Metric map pipelines are at the core of the flexibility offered by the present framework.
Such pipelines are defined as a bipartite, directed graphs with two kinds of nodes: 
data nodes (metric maps, as defined in Section~\ref{sect:metric.maps})
and action nodes (either filters or generators).
Directed edges may only exist between data nodes and action nodes.
Follow Figure~\ref{fig:pipeline.graph.definitions} as a reference
for the discussion below, keeping in mind that despite map layers
are represented there as point clouds,
arbitrary map types are allowed as long as they are compatible with their action nodes.

\emph{Generators} (see Figure~\ref{fig:pipeline.graph.definitions}(a)) 
are especial actions as they are the entry points for sensor data,
transforming raw range data, images, depth images, etc.
into more elaborated data structures, e.g. point clouds with different
per-point field annotations, or plane patches normals. 
They can also generate more than one output metric map layers, becoming the ideal
location where to run classification tasks, e.g. detecting edges or special objects 
in depth cameras or LiDAR observations.
We provide hooks for implementing custom detection algorithms that need to exploit the 
particular data structure for each possible input sensor, e.g. edge detection algorithms from LiDAR range images.
A generic default generator exists which just creates a map layer of a given type (by default, a point cloud)
and inserts the observation into it by using the abstract interface mentioned in Section~\ref{sect:metric.maps}.
For 2D or 3D LiDARs, this simply creates an unfiltered point cloud with all sensed points in a given scan.

\emph{Filters} are the other kind of action in this framework,
and are broadly defined as any arbitrary algorithm that takes at least one map layer
as input and generates or modifies other (or the same input) layers.
Our work already ships several such filters, while also offering a plug-in system
to add user-defined ones.
Some of the most important filters, 
used in the LiDAR odometry pipeline presented in Section~\ref{sect:icp.pipeline.default}, are briefly introduced next: 

\subsubsection{De-skewing}:
\label{sect:deskewing}
Most common LiDAR sensors today 
are rotating scanners, where a significant time elapses
between the first and the last points in a complete $360^\circ$ scan
(typically between 50~ms and 200~ms). 
De-skewing or un-distortion is a well-known scan data pre-processing
stage where the \emph{estimated} vehicle motion is used to interpolate
the sensor trajectory in SE(3) over time during the scan period 
in order to project all sensed points to 3D space in the most precise way.
Given a set of raw (distorted) points 
${}_s\mathbf{\tilde{R}}=\{{}_s\mathbf{\tilde{r}}_i\}_{i=1}^{N}$
in the sensor (``$s$'') frame of reference\footnote{The 
actual implementation left-multiplies these transformations with $\Pose{s}{b}$, 
the sensor frame ``$s$'' pose in the vehicle body frame ``$b$''
in order to give estimates of the vehicle trajectory and to enable data fusion from several sensors.},
with associated \emph{relative} time-stamps $t_i$
such as for $t=0$ the sensor pose in the map frame is $\Pose{s}=\{\Pos{s}, \Rot{s}\}$
(with $\Pos{s}\in\mathbb{R}^3$ the translation and $\Rot{s}\in SO(3)$ the orientation),
with linear velocity $\Vel{s}{s}$
and angular velocity $\AngVel{s}{s}$ (both in the sensor frame ``$s$''),
simple extrapolation on SE(3) is performed to obtain the 
corrected point cloud $\mathbf{R}=\{\mathbf{r}_i\}_{i=1}^{N}$ in the map frame:

\begin{figure*}
\centering
\includegraphics[width=0.99\textwidth]{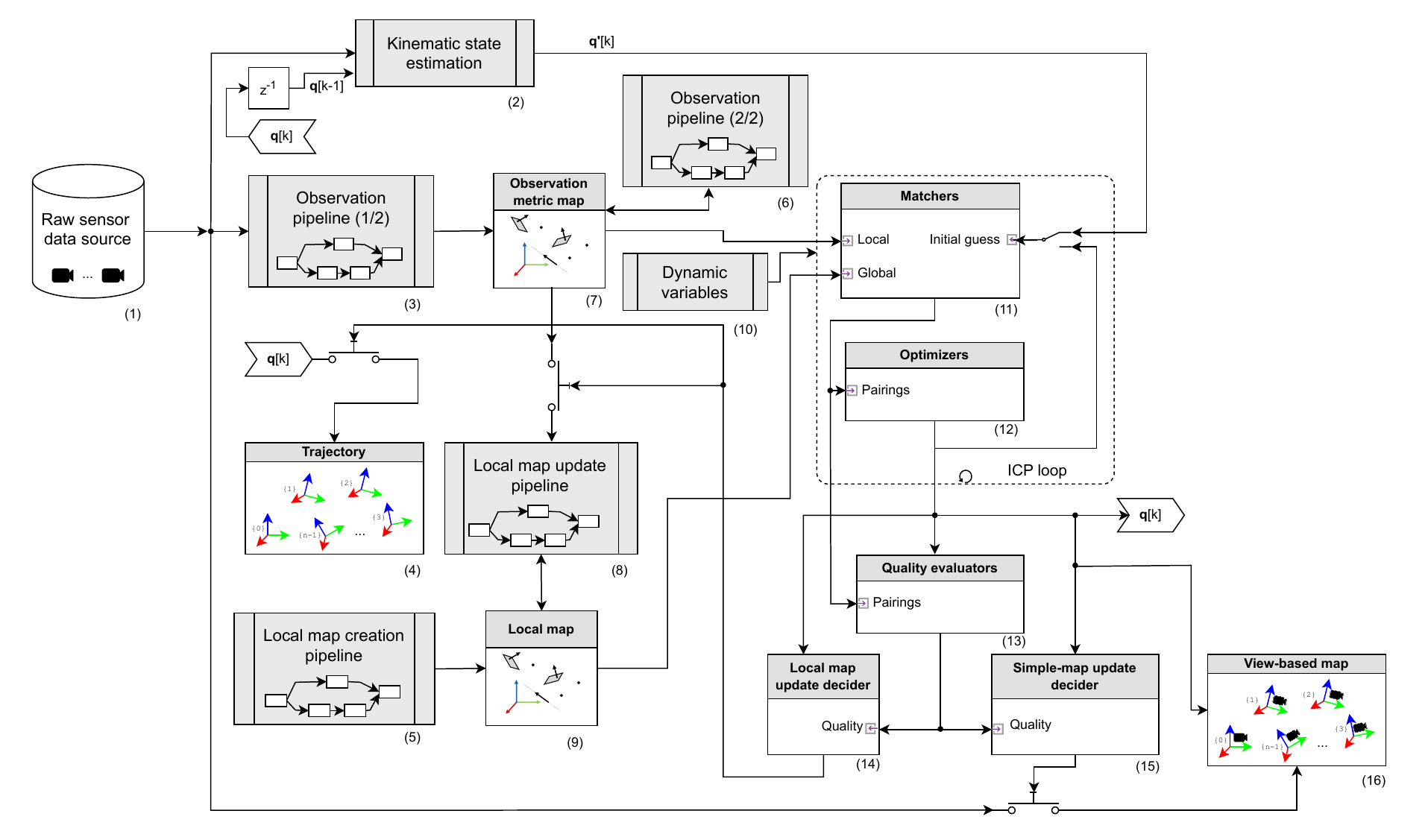}
\caption{Architecture of the proposed generic optimization-based LiDAR odometry system. Each process is highly configurable and parameterizable. See discussion in \S\ref{sect:lidar.odometry}.
}
\label{fig:lo-module}
\end{figure*}

\begin{equation}
\label{eq:deskew}
\begin{bmatrix}
\mathbf{r}_i\\
1
\end{bmatrix}
= 
\begin{bmatrix}
\Rot{s} & \Pos{s}\\
\mathbf{0}_{1\times 3} & 1
\end{bmatrix}
\begin{bmatrix}
\left(t_i ~ \AngVel{s}{s} \right)^\wedge & t_i ~ \Vel{s}{s}\\
\mathbf{0}_{1\times 3} & 1
\end{bmatrix}
\begin{bmatrix}
{}_s\mathbf{\tilde{r}}_i\\
1
\end{bmatrix}
\end{equation}

\noindent with ${\cdot}^\wedge$ the hat or wedge operator, mapping SO(3) tangent space vectors to
rotation matrices \citep{blanco2021tutorial}.

\subsubsection{Time-stamp adjustment}: Note that the exact moment used as a reference
for relative time-stamps in the de-skewing filter above is not defined.
Different conventions can be found in existing implementations, with the most common ones being
using the first point or the one in the middle as references. This decision is coupled with the
exact moment at which the vehicle pose and velocity are to be estimated by the LO or SLAM system, 
and for vehicles moving fast the difference between criteria may be in the order of tens 
of centimeters. Further complications arise from the diversity of per-point timestamp formatting in 
public datasets and sensor drivers, ranging from relative times in seconds, relative times in the
fixed range $[0,1]$, to just using absolute time stamps. 
Therefore, we provide this filter to adjust the timestamps in all incoming LiDAR scans
to ensure they observe one of a list of possible criteria.

\subsubsection{Down-sampling}: As shown while discussing related LO works (Section~\ref{sect:related.lo}),
sub-sampling point clouds is essential to achieve stable ICP-based mapping.
Our framework offers
several down-sampling filters, ranging from simply picking the first point within each 3D voxel
for a given uniform volumetric resolution, to keeping the average of each such voxels, or using 
non-linear spatial tessellation.

\subsubsection{Spatial filtering}: We also provide ternary filters to split an incoming map layer
in two categories: those points passing a given criterion and those that do not.
Filters include applying 3D bounding boxes (i.e. points within the box pass the test),
checking for a minimum distance to a given point (e.g. the vehicle),
or splitting by LiDAR ring number.

\subsubsection{Map insertion}:
\label{sect:pipeline.block.map.insertion}
 This fundamental binary filter takes an input point cloud layer
and an output layer of an arbitrary type and uses the abstract API of metric maps to merge
the cloud as if it was a single observation (e.g. from a 2D or 3D LiDAR) into the target.
A primary use of this filter is local map updating in our LO system, but it is also key for
map building in post-processing pipelines. 
Note that this operation generalizes several particular algorithms depending on the
class of the target map: point cloud insertion into a voxel-based point cloud,
ray-tracing on a 2D gridmap, raytracing and occupancy update for 3D volumetric grids, etc. 

\subsubsection{Dynamic object removal}: 
\label{sect:filter.dyn.obj.removal}
This filter takes an input layer with
an occupancy voxel map and uses its occupancy information to remove points from another point-cloud layer
belonging to voxels with a reduced occupancy likelihood, meaning that those locations were sensed probably
due to dynamic objects moving during the mapping process. Note that for a voxel to have a reduced occupancy probability
it needs to have been observed more often free than occupied.

\subsection{LiDAR odometry module}
\label{sect:lidar.odometry}

At this point, most basic blocks have been defined so we can introduce the architecture of the
proposed LO module. Please, use Figure~\ref{fig:lo-module} as a guide to the following discussion.
In the figure, $\mathbf{q}[k]$ stands for the kinematic state of the vehicle at the discrete time step $k$, 
and it includes the vehicle body ("$b$") pose $\Pose{b}\in SE(3)$ with respect to the local map
and its linear $\Vel{b}$ and angular $\AngVel{b}$ velocities.
Also, keep in mind that such architecture remains the same for 2D or 3D mapping, for local maps as 2D gridmaps or as 3D point-clouds, etc.
since those details are provided inside the configurable ``pipeline'' modules described later on.

Starting by the left of Figure~\ref{fig:lo-module}, raw sensor observations arrive from either an offline or live source 
(block \#1 in the diagram; see Section~\ref{sect:dataset.sources})
and fed three subsystems:
(i) the kinematic state estimation subsystem (block \#2), mainly in charge of fusing sensors (wheel odometry, IMU, etc.) and past 
LO outputs ($\mathbf{q}[k-1]$) to make \emph{short-term predictions} about the vehicle pose and velocity ($\mathbf{q}'[k]$); 
(ii) a first metric mapping pipeline (block \#3; see $Section~\ref{sect:pipelines}$), typically in charge of building filtered and down-sampled versions
of the incoming LiDAR scans into the \emph{observation} metric map data structure (see Section~\ref{sect:metric.maps}); 
and (iii) the \REVIEW{view-based map} (block \#16), one of the main LO outputs.

As discussed earlier, this work follows a scan-to-map design pattern
since it has proven benefits for 3D LiDAR mapping.
Hence, after using raw observations to build a metric map (block \#7),
it is aligned against a local metric map (block \#9),
which is initialized via a special instance of a metric map pipeline (block \#5)
that normally will just create empty metric map layers. This block is run only once
at system start up or when it is reset.
The local metric map data structure (block \#9),
is selectively updated using a custom pipeline algorithm (block \#8)
using one or more layers from the observation metric map (block \#7)
when certain criteria are met, as determined by a so-called local map update decider  (block \#14).
An independent decider (block \#15), normally with different criteria, is used
to determine what frames become key-frames in the output \REVIEW{view-based map}.
In general, such maps would benefit from sparser key-frames to reduce the size and memory requirements of final maps,
while the local map tends to benefit from more frequent updates from incoming sensor frames. 
However, note that most past works (e.g. \cite{deschaud2018imls,vizzo2023kiss}) 
update the local map for every single incoming sensor frame. 
This may be optimal in driving scenarios, but our experimental results show that this policy
may make the system 
\REVIEW{in other scenarios, as analyzed in Section~\ref{sect:ablation.localmap.updates}.}

Pose tracking itself is achieved by means of an ICP-like iterative optimizer (blocks \#11 and \#12),
in charge of finding the SE(3) pose $\mathbf{q}[k]$ that best explains the observation given a local map.
This optimization procedure is detailed in Section~\ref{sect:icp.pipelines}. 
For now, it is enough to abstract it as an algorithm taking as inputs: (i) the initial guess 
$\mathbf{q}'[k]$ from the kinematic state estimator, (ii) a local and a global metric map 
(with different geometric entities and map layers depending on the application),
and (ii) a set of dynamic variables (block \#10)
to adapt the behavior of the internal ICP pipeline (see Section~\ref{sect:dyn.vars}).

Once the optimizer converges, a set of quality evaluation algorithms (block \#13; see Section~\ref{sect:icp.pipelines.quality}) 
are used
to estimate the quality of the optimization.
This metric, together with other conditions (not shown in the diagram for the sake of clarity) 
are then used to conditionally trigger the two {decider blocks}: (i) one to accept the alignment as good enough 
to append the obtained pose to the estimated trajectory (block \#4) and to update the local metric map (block \#9)
via the purposely designed pipeline  (block \#8), and (ii) another one (block \#15) to add new key-frames to the
output map (block \#16).
Note that a localization-only mode exist (see Section~\ref{sect:results.localiz}), which disables both deciders while still producing an estimate of the vehicle state for each time step 
in $\mathbf{q}[k]$ by aligning the current observation against a prebuilt local map.

When the system is initialized, all blocks in Figure~\ref{fig:lo-module} are populated
with objects created dynamically using class factories as described in a human-readable 
configuration file (a YAML file, \cite{evans2017yaml}),
turning our framework into a flexible system easily re-configurable
and extensible by end users, enabling systematic investigations on the effects of changes in individual sub-systems.

Subsequent sections give further details on the ICP loop and the particular configuration used for 3D LiDAR mapping.

\subsection{Adaptive behavior: dynamic variables}
\label{sect:dyn.vars}

All software blocks in the LO system (Figure~\ref{fig:lo-module}) have parameters that
are initialized at system start-up. Some such parameters, however, would benefit from
being dynamically adapted depending on sensed conditions like the size of the environment 
(indoors vs. outdoors) or the motion profile (driving, handheld, drone, etc.).
For example, take the voxel-based down-sampling mechanism used in IMLS-SLAM \citep{deschaud2018imls}, 
which used a fixed resolution of 1.0 meters, adequate for driving datasets. If the SLAM system
is initialized in a small office-like environment instead, a much smaller resolution (i.e. less down-sampling)
would be needed indeed. Instead of relying on manual tuning of all those parameters, our
system would automatically detect the approximate size of the environment and
initialize the local map and down-sampling resolutions accordingly.

To make this adaptation possible, our framework offers two features:
(i) supporting mathematical expressions and basic scripting directly in the parameter specification files, 
and (ii) defining dynamic variables that can be used in such expressions.
On the former, we use the C++ Mathematical Expression Toolkit Library (ExprTk) \citep{partow2015c++}
due to its lightweight and efficient implementation.
On the latter, a set of variables are defined so they can be used as symbols in the expressions,
which are re-evaluated for each time step, and even for each ICP iteration within a given time step.
Next we review the most relevant such variables and how they are updated.

\subsubsection{Maximum sensor range}: 
\label{sect:max.sensor.range}
This is a fundamental variable as it roughly indicates the 
size of the environment and can be used as a hint to the required down-sampling resolution.
We define two such variables, the immediate maximum sensor range $r_{max}[k]$ for time-step $k$, 
and the low-pass filtered version $\hat{r}_{max}[k]$ using a first-order IIR (Infinite Impulse Response)
discrete-time filter with z-transform transfer function: 

\begin{equation}
\label{eq:low.pass.filter}
H(z) = \frac{1-\alpha}{z-\alpha} \quad \text{, with:} \quad \alpha = e^{-T \omega_c}
\end{equation}

\noindent where $\alpha$ is a fixed parameter (typically in the range [0.9,0.99]) that defines the low-pass cut frequency $\omega_c$ for
a sensor rate $1/T$.

\subsubsection{Related to robot kinematic state}: The latest vehicle state $\mathbf{q}[k]$, including
pose and linear and angular velocities in the map frame, 
are available as dynamic variables. For example, the local map update decider 
(block \#14 in Figure~\ref{fig:lo-module}) checks for a minimum linear and angular distances
between map updates. Our reference pipeline for 3D LiDAR mapping uses a heuristic formula for these
distances that takes into account both, the maximum sensor range, and the instantaneous robot angular velocity,
reducing the number of map updates while the sensor is experimenting higher accelerations.
The insight behind this policy is that, despite scan de-skewing, sensed points may be less accurate under these conditions 
since the assumption of constant velocity during each scan sweep is probably less accurate for higher angular velocities.
Variables exposing the instantaneous robot pose ($\mathbf{q}[k]$) are also fundamental in the context of map post-processing, for example, for distance-based filtering
while removing interferences from the vehicle itself or the person carrying a hand-held sensor.

\begin{figure}
\centering
\includegraphics[width=0.99\columnwidth]{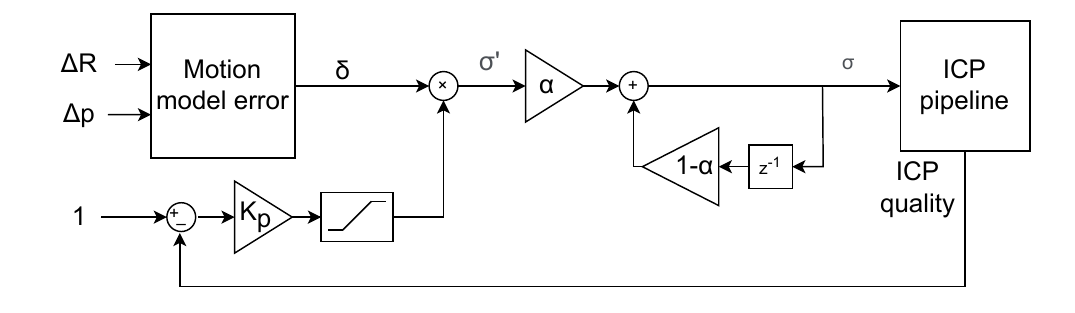}
\caption{The closed-loop proportional feedback controller strategy to dynamically adapt the 
ICP parameters, scaled by a threshold $\sigma$, based on both, 
errors in motion model predictions, and the overall ICP quality (in the range [0,1]). 
The calculation of the motion model error $\delta$ is given in \Eq{eq:delta.motion.error}.
Note that the input set-point for the controller is $1$, the ideal quality.
}
\label{fig:sigma.closed.loop}
\end{figure}

\subsubsection{Adaptive matching threshold}:
\label{sect:adaptive.sigma}
We have incorporated the idea of a dynamic threshold, proposed in KISS-ICP, to parameterize the data association and optimization stages of the ICP optimization loop
according to perceived changes. As justified geometrically in \cite{vizzo2023kiss}, given translational and rotational
corrections $\boldsymbol{\Delta} \mathbf{p} \in \mathbb{R}^3$ and $\boldsymbol{\Delta} \mathbf{R} \in SO(3)$, respectively,
of the ICP optimized pose with respect to the initial guess from
the kinematic state prediction (Section~\ref{sect:navstate.fuse}), the expected maximum point-to-point error $\delta$
can be approximated with: 

\begin{equation}
\label{eq:delta.motion.error}
\delta = \left\lVert \boldsymbol{\Delta} \mathbf{p} \right\rVert 
+
 2 r_{max} \sin \frac{1}{2} \left\lVert \REVIEW{(\log\boldsymbol{\Delta} \mathbf{R})}^{\vee} \right\rVert
\end{equation}

\noindent where $\log \mathbf{R}$ is the matrix logarithm, the averaged maximum sensor range $r_{max}$ is described in Section~\ref{sect:max.sensor.range}, 
\REVIEW{and $(\cdot)^\vee$ is the \emph{vee}-operator, the inverse of the wedge operator used in \Eq{eq:deskew}; that is, the $3 \times 3$ rotation correction matrix $\boldsymbol{\Delta} \mathbf{R}$ is first mapped to the tangent space via the matrix logarithm giving a $3 \times 3$ skew-symmetric matrix with only three degrees of freedom, then the \emph{vee}-operator simply takes out those three values as a $3 \times 1$ vector.}

With the idea of obtaining a threshold $\sigma$ to scale data association and optimization within ICP 
out of sequences of $\delta$ values over time, KISS-ICP proposes taking the standard deviation of all $\delta$ values,
saturating with a minimum $\sigma_{min}$ to prevent degeneration after long runs with good motion model predictions.
While quite successful for driving scenarios, this approach lacks the ability to quickly react against abrupt changes
in conditions (e.g. large occlusions, crossing a door moving from outdoors to an indoor area or handheld sensors with high dynamics).
In order to further improve the adaptability of the whole system, we propose using a closed-loop proportional negative feedback
controller to cope with sudden perturbations in the quality of ICP alignments,
as sketched in Figure~\ref{fig:sigma.closed.loop}.
This controller generates a multiplicative term $c$ that makes $\delta$ to grow ($\sigma' = c \delta$) 
whenever ICP quality starts to drop,
promoting the search for additional pairings in the next time step by increasing the search radius for nearest neighbors.
A first-order low pass filter, identical to \Eq{eq:low.pass.filter}, is then applied to obtain the actual 
threshold $\sigma$.
As it is well-known in control theory, a proportional controller always
exhibits an offset error, hence the obtained ICP quality will never reach the set-point, but in practice most often remains in the range $[0.9, 1]$.

\subsection{ICP optimization loop}
\label{sect:icp.pipelines}

One of the central blocks of the LO pipeline is the
flexible ICP algorithm, comprising blocks \#11 and \#12
in Figure~\ref{fig:lo-module}.
As the rest of the LO system,
those blocks are dynamically created from a plain text configuration
file describing the algorithms to use for data association, optimization, 
what metric map layers to use, and so on.

Our implementation is ``ICP-like'' in the sense that
the main components of a classical ICP loop remain
(i.e. data association, correction of the estimated transformation, and repeat until convergence)
but many key differences exist:
\begin{itemize}
\item Nearest neighbor search for each input point is generalized
into a set of matching algorithms (see Section~\ref{sect:icp.matchers}).
\item Several options are available as optimal pose \REVIEW{solver algorithms from a set of predefined correspondences} (see Section~\ref{sect:icp.solvers}).
\item An optional metric map pipeline algorithm can be applied inside the ICP loop itself (see \ref{sect:icp.local.map.update}).
\item Dynamic variables (discussed in Section~\ref{sect:dyn.vars}), including the ICP iteration count, can be used to parameterize any of the components above. This allows, for example, modifying the search radius over ICP iterations within one single time step (useful for loop closure).
\item An optional probabilistic prior distribution of the sought pose is accepted as input, allowing
prior knowledge to constraint the otherwise unbounded distance between the initial guess and the final result.
\item Generation of log records to debug and visually inspect the internals of the algorithm during the different ICP iterations 
has been a central feature kept in mind during our implementation. As far as we know, no other public ICP implementation allows inspecting the evolution of the pairings and all other variables over ICP iterations in a systematic way.
\item Importance is given to quality evaluators with a placeholder for such algorithms (see Section~\ref{sect:icp.pipelines.quality}), fundamental in assessing whether a map registration is successful or not.
\end{itemize}

Keep in mind that the same ICP structure is to be used during online mapping (LiDAR odometry), localization, 
and for loop closure (see Section~\ref{sect:lc}), although with different parameters and metric map layers.

A step-by-step description of the proposed ICP method is sketched in Algorithm~\ref{algo:icp}
and is explained next, 
where many low-level details have been omitted for clarity.
Overall, the ``I'' for \emph{iterative} from the original ICP method remains
as a fundamental feature required to solve such a strongly non-linear problem
as registering two metric maps with unknown pairings.
The loop is repeated until one of two conditions is reached: 
a maximum number of iterations $N_{max}$ (line 9 in the pseudocode)
or convergence (line 23). Successful registrations most often end up with convergence, hence 
$N_{max}$ is set to a large enough number ($N_{max} \ge 300$) to avoid undesired search interruptions.
Convergence is defined as small enough changes of the translational and rotational parts 
of the estimated transformation $\Pose=\{\Pos, \Rot\}$ with respect to the
previous step $\Pose'=\{\Pos', \Rot'\}$, that is, 
whether $\delta p < \varepsilon_p$ and $\delta R < \varepsilon_R$ with:

\begin{subeqnarray}
\label{eq:icp.convergence}
\delta p &=& \left\lVert \Pos - \Pos \right\rVert \\
\delta R &=& \left\lVert \log \left( \Rot^{\Inv} \Rot' \right)^\vee \right\rVert
\end{subeqnarray}

\noindent and $\varepsilon_p$ and $\varepsilon_R$ the corresponding thresholds,
which are set to $10^{-4}$ and $5\cdot 10^{-5}$, respectively,
in all presented experiments.
For each iteration, all matching algorithms are sequentially invoked 
over the two input metric maps
to populate a list of potential pairings $p$ (lines 11--14),
then solvers are invoked using this list (lines 15--20)
until a successful result $\mathbf{T}'$ is obtained (see Section~\ref{sect:icp.solvers}).
The optional input hook for invoking a metric map updating pipeline is 
invoked next (line 21), which enables our claimed new feature of
fluidly deskewing the input point cloud as the estimated pose is updated within the ICP iterations (see Section~\ref{sect:icp.local.map.update}).
The loop is run until convergence, then quality evaluation algorithms are
applied (lines 29--35) and their result is returned together with the estimated pose.

\begin{algorithm}[t]
\caption{Generic ICP-like optimization algorithm}
\label{algo:icp}
\scriptsize
\begin{algorithmic}[1]
\Function{align}{$m_l, m_g, \mathbf{T}_0, \mathbf{T}_p=\left\{ \bar{\mathbf{T}}_p,\boldsymbol{\Sigma}_p \right\}, \{M_i\}, \{S_i\},\{Q_i\}, \mathfrak{p} $}
	\LineComment{$m_l$ and $m_g$ are the local and global metric maps}
	\LineComment{$\mathbf{T}_0$ is the initial guess for the SE(3) transformation from local to global}
	\LineComment{$\mathbf{T}_p=\left\{ \bar{\mathbf{T}}_p,\boldsymbol{\Sigma}_p \right\}$ is an optional prior Gaussian for sought transformation}
	\LineComment{$\{M_i\}$, $\{S_i\}$, $\{Q_i\}$: sets of matchers, solvers, and quality evaluators.}
	\LineComment{$\mathfrak{p}$ is an optional metric map pipeline}
	
    \State $\mathbf{T} \gets \mathbf{T}_0$ \Comment{Initialize optimal SE(3) transform}
    \State $k \gets 0$ \Comment{Initialize iteration counter}
    \While{$k < N_{max}$}
      	\State $\textsc{UpdateDynamicVariables}()$ \Comment{For matchers and solvers}        	 

	    \State $p \gets \{\}$ \Comment{Initialize pairings with empty set}
        \ForEach{$M \in \{M_i\}$} \Comment{For each matcher} 
        	\State $p \gets p \cup M.\textsc{match}(m_l,m_g,\mathbf{T})$ \Comment{Data association}        	 
        \EndFor

        \ForEach{$S \in \{S_i\}$} \Comment{For each solver} 
        	\State{$\{\mathbf{T}', V\} \gets S.\textsc{solve}(p, \mathbf{T}_p)$} \Comment{Optimal SE(3) solvers}
        	\If{$V$} \Comment{Valid solution?}
        		\State{\textbf{break}}  \Comment{No need to try more solvers}
        	\EndIf
        \EndFor

        \State $\mathfrak{p}.\textsc{Apply}(m_l)$ \Comment{Apply observation pipeline (2/2)}

        \State $C \gets \textsc{CheckConvergence}(\mathbf{T},\mathbf{T}')$ \Comment{See \Eq{eq:icp.convergence}}
        
        \State $\mathbf{T} \gets \mathbf{T}'$ \Comment{Update estimation}
        
        \If{C} \Comment{Convergence?}
            \State \textbf{break}
        \EndIf

		\State $k \gets k + 1$ \Comment{Increment iteration counter}
    \EndWhile
    
	\State $q \gets 0 \quad, W \gets 0 \quad, bad \gets False$ \Comment{Initialize quality metric}
    \ForEach{$\{Q,w_i\} \in \{Q_i\}$} \Comment{For each quality algorithm} 
		\State ${q_i,bad_i} \gets Q.\textsc{evaluate}(\mathbf{T},m_l,m_g)$ \Comment{Evaluate metrics}
		\State $q \gets q + w_i ~ q_i$ \Comment{Accumulate}
		\State $bad \gets bad \text{~or~} bad_i$ \Comment{Logic or}
    \EndFor
	\State $q \gets bad ~ ? ~~ 0~ :~ q/W$ \Comment{Average}
    
    \State \textbf{return} $\{\mathbf{T}, q \}$ \Comment{Return estimated transformation and quality}
\EndFunction
\end{algorithmic}
\end{algorithm}

\subsubsection{Matchers}:
\label{sect:icp.matchers}
Different kinds of geometric pairs can conceivable be identified
between two metric maps: point-to-point, point-to-plane, point-to-line, plane-to-plane, etc.
Furthermore, the input local and global maps to be registered can be of different types:
point clouds, grid maps, voxel maps, etc.
Therefore, our framework provides a set of different match search algorithms
depending on the kind of geometric features to find, 
while the map variety is handled via a generic nearest neighbor (NN) abstract interface. 
It makes sense to have more than one matching algorithm in multiple situations: 
(i) looking for successive pairings for each local map point in a prioritized list
(e.g. first, try to find a point-to-plane pairing, but if it is not possible, then a point-to-point match), 
(ii) when several metric map layers exist and each local map layer is to be matched against a particular 
layer in the global map, (iii) when different such layers represent differentiated geometric features (e.g. planes, edges, or pole-like objects).

Regarding efficient parallelized implementation of matchers, 
we follow the findings reported in KISS-ICP \citep{vizzo2023kiss} and SIMPLE \citep{bhandari2024minimal}
about the superiority of \emph{Intel Threading Building Blocks} (TBB) \citep{tbb}
over other alternatives to exploit multi-core CPUs, hence our implementation also employs that
library to parallelize inner loops of most algorithms handling point clouds.
A more detailed description of all available matchers and benchmarking them against different metric map data structures
is out of the scope of this paper and is left for future works.

\subsubsection{Solvers}:
\label{sect:icp.solvers}
Having multiple solvers is in order due to the diversity of geometric entity combinations
that can be found by matchers.
\REVIEW{In particular, our implementation offers the classic Horn's method in 
\cite{horn1987closed} (apt for point-to-point pairings)
and a generic Gauss-Newton solver suitable for different types of 
geometries.
In the following we are focusing on
point-to-point pairings since they have been shown to lead to good results \citep{vizzo2023kiss}. 
Point-to-plane correspondences are also used for our 3D-NDT system configuration, but their equations are left out for space reasons
and will be described somewhere else.
More diverse geometric primitives are worth further investigation, e.g. see MULLS \citep{pan2021mulls}}. 

Therefore, assume $p$ in Algorithm~\ref{algo:icp} contains a set of $N$ 
local $\{ \mathbf{l}_i \}_{i=1}^N$ and global $\{ \mathbf{g}_i \}_{i=1}^N$ points in $\mathbb{R}^3$
that are supposed to correspond to each other.
Solving for the optimal pose $\Pose^\star \in SE(3)$ that minimizes the overall L2 registration error:

\begin{equation}
\label{eq:optimal.pose.ls}
\Pose^\star = \arg\min_{\Pose} \sum_{i=1}^N \lVert \underbrace{\Pose ~ \mathbf{l}_i  - \mathbf{g}_i }_{\mathbf{e}_i(\Pose)} \rVert ^2 = 
\arg\min_{\Pose} \sum_{i=1}^N  \left\lVert \mathbf{e}_i(\Pose)\right\rVert^2
\end{equation}

\noindent is a classic problem with well-known closed-form solutions
such as the quaternion method in \cite{horn1987closed}.
Unfortunately, in practice all pairings are not to be trusted equally
due to the uneven and discrete nature of LiDAR sampling, sensor noise, and 
the existence of dynamic objects.
Although our system offers Horn's algorithm among the possible solvers,
it is far from being the optimal choice 
\REVIEW{(see Section~\ref{sect:ablation.horn})}.
Recent research has proven that robust (non least-squares) problem formulations
lead to superior performance in the presence of pairing outliers and local minima, 
either using truncated least-squares and 
semidefinite programming (SDP) relaxation \citep{briales2017convex,yang2019polynomial,yang2020one}
or by simply applying a robust kernel $\rho(\cdot)$ (e.g. Huber, Geman-McClure,...) to \Eq{eq:optimal.pose.ls} \citep{chebrolu2021adaptive}:

\begin{equation}
\label{eq:optimal.pose.robust}
\Pose^\star = \arg\min_{\Pose} \sum_{i=1}^N \rho ( \REVIEW{ \left\lVert \mathbf{e}_i(\Pose)\right\rVert } )
\end{equation}

In this work we follow this latter approach for its simplicity and superior efficiency in terms of running time.
Furthermore, we propose adding an additional term to the cost function integrating an optional
probability distribution reflecting any prior knowledge about the relative map poses. 
In the context of LO, this term incorporates vehicle pose predictions from the kinematic state estimator
(block \#2 in Figure~\ref{fig:lo-module}, see Section~\ref{sect:navstate.fuse}),
while in the context of loop closure, it reflects uncertainty built up along a given topological loop (see Section~\ref{sect:lc}).
In any case, such prior information is modeled as a Gaussian distribution with mean $\bar{\mathbf{T}}_p \in SE(3)$ 
and covariance matrix $\boldsymbol{\Sigma}_p$. 
This prior term, denoted as $\mathbf{e}_p$, is not affected by the robust kernel and is introduced as a Mahalanobis distance 
of the \emph{vee}-operator applied to the Lie group logarithm map of the pose mismatch, that is:

\begin{equation}
\mathbf{e}_p(\Pose)=\left( \log \bar{\mathbf{T}}_p^{-1} \Pose \right)^{\vee} 
\end{equation}

\noindent leading to the final robustified least-squares cost function:

\begin{eqnarray}
\label{eq:optimal.pose.robust2}
\Pose^\star &=& \arg\min_{\Pose} c(\Pose) \\ 
\label{eq:optimal.pose.robust2.b}
c(\Pose)&=&
 {\left\lVert  \mathbf{e}_p(\Pose) \right\rVert}_{\boldsymbol{\Sigma}_p}^2
 +
 \sum_{i=1}^N \rho ( \REVIEW{ \left\lVert \mathbf{e}_i(\Pose)\right\rVert } )
\end{eqnarray}

\noindent with $\mathbf{e}_i(\Pose)$ the point-to-point pair costs defined in \Eq{eq:optimal.pose.ls}.

Solving \Eq{eq:optimal.pose.robust2}--(\ref{eq:optimal.pose.robust2.b}) is a non-linear optimization problem, hence we propose
using the iterative Gauss-Newton (GN) algorithm. 
Since the search space is non-Euclidean, the Lie group perturbation-model approach is employed \citep{sola2018micro,blanco2021tutorial}
where the unknown becomes an increment $\MyEps \in \mathfrak{se}(3)$ (SE(3) Lie group tangent Euclidean space):

\begin{eqnarray}
\MyEps^\star &=& \arg\min_{\MyEps} c(\Pose \boxplus \MyEps) \\
\Pose^\star &=& \Pose \boxplus \MyEps^\star
\end{eqnarray}

\noindent which, using the right-hand perturbation convention:

\begin{equation}
\Pose \boxplus \MyEps = \Pose ~ e^{\MyEps}
\end{equation}

\noindent allows us finding the required Jacobians to solve the normal equation of 
each GN iteration: 

\begin{equation}
\mathbf{H} \MyEps^\star = -\mathbf{g}
\end{equation}

\noindent with $\mathbf{H}$ and $\mathbf{g}$ the $6 \times 6$ Hessian matrix and $6 \times 1$ gradient vector, respectively:

\begin{eqnarray}
\mathbf{H} &=& \frac{\partial \mathbf{e}_p}{\partial \MyEps}^\top \boldsymbol{\Sigma}_p^{-1}  \frac{\partial \mathbf{e}_p}{\partial \MyEps}
+ \sum_{i=1}^{N} \sqrt{w( \lVert \mathbf{e}_i \rVert^2 )} 
\frac{\partial \mathbf{e}_i}{\partial \MyEps}^\top \frac{\partial \mathbf{e}_i}{\partial \MyEps} 
\nonumber 
\\ ~
\\
\mathbf{g} &=& \frac{\partial \mathbf{e}_p}{\partial \MyEps}^\top \boldsymbol{\Sigma}_p^{-1}  \mathbf{e}_p
+ \sum_{i=1}^{N} \sqrt{w( \lVert \mathbf{e}_i \rVert^2 )}
\frac{\partial \mathbf{e}_i}{\partial \MyEps}^\top \mathbf{e}_i
\end{eqnarray}

\noindent with $\frac{\partial \mathbf{e}_p}{\partial \MyEps}$ given by
\cite[Eq.~(10.34)]{blanco2021tutorial}, 
$w(\cdot)$ the \emph{weight} function of the employed robust kernel,
and:

\begin{equation}
\frac{\partial \mathbf{e}_i}{\partial \MyEps} = 
\left(
 \begin{bmatrix}
 \mathbf{e}_i^\top ~ 1
 \end{bmatrix} \otimes \mathbf{I}_3
\right)
\frac{\partial \Pose e^{\MyEps}}{\partial \MyEps}
\end{equation}

\noindent where the expression for $\frac{\partial \Pose e^{\MyEps}}{\partial \MyEps}$ 
can be found in \cite[Eq.~(10.19)]{blanco2021tutorial} and $\otimes$ is the Kronecker product.
Our implementation actually allows matchers to include individual weights for each 
pairing, which scale the corresponding Jacobians in the two equations above,
but in the present work we just assigned all pairings equal weights. Past works, 
such as \cite{dellenbach2022ct}, have exploited such weighting to give higher priority
to more reliable points, hence this feature deserves further research.
The cost function and associated Jacobians for other geometric entities 
in our implementation (point-to-plane, point-to-line, etc.)
are left out of this work for space reasons and shall be described in future works.

A configurable number of GN iterations can be run, but note that GN is invoked
inside an outer ICP loop, hence running just one or two iterations is normally enough: new data association
for the next ICP iteration means the cost function is likely to change so it is not worth wasting time iterating to find
what will only become partial solutions.

\subsubsection{Local map update pipeline}:
\label{sect:icp.local.map.update}
As shown in line 21 of Algorithm~\ref{algo:icp}, an optional metric map pipeline (Section~\ref{sect:pipelines})
hook can be invoked \emph{within} each ICP loop iteration.
This feature is not used for loop-closure, but it is key for sequential processing of scans in odometry.
In the default configuration for 3D LiDAR odometry (Section~\ref{sect:icp.pipeline.default})
this hook is used for two operations: (i) refining the estimated vehicle linear and angular velocities, 
and (ii) using such new estimated velocities to apply the de-skewing filter (Section~\ref{sect:deskewing}) to the final layers of 
the metric map built from the current LiDAR scan (refer to ``Observation pipeline (2/2)'' in \Figure{fig:lidar3d.obs.pipeline}).

Linear $\Vel{s}{s}$ and angular velocity $\AngVel{s}{s}$ vectors of the sensor, both in the sensor frame ``$s$'',
are estimated under the assumption of constant velocity from the current sensor pose\footnote{Again, the actual implementation
takes into account the sensor pose within the vehicle, but that change of coordinates is neglected here for the sake of clarity.}
estimation $\Pose{s}_{k}$
in the $i$-th ICP iteration and the sensor pose from last time step $(\Pose{s})_{k-1}$ with a sensor period $T$:

\begin{eqnarray}
\label{eq:rel.pose.incr.1}
((\Pose{s})_{k-1})^{-1} \Pose{s}_{k} &=& \begin{bmatrix}
 & \Delta\Rot & & \Delta \Pos \\
 0 & 0 & 0 & 1
\end{bmatrix}_{4\times 4} \\
\label{eq:rel.pose.incr.2}
\Vel{s} =  \frac{1}{T} \Delta \Pos \quad && \quad \Vel{s}{s} = \Pose{s}_{k}^{\Inv} \Vel{s} \\
\label{eq:rel.pose.incr.3}
\AngVel{s} =  \frac{1}{T} (\log\Delta \Rot)^{\vee} \quad && \quad \AngVel{s}{s} = \Pose{s}_{k}^{\Inv}  \AngVel{s}
\end{eqnarray}

In practice, this simple method makes the velocity vectors to be estimated alongside with the
vehicle incremental pose, enabling fluidly de-skewing points during ICP iterations so they better match 
the reference local map.
As shown in the experimental results
\REVIEW{ and in the ablation study in Section~\ref{sect:ablation.twist}},
this allows the effective tracking of abrupt
motion profiles (see, e.g. Section~\ref{sect:ncd}).

Our approach has a similar aspiration than Continuous-Time ICP \citep{dellenbach2022ct},
but there are two differences:
(i) ours does not introduce pose discontinuities between the end of a scan and the beginning of the next one, and
(ii) ours does not need to use heuristic weights for the different parts of the optimization 
(point registration errors, final pose, and velocity vectors)
since velocities are implicitly defined from the iteratively refined final pose, and the latter is unconstrained.

\subsubsection{Quality estimators}:
\label{sect:icp.pipelines.quality}
Except in degenerate cases, all solvers discussed above such as the classical quaternion method \citep{horn1987closed}
or the GN iterative solver will always produce a result, correct or wrong depending on the existence of outliers in the input pairings. 
It hence becomes essential to assess whether the output relative pose between the two metric maps is plausible or should
otherwise be discarded. In this latter case, our framework discards those key-frames with an unacceptable ICP quality.
If this happens right at the beginning of mapping, the metric map is discarded and started from scratch again.
Otherwise, amid navigation, the kinematic state estimator extrapolates the predicted vehicle motion to try to recover
a good pose tracking in subsequent frames.

As seen in lines 29--35 of Algorithm~\ref{algo:icp}, we propose using the weighted average of a configurable 
set of quality evaluation algorithms. The abstract interface of such algorithms takes as input the two metric maps
that has been aligned and the tentative optimal relative pose found by solvers, and returns two elements: 
(i) a quality metric $q_i \in [0,1]$, and (ii) a short-circuit logic discard flag, which shall be activated when
the alignment is clearly wrong. In that latter case, the $q_i$ values returned by other quality evaluators are ignored
and the registration is discarded.

\begin{figure}
\centering
\includegraphics[width=0.99\columnwidth]{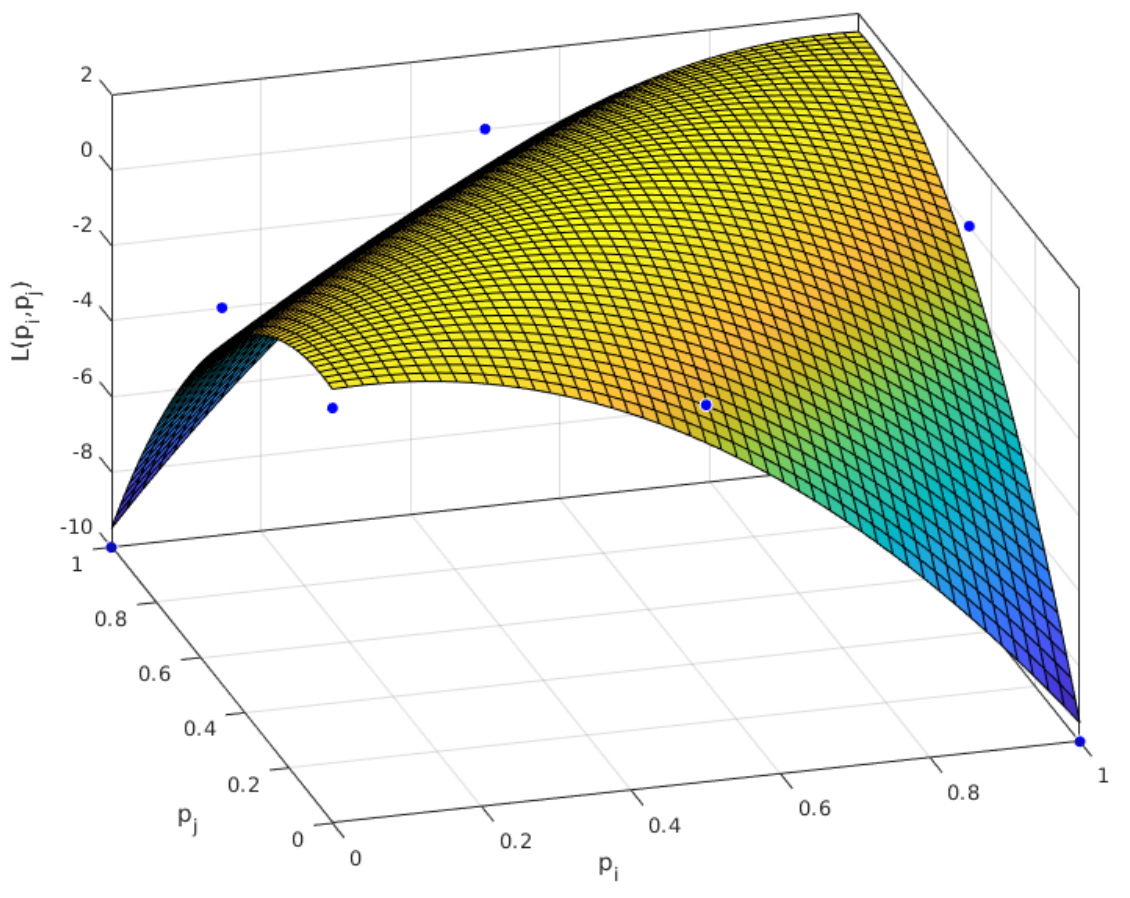}
\caption{Heuristic loss function to compare voxel occupancies $p_i, p_j ~ \in [0,1]$
from two maps $i$ and $j$ in the proposed alignment quality metric in Section~\ref{sect:icp.pipelines.quality}.
The exact equation represented here is 
$L(p_i,p_j)=1.5 + p_i + p_j - 12 p_i^2 + 22  p_i  p_j - 12 p_j^2$.
}
\label{fig:quality.loss.func}
\end{figure}

\begin{figure*}
\centering
\includegraphics[width=0.99\textwidth]{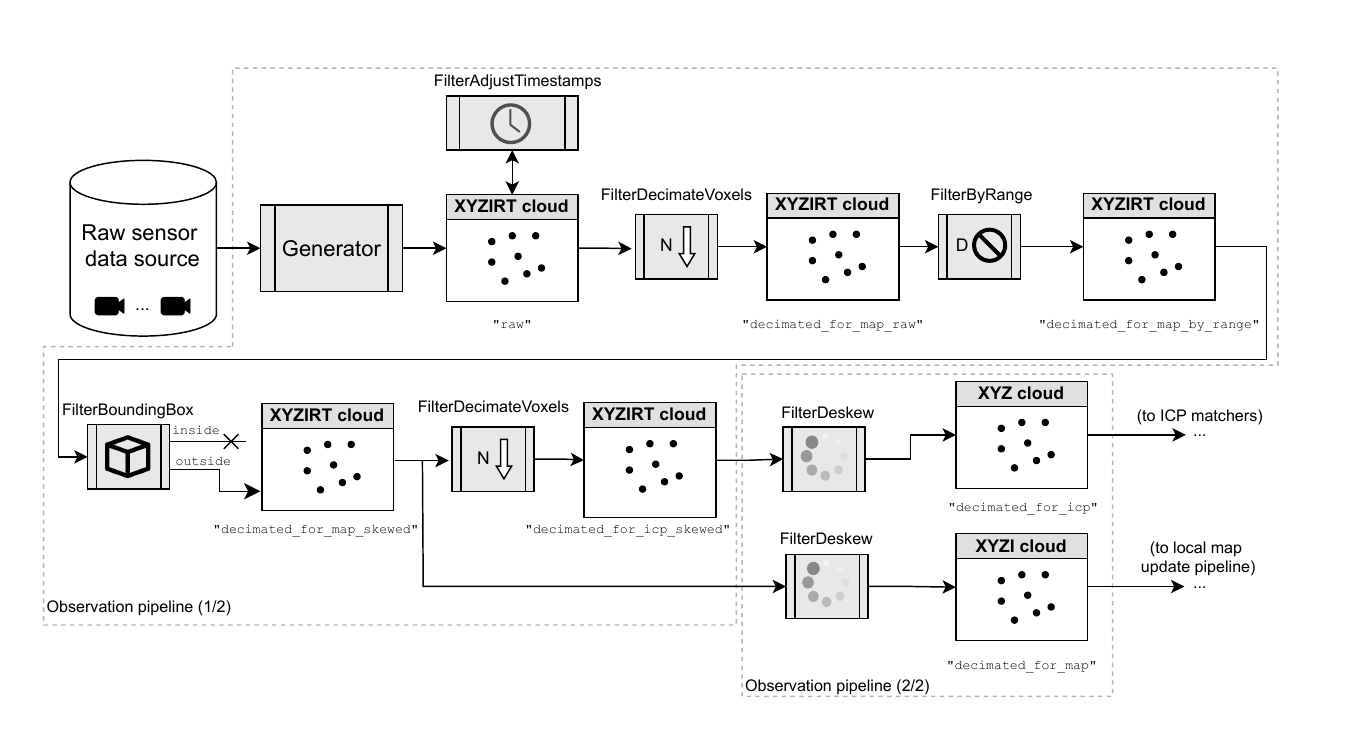}
\caption{Detailed view of the ``observation pipeline'' blocks in Figure~\ref{fig:lo-module} for the ``default 3D LiDAR'' LO system configuration.
Refer to discussion on pipelines in general in Section~\ref{sect:pipelines} and to the rationale behind this particular pipeline in Section~\ref{sect:icp.pipeline.default}.
}
\label{fig:lidar3d.obs.pipeline}
\end{figure*}

We have implemented three such algorithms:

\begin{itemize}
\item{\textbf{Point-wise paired ratio:} This simple method just uses the ratio of local map points that were assigned a pairing during data association
in the last ICP iteration:
\begin{equation}
q = \frac{N_{pairings}}{N_{local\_points}}
\end{equation}
This method is normally adequate for LO, although not informative enough for loop closure assessment.
}

\item{\textbf{Range reprojection:} An implementation of the method proposed in \cite{bogoslavskyi2017analyzing}, based
on re-projecting 3D points into a range image for comparison between the two maps.}

\item{\textbf{Voxel occupancy metric:} A novel algorithm that takes into account all volumetric occupancy information, 
and its mismatch between both metric maps. This method requires both maps to have an occupancy voxel layer, 
which is straightforward to add in our framework by adding such requirement to the metric map pipelines. 
The key insight is that voxels that are either free or occupied in both maps ``vote'' for a good quality,
but those with contradictory information ``vote'' for a bad quality with a stronger weight.
Inspired by Kullback-Leibler distance, we define a heuristic loss function $L(p_i,p_j)$ taking the occupancy probabilities $p_i$ and $p_j$ of each pair of voxels
(in the range [0,1]) 
corresponding to the same global coordinates for the two maps, with the shape illustrated in \Figure{fig:quality.loss.func}: 
similar occupancies have the highest score, any of the two map voxels being unobserved ($p \sim 0.5$) has a smaller score, and contradictory information
has a strong negative score. All $N$ voxels
are integrated into a single scalar quality $q$ value as:

\begin{equation}
q = \frac{1}{1 + e^{-\kappa d / N}} \quad \text{with:}\quad d = \sum_{k=0}^N L(p_i^k,p_j^k)
\end{equation}
\noindent where $\kappa$ is a heuristic scale parameter.
}
\end{itemize}

Apart of these quality metrics, note that the Gauss-Newton optimizer
described in Section~\ref{sect:icp.solvers} also provides an uncertainty estimation
for the solution, i.e. the Hessian matrix of the last iteration.
Although normally overconfident, its eigenvalues and eigenvectors can be also used to assess whether a given relative pose is well-conditioned.

\subsection{Default configuration for 3D LiDARs}
\label{sect:icp.pipeline.default}

Once the generic architecture of the LO module and the ICP algorithm
have been shown in Figure~\ref{fig:lo-module} and Algorithm~\ref{algo:icp}, respectively, we introduce a particular system configuration
which is dubbed \texttt{lidar3d-default} in the implementation
and which has been used in most experimental results.
A \emph{configuration} comprises the definition of what blocks populate the internals of generic blocks in Figure~\ref{fig:lo-module}, along with their parameters.

This configuration has been designed to successfully run in as many situations as possible, hence it is not particularly optimized for minimum tracking error
but for robustness and for its ability to run at sensor rates or faster.
Configurations with better performance for particular sensors and environments could be devised, but this is future work.

Observing Figure~\ref{fig:lo-module}, there are three ICP-related blocks to be defined (blocks \#11, \#12, \#13) and four metric map pipeline blocks 
(blocks \#3, \#5, \#6, \#8). Next we describe how they are set-up in this configuration.

\subsubsection{Local map creation pipeline}: A single map layer is created using a hashed voxel map holding point clouds, with a maximum of 20 points per voxel, only (X,Y,Z) attributes per point, and a voxel resolution of $1.5\%$ of the 
estimated sensor maximum range with a minimum and maximum of 0.5~m and 1.0~m, respectively. No minimum distance between points stored in each voxel is imposed.

\subsubsection{Observation pipelines}: Split in two parts, (1/2) and (2/2), 
with contents detailed in Figure~\ref{fig:lidar3d.obs.pipeline}. 
Recall that the pipeline is split to allow the first part to be invoked only once per incoming scan, and the latter once for each ICP iteration.
As sketched in the figure, a default generator first takes the raw sensor data 
and produces a point cloud map layer (``raw'') with contiguous memory layout (see Section~\ref{sect:metric.maps}). If per-point timestamps are present, they
are then adjusted, with the particular convention of using the middle of the scan as the time origin. Next, the cloud is down-sampled and nearby points removed (\texttt{FilterByRange} in the figure) to avoid polluting the map with vehicle or robot body measurements. Next, a bounding box filter is applied to 
remove nearby ceiling points, which have been shown to lead to divergence in
some scenarios. Then, this point cloud, which underwent decimation once, 
is decimated once again so we have two layers: a denser one to update the local map, and the sparser one to be used for registration in the ICP algorithm. 
This idea of using dual-resolution maps comes from its successful implementation
in KISS-ICP \citep{vizzo2023kiss}. Finally, note that de-skewing happens at the end, once for each down-sampled clouds, whereas all former works perform this step at the beginning instead.
The reason for this alternative placement is twofold: 
to enable the fluid de-skewing discussed in Section~\ref{sect:icp.local.map.update},
and increasing the efficiency since far less points are de-skewed in comparison
to performing this operation at the very beginning of the pipeline. 
The result of first down-sampling then de-skewing is not exactly the same
than switching the order of both operations, but in practice both methods lead to equivalent overall trajectory quality metrics.

\subsubsection{Local map update pipeline}: This pipeline comprises just one map insertion action (Section~\ref{sect:pipeline.block.map.insertion}), 
taking one of the observation map layer (the denser cloud) and merging it into the local metric map, effectively making the map to grow as the vehicle explores the environment.

\subsubsection{ICP-related blocks}: One matcher algorithm is defined to find out point-to-point correspondences between the local map
and one of the observation layers (the sparser cloud). Then, the Gauss-Newton solver described in Section~\ref{sect:icp.solvers} is defined as the unique solver. Following KISS-ICP \citep{vizzo2023kiss}, the matcher threshold and the solver robust kernel are parameterized with a dynamic threshold, although estimated in a different way (see Section~\ref{sect:adaptive.sigma}). Finally, quality for LO is assessed by using the matching ratio criterion (Section~\ref{sect:icp.pipelines.quality}).

\begin{figure*}
\centering
\includegraphics[width=1\textwidth]{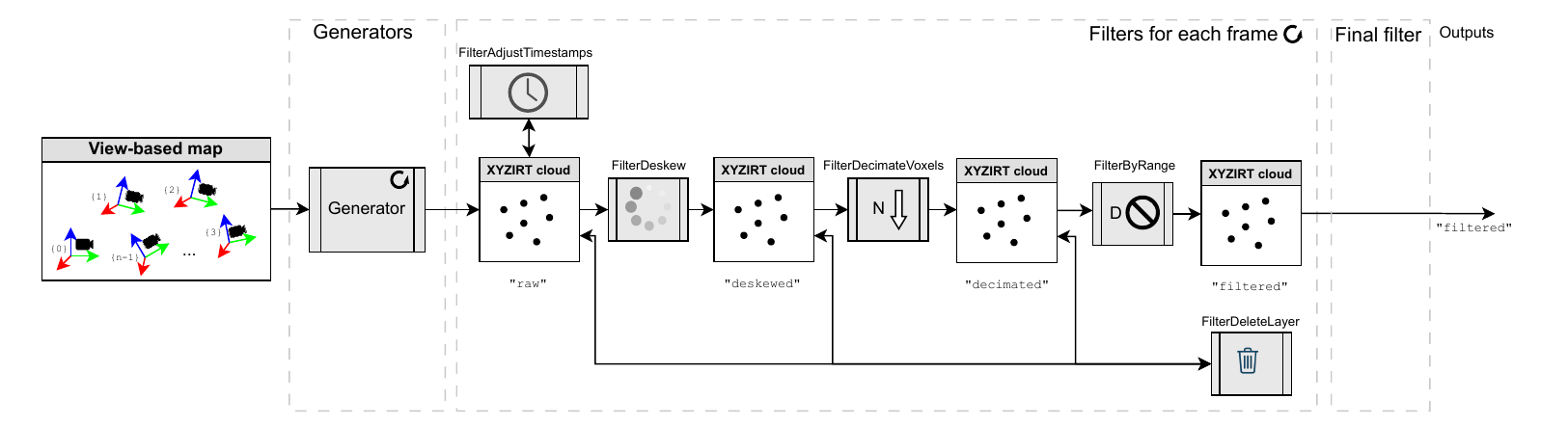}
\caption{Example configuration for building metric maps from view-based maps (using the application \texttt{sm2mm}): 
a basic pipeline for building one single point cloud by accumulating decimated versions of de-skewed LiDAR scans.
Refer to discussion in Section~\ref{sect:sm2mm}.
}
\label{fig:metric.map.pipeline.example.a}
\end{figure*}

\subsection{Kinematic state prediction}
\label{sect:navstate.fuse}
This role of this algorithm in the LO system (block \#2 in Figure~\ref{fig:lo-module})
is to generate short-term pose and velocity predictions for the vehicle for each incoming LiDAR scan,
based on the history of past sensor observations and former ICP registration outcomes.
Poses are needed to make the non-linear iterative search within ICP to start as close to the solution
as possible, and velocities are needed to perform the initial scan de-skewing.

We have experimented with two alternative methods: 
(i) simple kinematic extrapolation using only the two last vehicle poses, and 
(ii) a full probabilistic formulation based on factor graphs optimizing over a sliding window of past vehicle poses.
Initial results showed that the simple method gave more accurate trajectories and was less prone to divergence
under abrupt motions, hence we will only discuss the former method here and will leave the latter for future analysis.

The implemented method takes only two inputs: estimated poses from ICP optimization, and, optionally if available, wheels odometry readings.
Given the two last poses $\Pose{s}_k$ and $(\Pose{s})_{k-1}$ with elapsed time between them of $T$ seconds,
Eqs.~(\ref{eq:rel.pose.incr.1})--(\ref{eq:rel.pose.incr.3})
give us the estimated linear $\Vel{s}{s}$ and angular $\AngVel{s}{s}$ velocities, in the local sensor frame ``$s$''.
Then, when a kinematic state prediction is needed for an instant $\delta_t$ ahead of frame $\Pose{s}_k$,
velocities are left unmodified (constant velocity model assumption) and 
the predicted pose $\hat{\Pose{s}}_{k+1}$ is evaluated as follows. 
In the absence of wheel odometry, the former pose is extrapolated assuming constant velocities, that is: 

\begin{equation}
\label{eq:pose.prediction}
\hat{\Pose{s}}_{k+1}
= 
\Pose{s}_{k}
\begin{bmatrix}
\left(\delta_t ~ \AngVel{s}{s} \right)^\wedge & \delta_t ~ \Vel{s}{s}\\
\mathbf{0}_{1\times 3} & 1
\end{bmatrix}
\end{equation}

\noindent while, if wheel odometry is present, its values over time are used as a good approximation of 
short-term incremental motion and such increment $\Pose{k+1}{k}$ is just applied to the last frame, 
i.e. $\hat{\Pose{s}}_{k+1} = \Pose{s}_{k} \Pose{k+1}{k}$.

Note that an output from this prediction method is not available until at least two LiDAR scans have been processed
by the LO pipeline. In this initial time step, plus when there is a long 
period without incoming data (either intentional or due to temporary sensor failure), 
there will be no kinematic state prediction. Unless the LO system is started while already moving at high speed,
this does not present a particular challenge to the ICP-like optimizer. However, our framework offers 
the option to define an alternative ICP pipeline for time steps without state prediction, e.g. to use a much larger matching threshold.
Experimental results presented in this paper did not need to exploit such possibility, though.

\begin{figure*}
\centering
\includegraphics[width=1\textwidth]{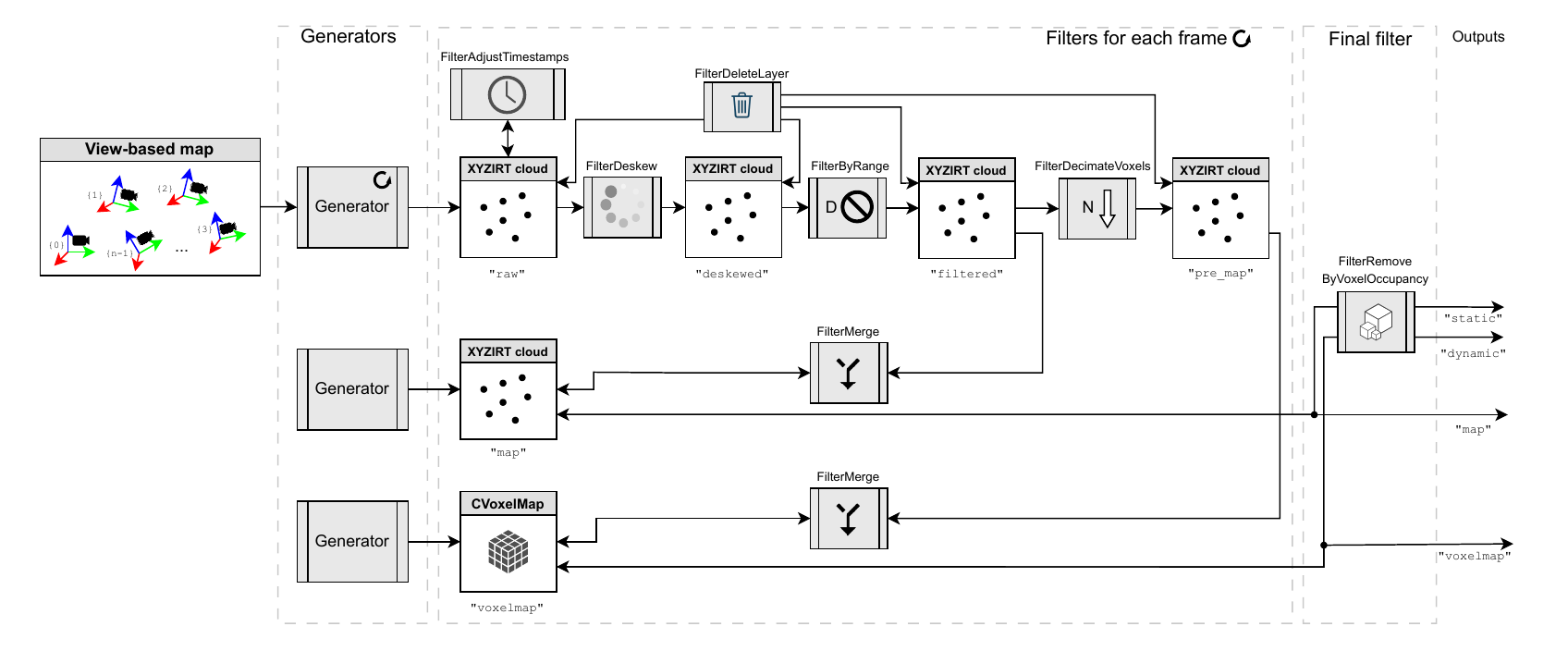}
\caption{Example configuration for building metric maps from view-based maps (using the application \texttt{sm2mm}): 
a more advanced pipeline for classification of points into static or dynamic via volumetric probabilistic occupancy. 
See output examples in \Figure{fig:sm2mm.example.paris} and discussion in Section~\ref{sect:sm2mm}.
}
\label{fig:metric.map.pipeline.example.b}
\end{figure*}

\subsection{\REVIEW{Implementation of view-based maps}}
\label{sect:simple.maps}
This section provides some insights into how view-based maps (see Section~\ref{sect:view.based.maps})
are implemented in our framework.
As stated above, these maps
are the output of LO, hence they must include all the required information
to build actual metric maps from them in the most flexible way.
Thus, each such map $\mathfrak{m}$ is defined as a sequence of $N$ key-frames $\mathbf{k}_i$:

\begin{eqnarray}
\label{eq:simple-map.1}
\mathfrak{m} &=& \left\{ \mathbf{k}_i \right\}_{i=1}^N \\
\label{eq:simple-map.2}
\mathbf{k}_i &=& \left( \mathbf{T}_i, (\Vel_i, \AngVel_i), \{ \mathfrak{o}_{i,j} \}_{j=1}^{M_i} \right)
\end{eqnarray}

\noindent with key-frames being tuples of three elements:
(i) the vehicle pose as a Gaussian distribution $\mathbf{T}_i=\left\{ \bar{\mathbf{T}}_i,\boldsymbol{\Sigma}_i \right\}$, 
(ii) the estimated vehicle linear ($\Vel_i$) and angular ($\AngVel_i$) velocities, and
(iii) a set of $M_i$ raw sensor observations $\mathfrak{o}_{i,j}$. 
Vehicle poses and velocities are given in global coordinates with an arbitrary origin, while observations include both, raw sensor data,
and the sensor pose within the vehicle.

Within observations we include an additional metadata observation which, instead of coming from
a real sensor, is filled during LO. It includes information such as the bounding box, in robocentric coordinates,
of all sensed points for each keyframe. This enables efficient determination of unfeasible loop-closures without
having to analyze the raw sensor data (see Section~\ref{sect:lc}).
Furthermore, given that these maps may make use of the lazy-load feature described in Section~\ref{sect:dataset.sources}
to make them efficient to load and parse, this metadata enables loading from disk only those sensor frames that actually
are needed for loop-closure hypothesis checking.

\begin{figure*}
\centering
\subfigure[Layer: ``map'']{\includegraphics[width=0.75\textwidth]{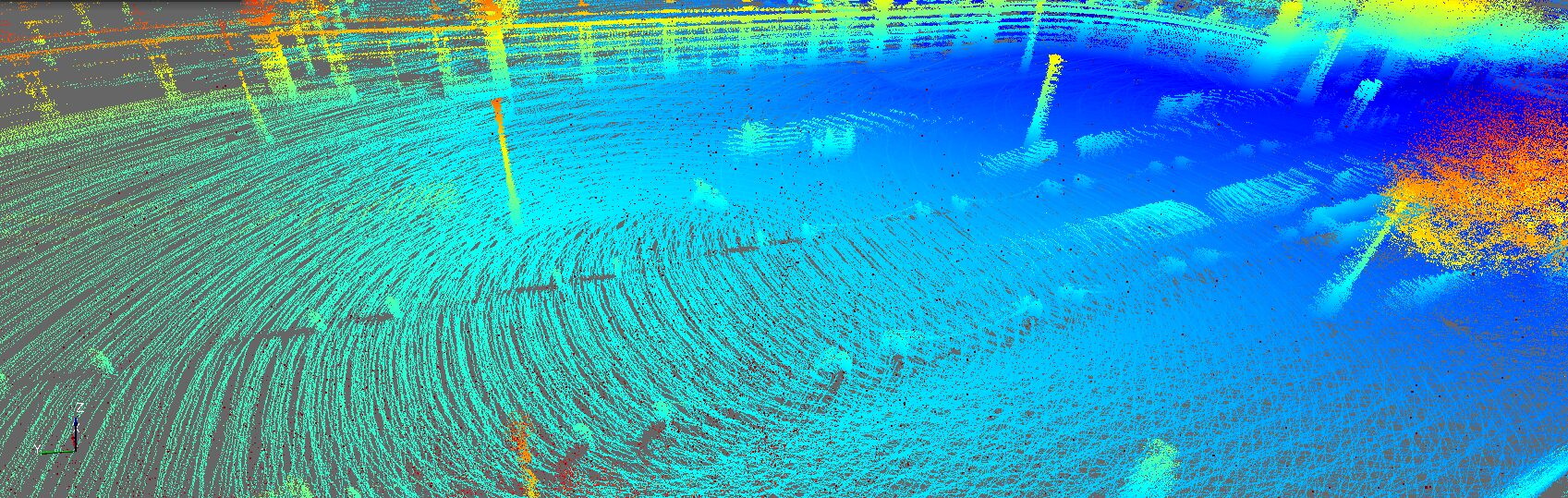}} 
\subfigure[Layer: ``static'']{\includegraphics[width=0.75\textwidth]{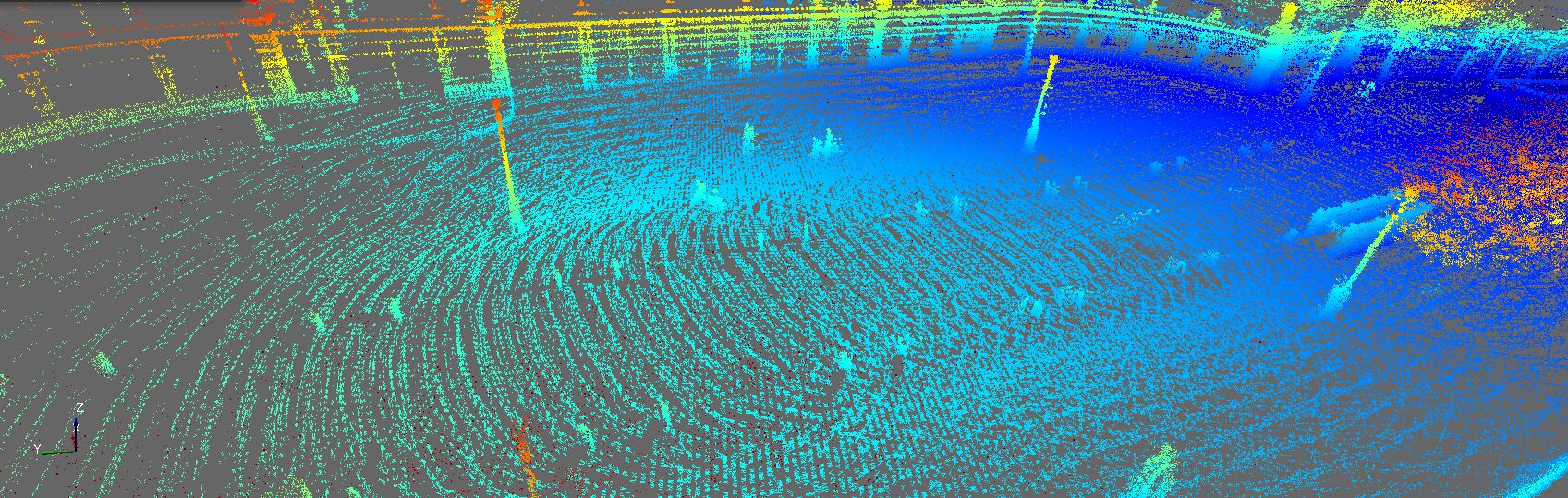}} 
\subfigure[Layer: ``dynamic'']{\includegraphics[width=0.75\textwidth]{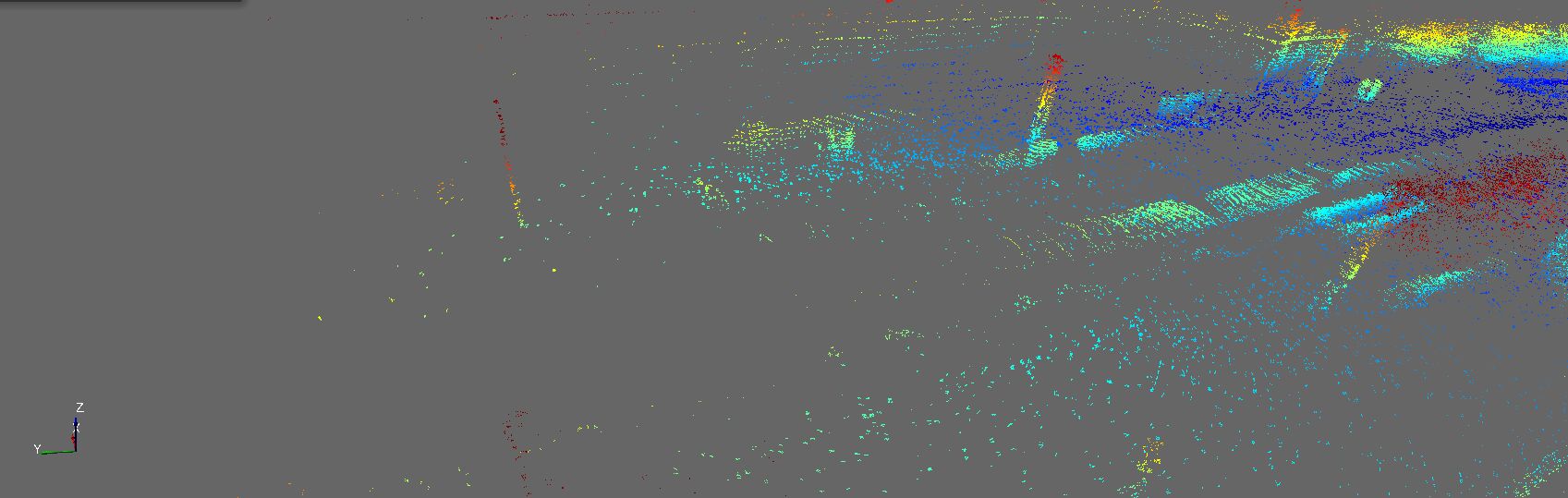}} 
\subfigure[Layer: ``voxelmap'']{\includegraphics[width=0.75\textwidth]{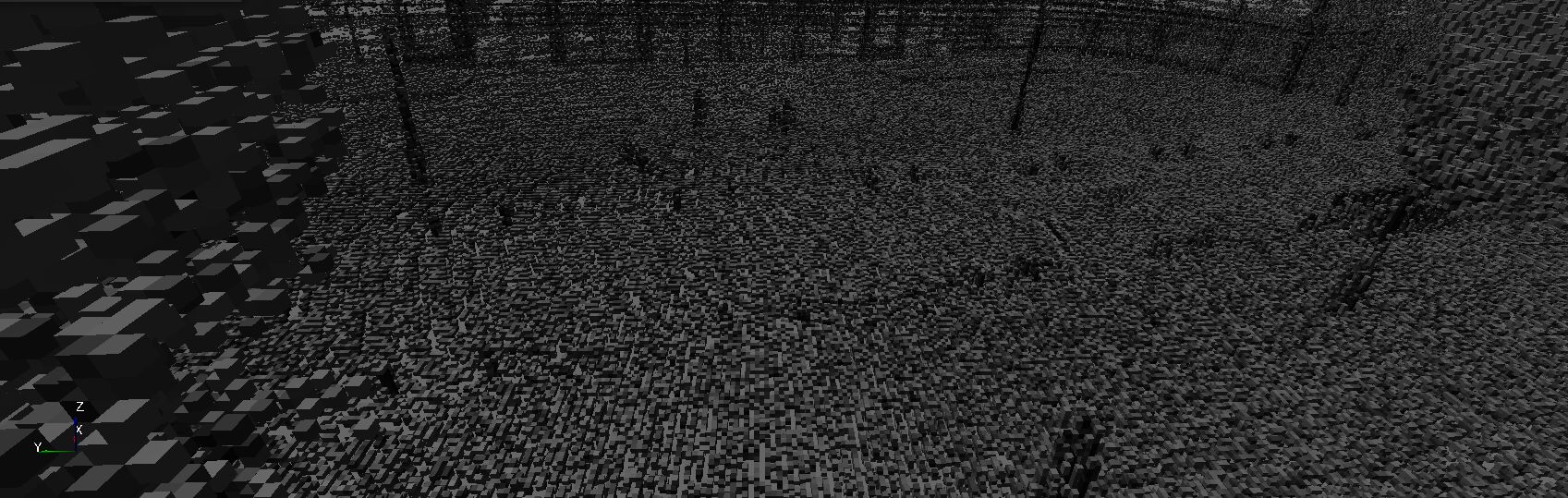}} 
\caption{Layers of the metric map built using the pipeline of \Figure{fig:metric.map.pipeline.example.b}
from the Luxembourg Garden automotive dataset in \cite{dellenbach2022ct}. 
Note how most traces of moving pedestrians and vehicles in (a) have been correctly removed in (b).
See discussion in Section~\ref{sect:sm2mm}.
}
\label{fig:sm2mm.example.paris}
\end{figure*}

\begin{figure*}
\centering
\subfigure[Layer: ``map'' (original map)]{
	\includegraphics[width=0.75\textwidth]{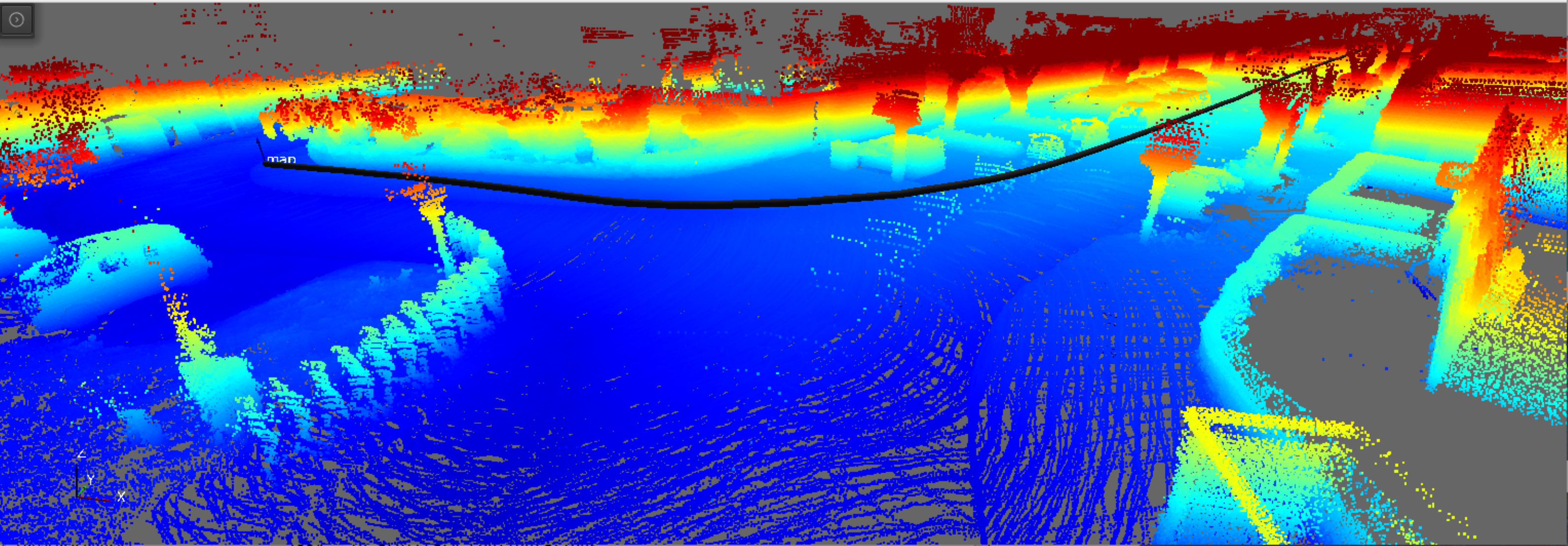}
    \begin{tikzpicture}[overlay, remember picture]
        
        \draw[white, thick] (-10,+2) ellipse (1.15cm and 1.5cm);
        \node[align=center, text=white] at (-8, +1.15) {Pedestrian};

        \draw[white, thick] (-5.5,+3) ellipse (1.5cm and 1.5cm);
        \node[align=center, text=white] at (-5.4, +1.15) {Vehicle};
    \end{tikzpicture}
} 
\subfigure[Layer: ``static'']{\includegraphics[width=0.75\textwidth]{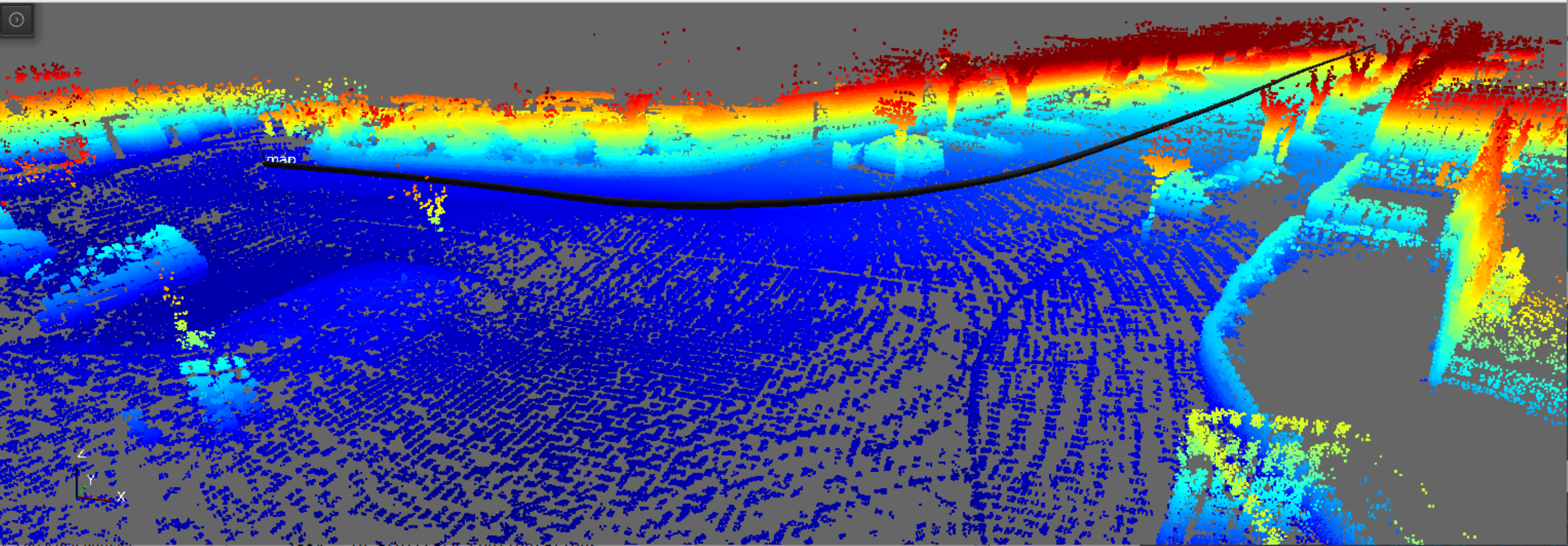}} 
\subfigure[Layer: ``dynamic'']{\includegraphics[width=0.75\textwidth]{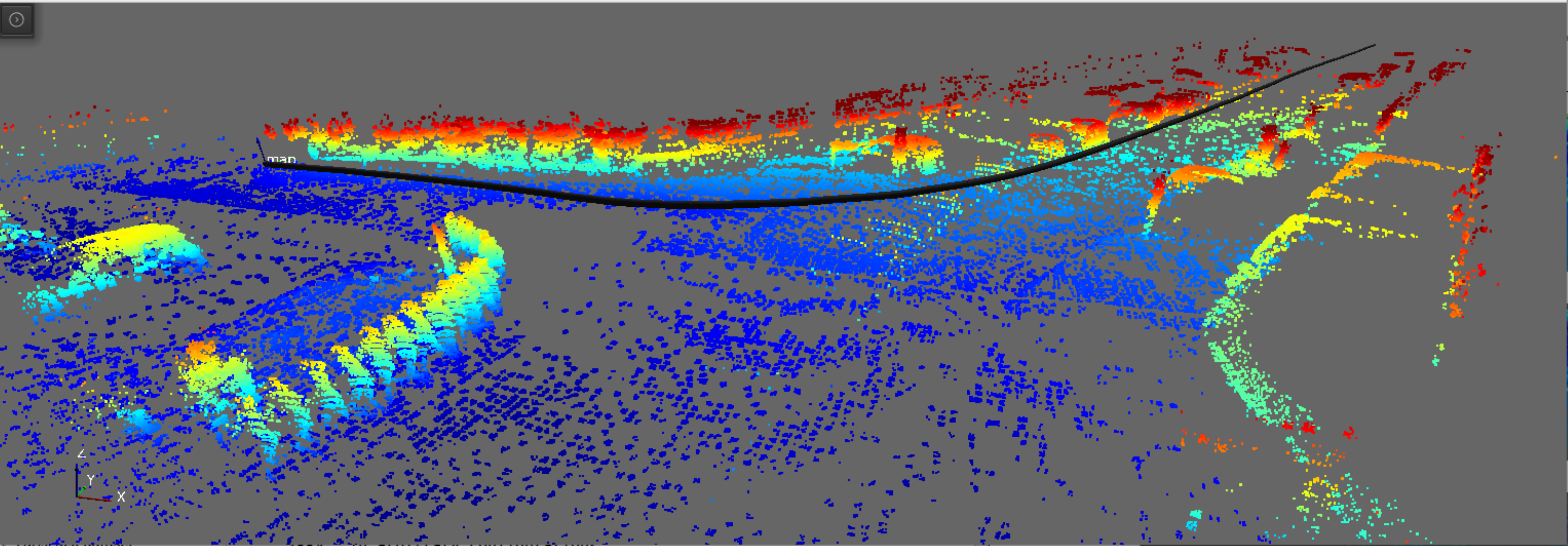}} 
\caption{\REVIEW{Layers of the metric map built using the pipeline of \Figure{fig:metric.map.pipeline.example.b}
from the sequence \texttt{00\_4390\_to\_4530} in the ERASOR dataset \citep{Lim2021erasor}. The estimated vehicle trajectory 
is shown as a thick black solid line. 
Note how traces of moving pedestrians and vehicles in (a) have been correctly removed in (b). The ``dynamic'' layer in (c) contains 
those moving objects, together with noisy points from environment voxels
that were not consistently observed as occupied with a level of confidence high enough.
See discussion in Section~\ref{sect:sm2mm}.}
}
\label{fig:sm2mm.example.erasor}
\end{figure*}

\subsection{\REVIEW{Converting view-based map to metric maps}}
\label{sect:sm2mm}

\REVIEW{Once a view-based map is populated with key-frames, 
one can use it to build a particular kind of metric map from it;}
this corresponds to the task 
represented in \Figure{fig:overview}(c) while discussing the framework overview.

Pipelines comprise three stages:
\begin{itemize}
\item \textbf{Generators:} Can be evaluated once to create empty metric maps of a given type with a particular set of parameters,
or evaluated for each incoming frame in the view-based map to populate a map layer from its raw sensory data.

\item \textbf{Per-frame filters:} The main part, where all filters in a sequence are evaluated in a predefined order,
once per incoming map frame.

\item \textbf{Final filters:} This optional stage can be used to post-process the maps, further filtering them, etc.
\end{itemize}

The utility of such flexible design is illustrated with two example metric map pipelines\footnote{Their
corresponding YAML file descriptions can be checked out online under \texttt{demos} in the \texttt{mp2p\_icp} project.}. 
Possibly the simplest pipeline for 3D LiDAR datasets, sketched in \Figure{fig:metric.map.pipeline.example.a}, 
uses a generator converting raw observations into a ``raw'' point cloud, which is 
then de-skewed after adjusting per-point timestamps to a given fixed convention. Recall de-skew in this 
context is possible since estimated velocities are stored within \REVIEW{map key-frames} (see Section~\ref{sect:simple.maps}).
Next, scan points are down-sampled and spatially filtered to remove observations from the vehicle body.
Finally, all layers are cleared so they start empty for the next map frame, except the map layer used as output, 
where all points accumulate.

A more elaborate pipeline is presented in \Figure{fig:metric.map.pipeline.example.b}, with the goal of 
generating point clouds where mobile objects are removed, which is accomplished as follows.
On the left, we have three generators: one to convert raw sensory data into point clouds (just as in the former pipeline),
plus other two generators, each only invoked once to create two of the final metric maps. Note that each generator
will create metric map layers of different types and with a particular sets of parameters (e.g. resolution, log-odds occupancy update weights, etc.).
These two maps are a point cloud (layer ``map'') and a 3D occupancy voxel map (Section~\ref{sect:metric.maps}).
Raw points are de-skewed and spatially filtered to remove self-body interferences. That version of the cloud is inserted, for each frame, into the 
final point cloud ``map'', whereas its down-sampled version (for efficiency reasons) is used to 
perform ray-tracing and update the occupancy value of the voxel map.
Once all map frames are processed, the final filtering stage makes use of the filter described in 
Section~\ref{sect:filter.dyn.obj.removal} to split the map point cloud into those likely to be static and dynamic.
\REVIEW{As a first example of the attainable results}, 
\Figure{fig:sm2mm.example.paris} shows the output map layers
for a small fragment of the Paris LuCo dataset \citep{dellenbach2022ct},
where it can be seen how most 3D points corresponding to pedestrians, vehicles, or noisy points 
(e.g. near the edges of light poles),
are removed from the static layer. 
Note that since there is no filter implementing learning-based semantic segmentation yet,
vehicles and pedestrians that remain still are classified as static objects.

\REVIEW{As an additional demonstration of this dynamic object removal pipeline we have tested it against one of the sequences
of the ERASOR dataset introduced in \cite{Lim2021erasor}. This dataset is specifically designed to capture situations with dynamic objects such as pedestrians or vehicles in motion.
The same parameters were used than in the Paris LuCo dataset above, and results are shown in Figure~\ref{fig:sm2mm.example.erasor}. 
It can be verified how all traces of vehicles and pedestrians are correctly removed in the ``static'' layer.
}

To sum up, the flexibility of our framework enables generators or filters to be combined
in arbitrary configurations
to easily manipulate and filter metric maps in ways that have not been possible before, 
and all this without coding.

\subsection{Georeferencing view-based maps}
\label{sect:sm.georef}

\begin{figure}
\centering
\subfigure[Coordinate frames involved in georeferencing]{\includegraphics[width=0.85\columnwidth]{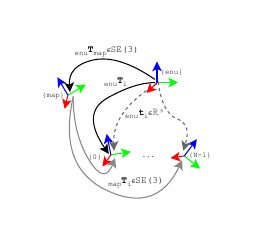}}
\subfigure[Georeferencing as a factor graph]{\includegraphics[width=0.85\columnwidth]{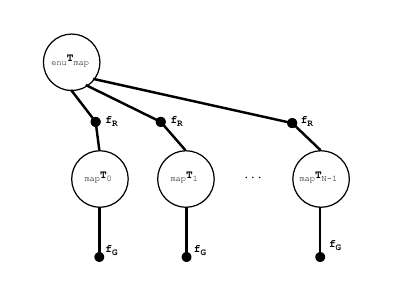}}
\caption{The problem of georeferencing a simple-map involves
(a) finding the unknown relative pose between the ENU and metric map frames of reference, 
which can be turned up into optimizing the factor graph in (b). See discussion in Section~\ref{sect:sm.georef}.
}
\label{fig:georef}
\end{figure}

Georeferencing a map consists in establishing the correspondence between 
coordinates in the map frame and either geodetic coordinates 
or any other global coordinate system such as UTM (Universal Transverse Mercator).

In particular, we address the problem of georeferencing view-based maps $\mathfrak{m}$
(see \Eq{eq:simple-map.1}--\ref{eq:simple-map.2}), or segments of such maps, 
that contain frames with observations $\mathfrak{o}_{i,j}$ from low-cost GNSS receivers with positioning errors in the range of meters.
Apart of its utility on itself, georeferencing segments of large maps becomes a valuable tool
for loop-closure detection outdoors (see Section~\ref{sect:lc}).

Following \Eq{eq:simple-map.2}, we assume without loss of generality that there are $N$ map frames
with global poses with respect to the map origin being $\Pose{i}\equiv \Pose{i}{map} \in SE(3)$ 
and where all frames contain exactly one GNSS positioning observation $\mathfrak{o}_{i}$
with geodetic coordinates, i.e. latitude, longitude, and ellipsoid height.
Let the first such observation, $\mathfrak{o}_{0}$, be used to determine the East-North-Up (ENU) frame of reference
with respect to which all other observations have Euclidean coordinates $\Trans{i}{enu} \in \mathbb{R}^3$; see \cite[\S4.1]{blanco2009collection} for
the geodetic transformation formulas.

Therefore, note that once we have run LO and have GNSS observations, two coordinate origins coexist: 
the LO metric map origin (``map'') and that for ENU coordinates (``enu'').
Georeferencing the map hence becomes finding out the rigid transformation $\Pose{map}{enu} \in SE(3)$ that best
explains all observations, as sketched in \Figure{fig:georef}(a).
The placement of the ENU frame may be initialized from the very first GNSS observation in the view-based map,
or may be given already by another past observation in the context of sub-mapping (see LC in Section~\ref{sect:lc}).
We propose using \REVIEW{factor graphs \citep{Ni2007ICRA,murphy2012machine}}, a popular kind of bipartite graphical model,
as the ideal tool to solve such estimation problem from noisy, possibly containing spurious, GNSS readings.
As shown in \Figure{fig:georef}(b), the graph unknowns are the relative pose $\Pose{map}{enu}$
and the vehicle trajectory poses $\Pose{i}{enu}$, in the ENU frame.
The relative pose between the latter ones will normally not change significantly, 
since they were estimated using LO with a small uncertainty, but their values are left as free variables such as, for long trajectories,
global GNSS observations are able to correct the LO drift, up to the GNSS receiver accuracy.
\REVIEW{In particular, GNSS observations are specially beneficial to 
reduce the larger error that exists in the vertical axis due to LO drift
in long trajectories.}
Two types of factors are defined. First, 
unary GNSS factors ($f_G$), whose error vector $\mathbf{e}_G(\Pose{i}{enu})$ is the mismatch 
between the measured GNSS ENU coordinates $\Trans{i}{enu}$ and the predicted position of the GNSS antenna:

\begin{equation}
\mathbf{e}_G = ( \Pose{i}{enu} \oplus \Trans{g}{v} ) - \Trans{i}{enu}
\end{equation}

\noindent with $\Trans{g}{v} \in \mathbb{R}^3$ the GNSS antenna coordinates on the vehicle (``v'').
These factors use: (i) robust kernels to cope with large errors in areas with satellite signal occlusions, 
and (ii) a zero-mean Gaussian noise model based on the GNSS self-reported accuracy.
The other type of factor in \Figure{fig:georef}(b) is the relative pose binary factor ($f_R$)
whose error function $\mathbf{e}_R(\Pose{map}{enu}, \Pose{i}{enu})$ is proportional to the pose
mismatch between key-frame pose variables $\Pose{i}{enu}$ and their pose as stored in the simple-map:

\begin{equation}
\mathbf{e}_R =
\log \left( 
(\Pose{i}{map})^{-1} \left(\Pose{map}{enu}\right)^{-1} \Pose{i}{enu}  
\right)^{\vee}
\end{equation}

\noindent which follows from pose compositions through the two possible paths
between the frame ``map'' and the $i$-th key-frame in \Figure{fig:georef}(a).

We use GTSAM (\cite{dellaert2021factor})
to implement and solve the optimization problem defined by this factor graph. 
The quality of the resulting georeferencing is assessed via 
the square roots of the diagonal of the marginal covariance matrix for $\Pose{map}{enu}$, 
which reflect whether all six degrees of freedom in SE(3) were properly
constrained by observations or not, e.g. a common caveat is the ``roll'' angle
not be estimated accurately when processing a map segment with an almost perfect straight line trajectory.
Adding multiple GNSS sensors with a large baseline could be interesting for large vehicles.
Experimental results regarding georeferencing maps are shown in Section~\ref{sect:results.geo}, while this feature is also a building
block of LC, addressed next.

\subsection{Loop closure algorithm}
\label{sect:lc}
Having loop closing (LC) mechanisms is what makes the difference between an odometry 
and a SLAM system, capable of creating large, globally-consistent maps.
LC in our framework is not run concurrently to odometry 
but as a post-processing stage. This could change in future versions,
but the fundamentals would remain the same.

\begin{algorithm}[t]
\caption{Overview of the loop closure algorithm}
\label{algo:lc}
\scriptsize
\begin{algorithmic}[1]
\Function{loop\_closure}{$\mathfrak{m}$}
	\LineComment{$\mathfrak{m}$ is the input/output simple-map}
	
    \State $S = \{s_i\}_{i=1}^M \gets \textsc{DivideSubmaps}(\mathfrak{m})$ \Comment{Split into sub-maps}

    \ForEach{$s_i \in S$} \Comment{For each sub-map} 
		\State $b_i \gets \textsc{BoundingBox}(s_i)$ \Comment{Evaluate bounding box}
		\State $g_i \gets \textsc{GeoReferencing}(s_i)$ \Comment{Geo-reference sub-map, if possible}
    \EndFor
    
	\State $G_s \gets \textsc{BuildSubmapGraph}(S, \{g_i\})$
	\State $G_k \gets \textsc{BuildKeyFramesGraph}(S, \{g_i\})$

    \If{$\texttt{any } \{g_i\}$} \Comment{Optimize poses from GNSS}
    	\State $\{G_k,G_s,\mathfrak{m}\} \gets \textsc{OptimizeGraph}(G_k,G_s,\mathfrak{m})$
    \EndIf

	\While{\texttt{true}}
	
	\State{$L \gets \textsc{LC\_Candidates}(G_s,\{b_i\})$}
	
    \If{$|L| = 0$}
		\State{\textbf{break}}  \Comment{No more potential loop closures}	
	\EndIf

    \ForEach{$L_i \in L$} \Comment{For each loop-closure candidate} 
		\State{$\{G_k,G_s,\mathfrak{m}\} \gets \textsc{Process\_LC}(L_i,G_k,G_s,\mathfrak{m})$}
	\EndFor
	
   	\State $\{G_k,G_s,\mathfrak{m}\} \gets \textsc{OptimizeGraph}(G_k,G_s,\mathfrak{m})$
	
	\EndWhile

    \State \textbf{return} $\mathfrak{m}$ \Comment{Return map with optimized key-frame poses}
\EndFunction
\end{algorithmic}
\end{algorithm}

\begin{figure*}
\centering
\includegraphics[width=0.9\textwidth]{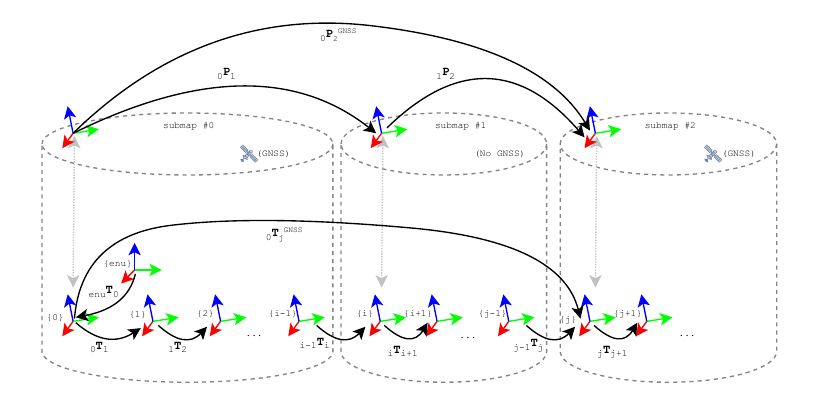}
\caption{The two-level hierarchical graph used to determine loop closures 
and optimize the resulting key-frames. 
On the bottom, all map key-frames; on a higher level, one single frame per sub-map.
Some sub-maps may include GNSS observations, allowing the determination of their coordinates
in an ENU frame of reference. 
Refer to discussion in Section~\ref{sect:lc}.
}
\label{fig:lc.submaps}
\end{figure*}

In the following, please check Algorithm~\ref{algo:lc} to follow the discussion on 
the top-level structure of the LC procedure.
First of all, the input and output of LC are both the view-based map
$\mathfrak{m}$ (see definition in Section~\ref{sect:simple.maps}),
comprising key-frames with SE(3) poses with respect to some arbitrary origin
and raw sensory data (i.e. GNSS observations, LiDAR scans).
The goal of LC is to detect when key-frames correspond to observations of the same 
place, even if their global coordinates from LO are far apart, and to correct all
poses so the trajectory, and consequently the associated metric maps, are globally consistent.
The problem is attacked from a hierarchical view-point by first splitting 
the input map into sub-maps (line 3 in the pseudocode),
for which their local metric map bounding box (line 5)
and geo-referenced location from GNSS data (line 6, see Section~\ref{sect:sm.georef})
are evaluated. Note that GNSS is totally optional, but it will indeed help constraining the posterior search for LC candidates
\REVIEW{by reducing the uncertainty of relative poses between key-frames that are topologically far apart}.
Then, we build a two-level graph structure: 
on the bottom-level, a key-frame graph ($G_k$) has one node per map key-frame, 
while on a higher level, a sub-map graph ($G_s$) has one node per sub-map, 
as sketched in \Figure{fig:lc.submaps}.
\REVIEW{This approach follows a tradition of splitting large metric maps
into submaps, such as done with Tectonic SAM in \cite{Ni2007ICRA}
or with condensed factors in \cite{Grisetti2012IROS}.
Two relevant differences of our approach with respect to those past works are: 
(i) the explicit distinction
between condensed factors coming from LiDAR odometry and those
from GNSS observations, 
and (ii) the usage of each of the two graphs for different purposes.
In particular,} the low-level graph
is tightly coupled with a corresponding factor-graph in charge of
actually optimizing the global poses for the entire map, 
while the higher-level sub-map graph is used to efficiently
search for LC candidates.
Initially, the low-level (key-frames) graph has a linear connection pattern, that is, 
each node $i$ is connected via a pose constraint ($\Pose{i}{i-1}$) to the immediately previous node $i-1$.
Exactly the same happens on the sub-map (higher) level, where we use the notation ${}_{i-1}\mathbf{P}_{i}$
to denote relative poses between sub-maps. These two graphs are built in lines 8--9 of  Algorithm~\ref{algo:lc}.
Then, if there is more than one sub-map where georeferencing was successful, we can exploit that information
by means of additional connections between the reference key-frames of the first ever sub-map with GNSS data and
all others; those additional edges are labeled as ${\Pose{a}{b}}^{GNSS}$ and ${{}_{b}\mathbf{P}_{a}}^{GNSS}$
in \Figure{fig:lc.submaps}
for the lower and higher levels, respectively.
If there are such additional GNSS constraints, the key-frame graph is optimized using a factor graph 
(lines 10--12 of the pseudocode) to reduce absolute coordinate errors from those accumulated by pure LO (unbounded), 
to those of the GNSS receivers (maybe several meters, but bounded).
It is worth mentioning that optimization is run in two passes: 
a first one without robust kernels, then a second robust optimization.
\REVIEW{Note that this is the opposite of what is 
typically done with SLAM frontends, where robust optimization 
comes first to reject outliers.
However, if we want to handle loop closures with a large
initial mismatch between the global poses of the two involved submaps
(e.g. thanks to GNSS data, if available)
robust kernels cannot be applied at first, or the error would 
be so large that it will be ignored as if it was an outlier.}

Then, lines 13--22 comprise the loop to search for potential LC candidates, 
evaluate and apply them if found plausible, re-optimize all poses, and repeat
until no further feasible candidates are found.
A LC candidate is a pair of sub-maps that may correspond to the same physical place, 
at least, partially. 
\REVIEW{Potential candidates are determined by evaluating
the volumetric intersection ratio between the bounding boxes of every pair of sub-maps and 
accepting those above a given threshold (line 14).
Given the potential large uncertainty in the relative poses of two submaps, 
metrically close to each other after a long topological loop,
we use a Monte Carlo method to draw relative pose samples to evaluate the 
mathematical expectation of the intersection ratio.}
To find out the probabilistic relative pose between all sub-maps, 
we apply Dijkstra's algorithm to the higher-level graph
to obtain, for each sub-map, a tree reflecting the shortest paths to all other sub-maps. 
Relative poses are then computed by composing transformations following the tree edges.
Note that adding GNSS edges makes that sub-maps that initially may be far from each other
(large uncertainty in their relative pose)
now are topologically closer, hence their relative pose based on GNSS will be used instead
of the pose composition chain throughout the lower key-frame level.
This graph-based procedure is similar to many past works, such as 
\cite{bosse2004simultaneous} or \cite{blanco2008toward}.
\REVIEW{Note that an alternative way to determine potential 
loop closure candidates would be using descriptors for local maps, 
as classically done in visual place recognition or, 
in the context of LiDAR SLAM, in methods such as Scan Context \citep{Kim2018IROS} or ScanContext++ \citep{gskim-2021-tro}.}

Each such potential LC candidate is then actually evaluated (line 19).
Our approach is based on running a configurable ICP pipeline on the
metric maps built from the view-based maps of the sub-maps, including
the final evaluation of the registration quality (see Section~\ref{sect:icp.pipelines.quality})
to decide whether the LC seems plausible or not.
If it does, new edges are added to both graph levels with the obtained relative pose.

A few words are in order regarding this ICP pipeline, which is different than 
that in \Figure{fig:lidar3d.obs.pipeline} for LO.
The first difference is the use of two point-cloud map layers: one for all objects near the ground plane, 
and another for everything else. The ICP pipeline
then includes two instances of point-to-point matchers
such that ``ground'' and ``non-ground'' points are only attempted to be paired against 
points belonging to the same category in the other sub-maps. 
\REVIEW{Even a naive method to classify the point cloud into these 
two categories such as split by $z$ (vertical) coordinates leads to
good results, as long as local maps do not have significant tilt and
the ground floor can be successfully segmented. 
More advanced methods have been used in the literature 
to segment the ground points in the context of loop closure, 
such as in Quatro++ \citep{Lim2024IJRR},
and could be used instead of our naive classifier.
}

Regarding ICP quality assessment, here we include the voxel occupancy-based method described
in Section~\ref{sect:icp.pipelines.quality}.
Furthermore, the point-cloud metric maps used for LC have a different implementation 
than that used during LO: instead of the hashed-voxels maps used for LO, we use simple clouds
with contiguous memory layout. The main motivation for this is that search for nearest neighbors (NN)
(one of the dominating costs in ICP algorithms)
with large search radii is more efficient using KD-Trees 
(the implementation of NN for the latter map type) than searching neighboring voxels in 
hashed maps.

\begin{table*}[t]
\caption{Summary of all datasets on which the proposed system has been experimentally validated.}
\resizebox{\textwidth}{!}{
\setlength\extrarowheight{1pt}
\begin{tabular}{|c||c|c|c|c|c|c|r|r|}
\hline
  \textbf{Dataset} & 
  \TwoRowsBf{LiDAR}{sensor(s)} & 
  \TwoRowsBf{LiDAR}{rings} & 
  \REVIEW{\textbf{GNSS}} & 
  \textbf{Motion} & 
  \textbf{Environment} & 
  \TwoRowsBf{Tested}{sequences} &
  \TwoRowsBf{Total~~~~}{scans~~~} &
  \TwoRowsBf{Total~~~~}{length~~~}
\\
\hline

\TwoRows{Newer College Dataset $\rightarrow$ Section~\ref{sect:ncd} }{\cite{zhang2021multi}} & OS0-128 & 128 & - & Handheld & College & 12 & 78.2k & 9.8~km
\\
\hline

\TwoRows{Newer College Dataset $\rightarrow$ Section~\ref{sect:ncd} }{\cite{ramezani2020newer}} & OS1-64 &  & - & Handheld & College & 2 & 41.9k & 4.7~km
\\
\hhline{|-|-|~|*{6}{-|}}

\TwoRows{MulRan $\rightarrow$ Section~\ref{sect:mulran} }{\cite{gskim-2020-mulran}}
 & OS1-64 & ~ & \checkmark & Vehicle & Campus/City & 12 & 127.7k & 123.4 km
\\
\hhline{|-|-|~|*{6}{-|}}

\TwoRows{KITTI $\rightarrow$ Section~\ref{sect:kitti}}{\cite{geiger2013vision}} & HDL-64E & 64 & - & Vehicle & Road/Urban & 22 & 43.5k & 44.24~km
\\
\hhline{|-|-|~|*{6}{-|}}

\TwoRows{Voxgraph $\rightarrow$ Section~\ref{sect:voxgraph} }{\cite{reijgwart2019voxgraph}} & OS1-64 & ~ & - & Drone & Urban & 1 & 2.2k & 0.24~km
\\
\hhline{|-|-|~|*{6}{-|}}

\TwoRows{HILTI 2021 $\rightarrow$ Section~\ref{sect:hilti} }{\cite{hilti2021}} & OS0-64 & ~  & - & Drone & \TwoRows{Industrial}{unit} & 1 & 0.9k & 0.08~km
\\
\hhline{|-|-|~|*{6}{-|}}

\TwoRows{KITTI-360 $\rightarrow$ Section~\ref{sect:kitti360} }{\cite{Liao2022PAMI}} & HDL-64E & ~  & - & Vehicle & Urban & 12 & 70.1k &  58.7 km
\\
\hline

\TwoRows{ParisLuco $\rightarrow$ Section~\ref{sect:paris} }{\cite{dellenbach2022ct}} & HDL32 & 32 & - & Vehicle & City & 1 & 12.7k & 4.0~km 
\\
\hhline{|-|-|~|*{6}{-|}}

\TwoRows{Almería forests $\rightarrow$ Section~\ref{sect:backpack} }{\cite{aguilar2024lidar}} & OS0-32 & ~ & - & Backpack & Forests & 6 & 19.9k  & 1.6~km
\\
\hline

\TwoRows{NTU-VIRAL $\rightarrow$ Section~\ref{sect:ntu} }{\cite{nguyen2022ntu}} & 2 $\times$ OS1-16 &  & - &  Drone & Campus & 9 & 32.6k & 2.2~km
\\
\hhline{|-|-|~|*{6}{-|}}

\TwoRows{DARPA Subterranean $\rightarrow$ Section~\ref{sect:darpa}}{\cite{tranzatto2022team}} & VLP16 & 16 &- &  ANYmal C & Caves & 4 & 143.1k & 2.0~km
\\
\hhline{|-|-|~|*{6}{-|}}

\TwoRows{UAL campus $\rightarrow$ Section~\ref{sect:ual.campus} }{\cite{blanco2019benchmarking}} & VLP16 &  & \checkmark & Vehicle & Campus & 1 & 15.5k & 3.4~km
\\
\hline

\TwoRows{fr079 $\rightarrow$ Section~\ref{sect:2dlidar} }{\cite{Radish}} & SICK LRF & 1 & - & PowerBot & Offices & 1 & 1.2k & 0.1~km
\\
\hhline{|-|-|~|*{6}{-|}}

\TwoRows{Málaga CS faculty
$\rightarrow$ Section~\ref{sect:2dlidar} 
}{\cite{blanco2024uma2d}} & SICK LRF &  & - & Wheelchair & Campus & 1 & 4.7k & 1.9~km
\\
\hline
\multicolumn{5}{c}{} &
\multicolumn{1}{r}{\textbf{Overall:}} & 
\multicolumn{1}{c}{\textbf{85}} &  %
\multicolumn{1}{r}{\textbf{594.2k}} &  %
	\multicolumn{1}{r}{\textbf{256.4~km}} %
\end{tabular}
}
\label{tab:summary.datasets}
\end{table*}

\subsection{Dataset sources}
\label{sect:dataset.sources}

\REVIEW{Finally, it is worth mentioning that} the former work in \cite{blanco2019mola} already introduced
data source MOLA modules as an abstraction of either, real live
sensors or offline datasets.
Such abstraction aims at enabling algorithm implementations in MOLA to immediately work with any of the supported datasets, without worrying about the details 
of each dataset file and data structures, sensor intrinsic and extrinsic calibration,
ground truth (GT) axes conventions, etc. 

This work extends the former modules in two substantial ways.
Firstly, new dataset sources have been defined, now offering access to:
\begin{itemize}
\item \textbf{KITTI dataset}:
Read LiDAR scans (with per-point intensity field), camera images, and ground truth of this popular dataset introduced in \cite{geiger2013vision}.

\item \textbf{KITTI-360 dataset}:
Read LiDAR scans (with intensity field and interpolated timestamps), camera images, and ground truth of the dataset, introduced in \cite{Liao2022PAMI}.

\item \textbf{MulRan dataset}:
Read LiDAR scans (with intensity and timestamp fields), GNSS, IMU, and ground truth of \cite{gskim-2020-mulran}.

\item \textbf{ParisLuco dataset}:
Read LiDAR scans (with timestamps, and reconstructed per-point ring field), and ground truth of \cite{dellenbach2022ct}.

\item \textbf{EuROC MAV dataset}:
Read camera images and IMU for \cite{burri2016euroc}.
This dataset is not benchmarked in the present work since it does not contain LiDAR.

\item \textbf{ROS 2}:
Support is given for reading from both, a live system, or from \emph{bag} files.
Sensor messages for LiDAR point clouds, camera images, IMU readings, wheels odometry, and GNSS are all
parsed for MOLA modules to process them.

\item \textbf{MRPT rawlog files}:
Rawlog files are an equivalent to ROS bag files, introduced in the MRPT framework
in 2005. It has been used in public datasets, e.g. \cite{blanco2014malaga,blanco2019benchmarking}.
Backwards compatibility and Operative System independence has been ensured during the whole life
of the project, hence software written today using the latest MRPT version
is able to read all past datasets, even if generated in a different
OS or processor architecture.
There exists an import tool from ROS~1 bag files to rawlogs, hence this MOLA input module 
also offers easy access to
many of the ROS~1 public datasets, e.g. HILTI challenge datasets \citep{hilti2021} or 
the Newer College dataset \citep{zhang2021multi}.
\end{itemize}

Secondly, two new abstract interfaces have been implemented in all the dataset sources above:

\begin{itemize}
\item \textbf{Replay control interface:}
If used to replay datasets within an asynchronous network
of processing nodes, this interface allows dynamically
setting the replay speed, pausing and resuming, or fast-forwarding
to a particular time step. 
The application GUI makes use of this generic interface
to allow users to control the replay of any dataset.

\item \textbf{Random-access interface:} For offline, batch processing,
this interface allows randomly accessing to the raw sensory observations
in the dataset. This enables, for example, running a given SLAM
algorithm on a dataset as fast as possible while ensuring no frame is lost
for SLAM methods running slower than the sensor rate.
\end{itemize}

Finally, a word is in order regarding memory management for dataset input modules.
Loading and keeping in memory whole datasets may become unfeasible 
for a typical desktop computer, e.g. some individual sequences 
of the MulRan dataset \citep{gskim-2020-mulran} take 29~GiB.
The present framework exploits the usage of MRPT lazy-load
mechanism in different areas, including dataset sources.
Observations that tend to be heavy in memory requirements 
(i.e. images, 3D LiDAR scans, and RGB+D frames)
can be automatically loaded to memory when accessed,
to be unloaded once a fixed number of additional frames
have been processed.
Most MOLA dataset input modules also feature 
read-ahead in a multi-thread fashion to minimize the performance 
impact of such lazy-load mechanism.
In practice, this creates a cache of recently-used and 
about-to-be-used
sensory data quickly accessible in RAM while SLAM is processing
them.

\section{LiDAR odometry quantitative experimental validation}
\label{sect:results.quant}

To measure the accuracy and robustness of the proposed system as a whole
against different LiDAR scanning and vehicle motion patterns,
we performed an exhaustive benchmarking against several
public datasets that include 3D LiDAR and ground truth
trajectories.

Please, refer to Table~\ref{tab:summary.datasets} for an overview of
all the used datasets.
Note that we selected an extensive list of public datasets in order to 
cover a large variety of motion models (vehicles, drones, hand-held, etc.)
and sensor models and resolutions.
In particular, it is noteworthy that we used exactly the same system 
configuration \REVIEW{(called ``default configuration'' in the following)}
for all 3D LiDAR datasets, from 16 to 128 beams. In all cases,
our system provided accurate trajectory estimations without diverging, while
keeping execution times fast enough for running significantly faster than the real sensor rate.
\REVIEW{That said, these additional configurations have been also evaluated for selected datasets
with the aim of serving for ablation studies in Section~\ref{sect:ablation}, to demonstrate a particular feature, or 
to illustrate the flexibility of the proposed system:}
\begin{itemize}
\item \REVIEW{A 2D-LiDAR configuration, addressed in Section~\ref{sect:2dlidar}.}
\item \REVIEW{A configuration where the local map update decider (see Section~\ref{sect:lidar.odometry}) is not used 
and the map is updated for all time steps.}
\item \REVIEW{A configuration using an alternative 3D-NDT metric map instead of plain point clouds.}
\end{itemize}

In order to evaluate our solution against existing state of the art LO methods,
the next subsections provide these quantitative error metrics:

\begin{itemize}
\item Relative translational error (RTE) (\%) and relative rotational error (d/hm, that is, degrees per 100~m): 
these popular metrics were defined in \cite{geiger2013vision}, and measure relative pose errors, 
averaging for poses that are a given distance apart. These metrics have been evaluated using a derived work
from the original reference metrics implementation provided by \cite{geiger2013vision}, which
assumed that LiDAR scans and ground-truth are perfectly synchronized and given at the same rate. 
This is not possible for all dataset sources, hence these metrics will not be available for them.

\item Absolute Trajectory Error (ATE) (m): We use the RMSE value of the distances between time-aligned
ground-truth and estimated trajectories, after spatially aligning them using Umeyama'a method \citep{umeyama1991least}.
Evaluation is performed using the open-sourced tool \texttt{evo\_traj} \citep{grupp2017evo}.
\end{itemize}

Unless noted otherwise, metrics for alternative methods have been taken from their
original publications. The exception is KISS-ICP \citep{vizzo2023kiss}, which due to its
simplicity of integration into third-party systems, has been integrated as
another MOLA LO module, hence enabling running all datasets on it and obtaining trajectories
in the same format than our proposed system, becoming a excellent baseline SOTA reference
for all the datasets.

Sometimes a particular LO method \emph{diverges}, which is represented in the presented metrics
as $\times$. Divergence means a significant drift, large enough to render the overall 
trajectory (and map) useless. In particular, our criteria for divergence 
means that at least one of the following conditions hold: 
(i) trajectory gets ``trapped'' into a circular orbit, 
(ii) relative angular errors larger than $45^\circ$ for short distances (e.g. less than 10 meters).

\begin{figure*}
\centering
\begin{tabular}[t]{cc}
\subfigure[\texttt{KAIST01} (6.1~km long)]{\hspace*{-0.4cm}\includegraphics[width=0.48\textwidth]{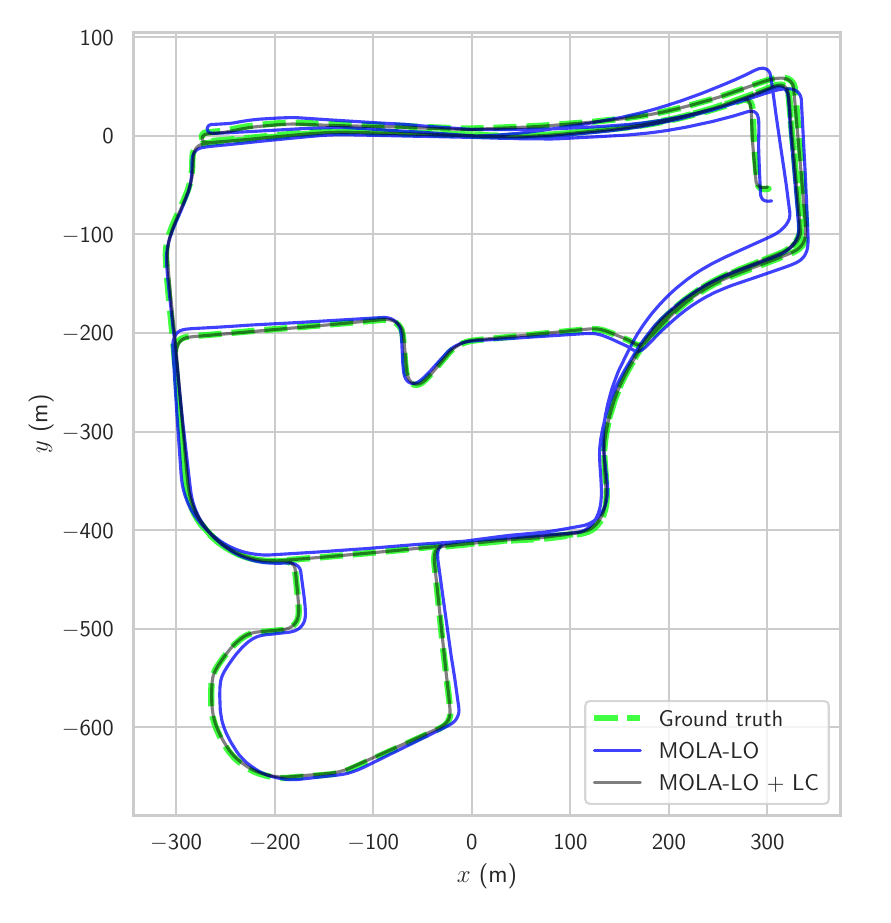}}
&
\raisebox{4.5cm}{
\begin{tabular}{l}
\subfigure[\texttt{Riverside01} (6.4~km)]{\hspace*{-1.2cm} \includegraphics[width=0.54\textwidth]{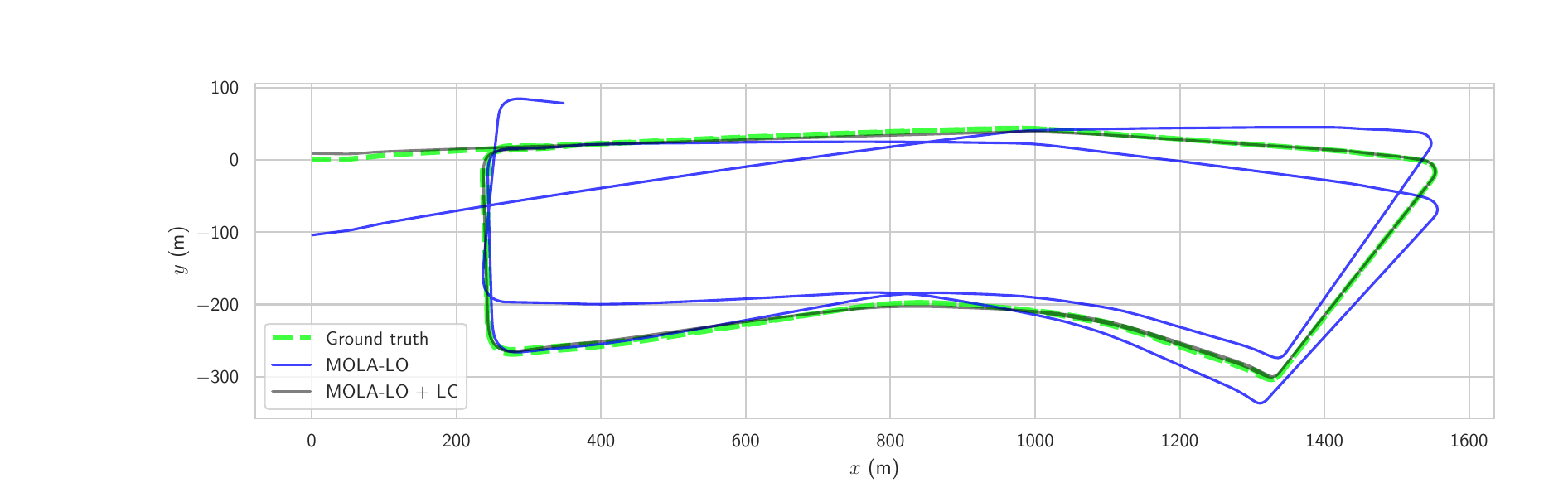}}
\\
\subfigure[\texttt{DCC01} (4.9~km)]{\hspace*{-1.cm}\includegraphics[width=0.24\textwidth]{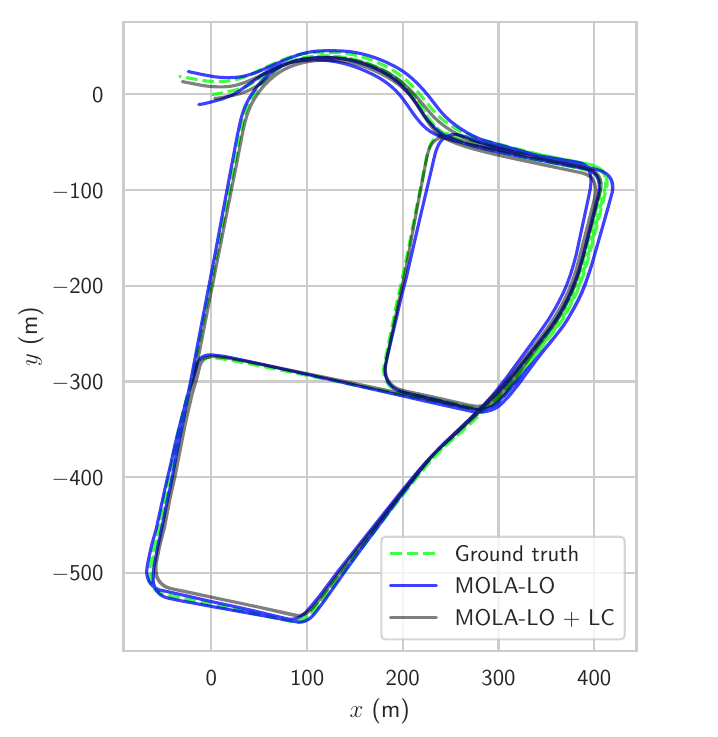}}
\subfigure[\texttt{Sejong01} (\textbf{23.2~km})]{\hspace*{-0.4cm}\includegraphics[width=0.30\textwidth]{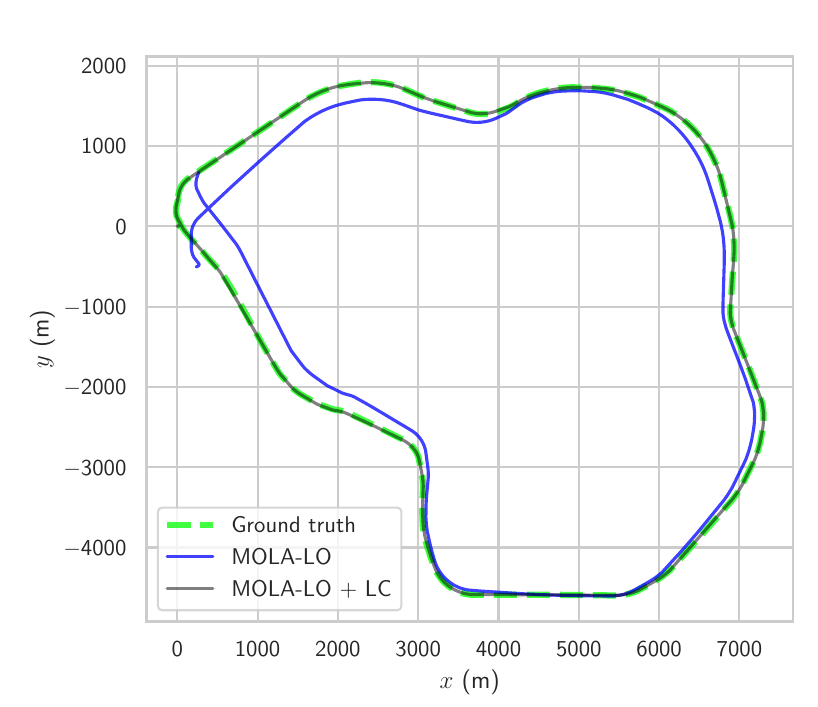}}
\end{tabular}
}
\end{tabular}
\caption{Estimated trajectories for the proposed LiDAR odometry system \REVIEW{(in its default configuration)} applied to the \textbf{MulRan dataset}, compared to ground truth.
Only one sequence is shown for each of the four locations. Estimated paths including loop closure 
(section~\ref{sect:lc}) are denoted with legend ``MOLA-LO+LC''.}
\label{fig:mulran.paths}
\end{figure*}

\begin{table*}
\caption{Relative translational error (RTE) (\%), relative rotational error (RRE) (d/hm, that is, degrees per 100~m) (as defined in \cite{geiger2013vision}), and absolute translational error (ATE) (rmse values in meters as reported by ``\texttt{evo\_ape -a}'')
for the \textbf{MulRan dataset}.
All sequences have loop closures. Smaller numbers are better. Bold means best accuracy in each sequence and category.}
\centering
\resizebox{\textwidth}{!}
{
\setlength\extrarowheight{1pt}
\begin{tabular}{|c||r|r|r|r|r|r|r|r|r|r|r|r|r|}
\hline
\textbf{Method}
& \multicolumn{3}{c|}{\texttt{KAIST}}
& \multicolumn{3}{c|}{\texttt{DCC}}
& \multicolumn{3}{c|}{\texttt{Riverside}}
& \multicolumn{3}{c|}{\texttt{Sejong}}
& Avr. time
\\
\hhline{|~|*{12}{-|}~|}
~
& \texttt{01} & \texttt{02} & \texttt{03} 
& \texttt{01} & \texttt{02} & \texttt{03} 
& \texttt{01} & \texttt{02} & \texttt{03} 
& \texttt{01} & \texttt{02} & \texttt{03} 
& per frame
\\ %
\hline
\TwoRows{KISS-ICP}{\cite{vizzo2023kiss}} & 
\ThreeRows{2.22\%}{0.66~$d/hm$}{13.36~m} &
\ThreeRows{2.15\%}{0.68~$d/hm$}{11.76~m} &
\ThreeRows{2.43\%}{{0.70}~$d/hm$}{12.32~m} &
\ThreeRows{2.62\%}{0.65~$d/hm$}{12.56~m} &
\ThreeRows{2.32\%}{0.63~$d/hm$}{10.13~m} &
\ThreeRows{1.92\%}{{0.62}~$d/hm$}{10.54~m} &
\ThreeRows{3.33\%}{0.74~$d/hm$}{41.50~m} &
\ThreeRows{3.15\%}{0.68~$d/hm$}{35.96~m} &
\ThreeRows{2.55\%}{{0.53}~$d/hm$}{29.32~m} &
\ThreeRows{4.28\%}{0.62~$d/hm$}{373.53~m} &
\ThreeRows{4.79\%}{0.68~$d/hm$}{417.26~m} &
\ThreeRows{4.93\%}{0.82~$d/hm$}{344.05~m} 
&
23 ms 
\\ %
\hline
\TwoRows{SiMpLE (MulRan)}{\cite{bhandari2024minimal}} & 
\ThreeRows{\textbf{2.03}\%}{\textbf{0.58}~$d/hm$}{13.29~m} &
\ThreeRows{\textbf{1.97}\%}{\textbf{0.60}~$d/hm$}{{9.56}~m} &
\ThreeRows{\textbf{2.22}\%}{\textbf{0.65}~$d/hm$}{{10.10}~m} &
\ThreeRows{\textbf{2.45}\%}{\textbf{0.60}~$d/hm$}{12.17~m} &
\ThreeRows{\textbf{1.99}\%}{\textbf{0.53}~$d/hm$}{{8.04}~m} &
\ThreeRows{\textbf{1.75}\%}{\textbf{0.56}~$d/hm$}{{7.38}~m} &
\ThreeRows{\textbf{3.10}\%}{\textbf{0.61}~$d/hm$}{\textbf{23.90}~m} &
\ThreeRows{\textbf{2.97}\%}{\textbf{0.58}~$d/hm$}{30.63~m} &
\ThreeRows{\textbf{2.40}\%}{0.51~$d/hm$}{50.84~m} &
\ThreeRows{4.15\%}{\textbf{0.47}~$d/hm$}{636.11~m} &
\ThreeRows{5.06\%}{\textbf{0.55}~$d/hm$}{690.61~m} &
\ThreeRows{5.26\%}{\textbf{0.70}~$d/hm$}{566.00~m}
&
32 ms 
\\ %
\hline
\ThreeRowsC{MOLA-LO}{Config.: Default}{(ours)} & 
\ThreeRows{{2.13}\%}{{0.65}~$d/hm$}{{10.23}~m} &
\ThreeRows{2.12\%}{0.70~$d/hm$}{{10.51}~m} &
\ThreeRows{{2.38}\%}{0.71~$d/hm$}{{10.79}~m} &
\ThreeRows{{2.56}\%}{{0.63}~$d/hm$}{{10.90}~m} &
\ThreeRows{2.34\%}{0.62~$d/hm$}{{9.48}~m} &
\ThreeRows{1.88\%}{0.65~$d/hm$}{{8.74}~m} &
\ThreeRows{3.27\%}{0.72~$d/hm$}{{38.05}~m} &
\ThreeRows{3.10\%}{{0.66}~$d/hm$}{{32.89}~m} &
\ThreeRows{2.52\%}{{0.53}~$d/hm$}{30.27~m} &
\ThreeRows{{4.17}\%}{{0.60}~$d/hm$}{429.76~m} &
\ThreeRows{4.67\%}{{0.67}~$d/hm$}{455.47~m} &
\ThreeRows{4.83\%}{{0.81}~$d/hm$}{333.63~m}
&
24 ms 
\\ %
\hline
\ThreeRowsC{\REVIEW{MOLA-LO}}{\REVIEW{Config.: Always update map}}{(ours)} & 
\ThreeRows{2.11\%}{0.65~$d/hm$}{10.58~m} &
\ThreeRows{2.09\%}{0.69~$d/hm$}{9.65~m} &
\ThreeRows{2.33\%}{0.71~$d/hm$}{10.18~m} &
\ThreeRows{2.53\%}{0.63~$d/hm$}{10.88~m} &
\ThreeRows{2.27\%}{0.61~$d/hm$}{9.18~m} &
\ThreeRows{1.87\%}{0.66~$d/hm$}{9.14~m} &
\ThreeRows{3.25\%}{0.71~$d/hm$}{35.13~m} &
\ThreeRows{3.08\%}{0.65~$d/hm$}{31.25~m} &
\ThreeRows{2.50\%}{0.52~$d/hm$}{29.55~m} &
\ThreeRows{4.06\%}{0.58~$d/hm$}{466.81~m} &
\ThreeRows{4.58\%}{0.66~$d/hm$}{512.25~m} &
\ThreeRows{4.85\%}{0.81~$d/hm$}{413.81~m} &
24 ms 
\\ %
\hline
\ThreeRowsC{\REVIEW{MOLA-LO}}{\REVIEW{Config.: 3D-NDT}}{(ours)} & 
\ThreeRows{2.09\%}{0.64~$d/hm$}{\textbf{8.77}~m} &
\ThreeRows{2.05\%}{0.68~$d/hm$}{\textbf{7.44}~m} &
\ThreeRows{2.30\%}{0.71~$d/hm$}{\textbf{7.79}~m} &
\ThreeRows{2.53\%}{0.61~$d/hm$}{\textbf{9.20}~m} &
\ThreeRows{2.18\%}{0.61~$d/hm$}{\textbf{6.74}~m} &
\ThreeRows{1.89\%}{0.65~$d/hm$}{\textbf{6.83}~m} &
\ThreeRows{3.19\%}{0.69~$d/hm$}{24.91~m} &
\ThreeRows{3.08\%}{0.66~$d/hm$}{\textbf{21.51}~m} &
\ThreeRows{2.49\%}{\textbf{0.50}~$d/hm$}{\textbf{20.30}~m} &
\ThreeRows{\textbf{3.91}\%}{0.56~$d/hm$}{\textbf{317.56}~m} &
\ThreeRows{\textbf{4.45}\%}{0.63~$d/hm$}{\textbf{337.94}~m} &
\ThreeRows{\textbf{4.68}\%}{0.78~$d/hm$}{\textbf{199.25}~m} &
39 ms 
\\
\hhline{|*{14}{=|}}
\TwoRows{MOLA-LO + LC (ours)}{(default LO + GNSS)}
 & 
\ThreeRows{\textbf{2.04}\%}{{0.64}~$d/hm$}{\textbf{2.42}~m} &
\ThreeRows{\textbf{1.91}\%}{\textbf{0.64}~$d/hm$}{\textbf{2.13}~m} &
\ThreeRows{\textbf{2.15}\%}{\textbf{0.69}~$d/hm$}{\textbf{1.93}~m} &
\ThreeRows{\textbf{2.34}\%}{\textbf{0.61}~$d/hm$}{\textbf{5.14}~m} &
\ThreeRows{{2.27}\%}{{0.67}~$d/hm$}{{4.21}~m} &
\ThreeRows{\textbf{1.70}\%}{\textbf{0.60}~$d/hm$}{\textbf{1.90}~m} &
\ThreeRows{\textbf{2.66}\%}{\textbf{0.62}~$d/hm$}{\textbf{3.98}~m} &
\ThreeRows{\textbf{2.44}\%}{\textbf{0.57}~$d/hm$}{\textbf{2.92}~m} &
\ThreeRows{\textbf{2.08}\%}{\textbf{0.42}~$d/hm$}{\textbf{3.56}~m} &
\ThreeRows{\textbf{3.20}\%}{\textbf{0.50}~$d/hm$}{\textbf{36.64}~m} &
\ThreeRows{\textbf{3.67}\%}{\textbf{0.57}~$d/hm$}{\textbf{41.90}~m} &
\ThreeRows{\textbf{4.21}\%}{\textbf{0.74}~$d/hm$}{\textbf{26.55}~m}
&
40 ms 
\\
\hline
\TwoRows{\REVIEW{MOLA-LO + LC (ours)}}{\REVIEW{(default LO, no GNSS)}}
 & 
\ThreeRows{2.08\%}{\textbf{0.63}~$d/hm$}{2.57~m} &
\ThreeRows{1.98\%}{0.66~$d/hm$}{2.56~m} &
\ThreeRows{2.34\%}{0.72~$d/hm$}{2.76~m} &
\ThreeRows{2.50\%}{0.65~$d/hm$}{5.31~m} &
\ThreeRows{\textbf{2.05}\%}{\textbf{0.58}~$d/hm$}{\textbf{2.75}~m} &
\ThreeRows{1.76\%}{0.61~$d/hm$}{1.91~m} &
\ThreeRows{3.79\%}{0.87~$d/hm$}{22.76~m} &
\ThreeRows{3.21\%}{0.65~$d/hm$}{9.09~m} &
\ThreeRows{2.52\%}{0.49~$d/hm$}{6.93~m} &
\ThreeRows{4.08\%}{0.57~$d/hm$}{381.54~m} &
\ThreeRows{4.61\%}{0.65~$d/hm$}{419.82~m} &
\ThreeRows{4.80\%}{0.80~$d/hm$}{315.43~m} &
42 ms 
\\
\hline
\end{tabular}
}
\label{tab:mulran.metrics}
\end{table*}

\begin{figure}
\centering
\includegraphics[width=1\columnwidth]{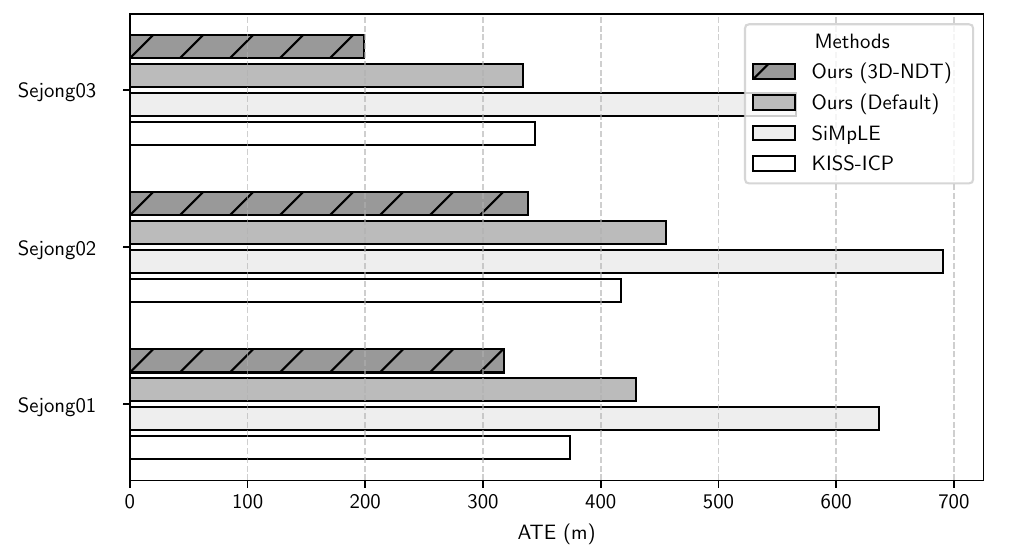}\\
\includegraphics[width=1\columnwidth]{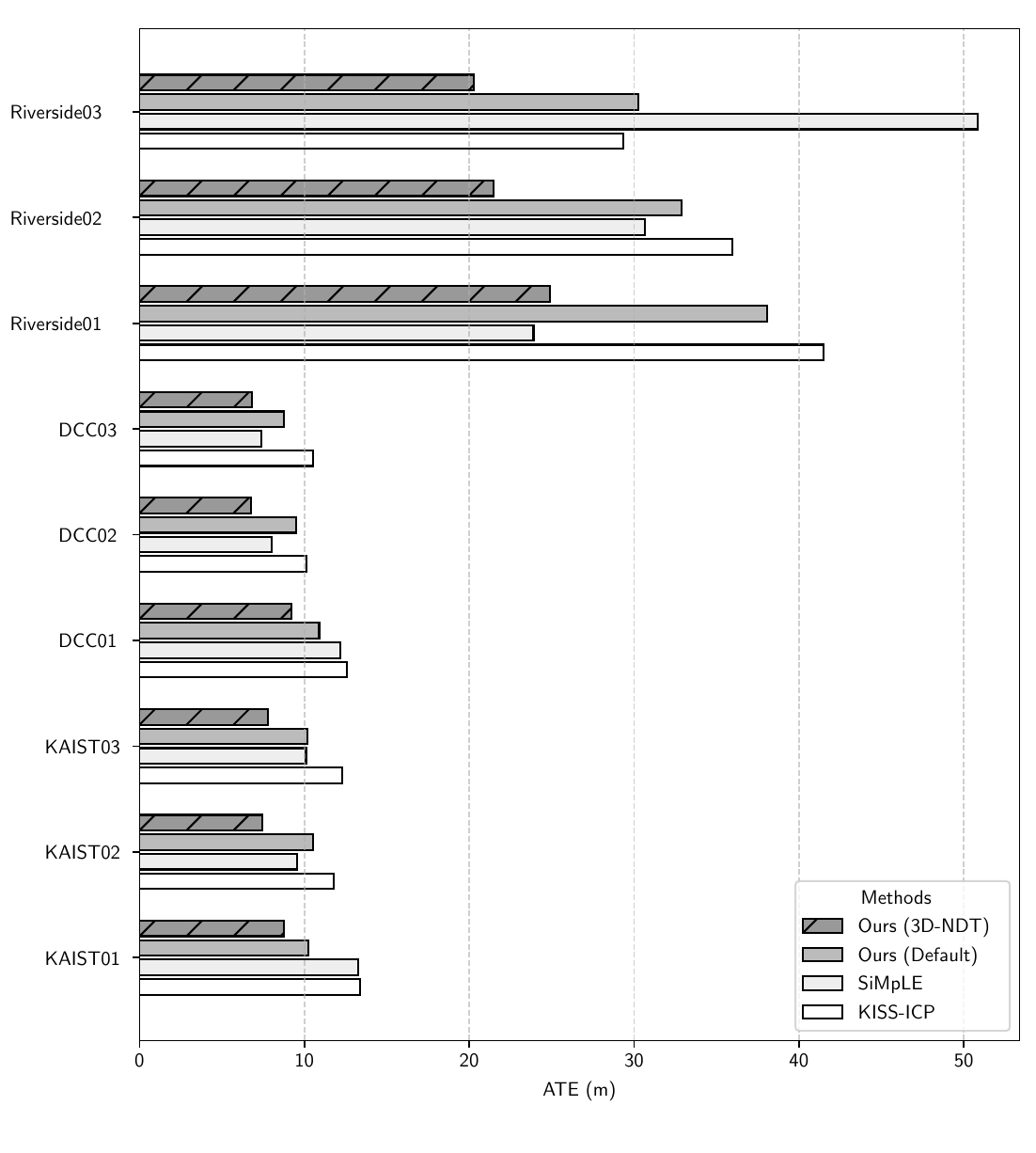}
\caption{\REVIEW{Summary of absolute translational error (ATE) for the \textbf{MulRan dataset}.}}
\label{fig:mulran.summary.ate}
\end{figure}

All results are exactly reproducible by interested readers following the software instructions
on the project website. Since all of our proposed methods and software implementations are deterministic,
there will be no differences in the results disregarding different CPU speed or number of parallel cores.

\begin{figure}
\includegraphics[width=1.0\columnwidth]{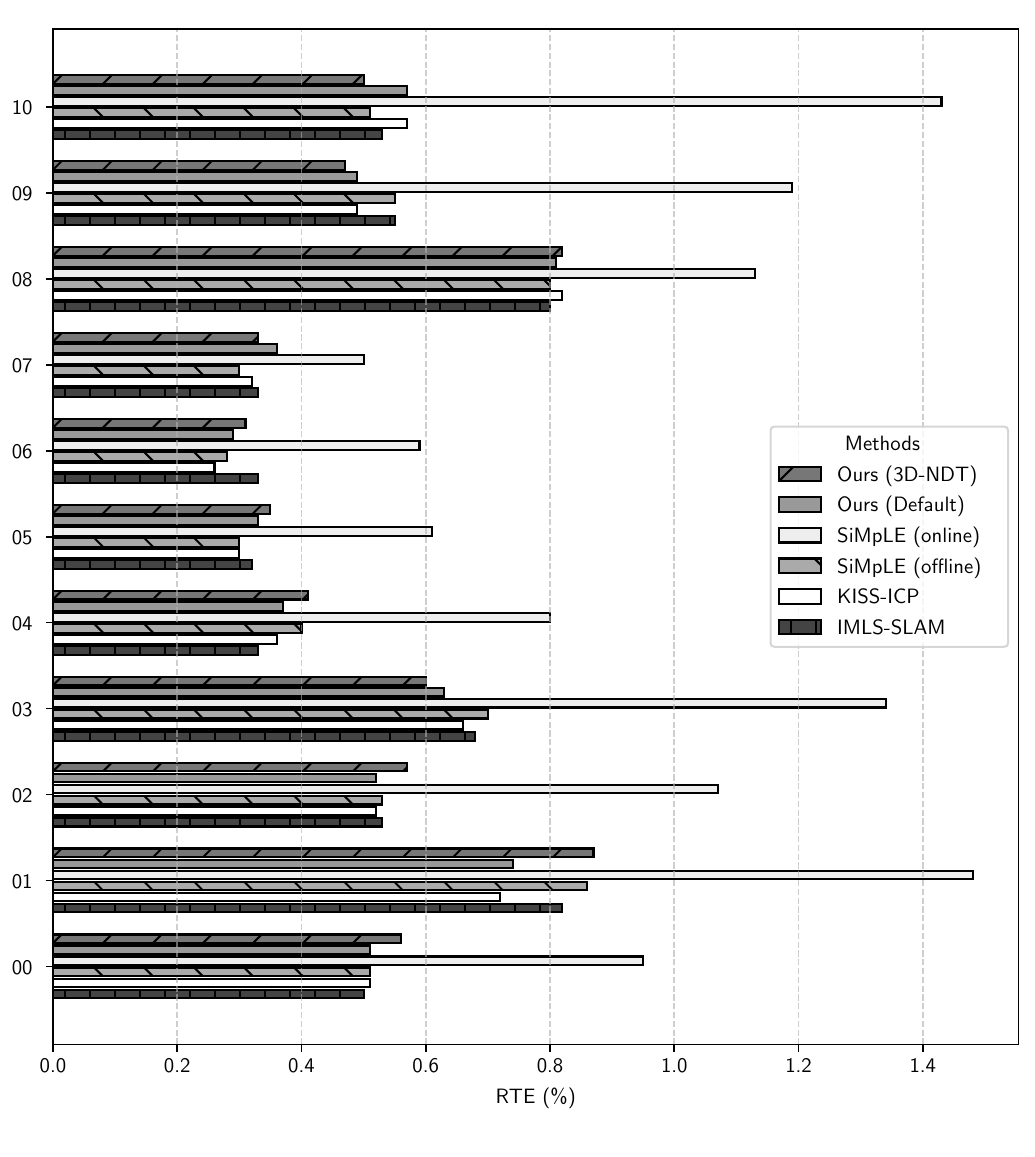}
\caption{\REVIEW{Relative translational error (RTE) (\%) for the training sequences in the \textbf{KITTI visual-LiDAR odometry dataset}}.}
\label{fig:kitti.rte.summary.graph}
\end{figure}

\subsection{MulRan dataset}
\label{sect:mulran}

This multimodal dataset comprises Ouster OS1-64 LiDAR data, consumer-grade GNSS, and precise ground truth
for 12 driving sequences \citep{gskim-2020-mulran}. All sequences include several loop closures, except the
longest ones (\texttt{Sejong\{01,02,03\}}) which only include one, although the longest loop in all 
the benchmarked datasets ($>20$ km).

We have compared our LO system in two configurations, (i) its default one and (ii) its 3D-NDT version, 
against two other state-of-the-art LO methods:
KISS-ICP \citep{vizzo2023kiss} using its default settings, and SiMpLE \citep{bhandari2024minimal} using 
their parameter set optimized for this dataset. 
\REVIEW{Note that only SiMpLE has parameters optimized for this particular dataset, while our
two configurations and KISS-ICP use the same parameters for all datasets}.
MOLA wrappers have been provided for both methods to provide a fair comparison in terms of identical API
to provide input raw data and to collect estimated trajectories.
The MOLA dataset source (see Section~\ref{sect:dataset.sources}) for MulRan takes care of 
removing LiDAR scans at the beginning or end of each sequence that have no associated ground truth. 
In this way, we can use the KITTI evaluation metrics (RTE and RRE) apart of APE (see metrics definitions in Section~\ref{sect:results.quant}).
Consumer-grade GNSS observations are not used by our LO system, but they are while applying the loop-closure post-processing stage (explained in Section~\ref{sect:lc}).

\begin{figure*}
\centering
\resizebox{\textwidth}{!}
{
\begin{tabular}[t]{cccc}
\multicolumn{2}{c}{
\subfigure[\texttt{00}]{
\includegraphics[width=0.49\textwidth]{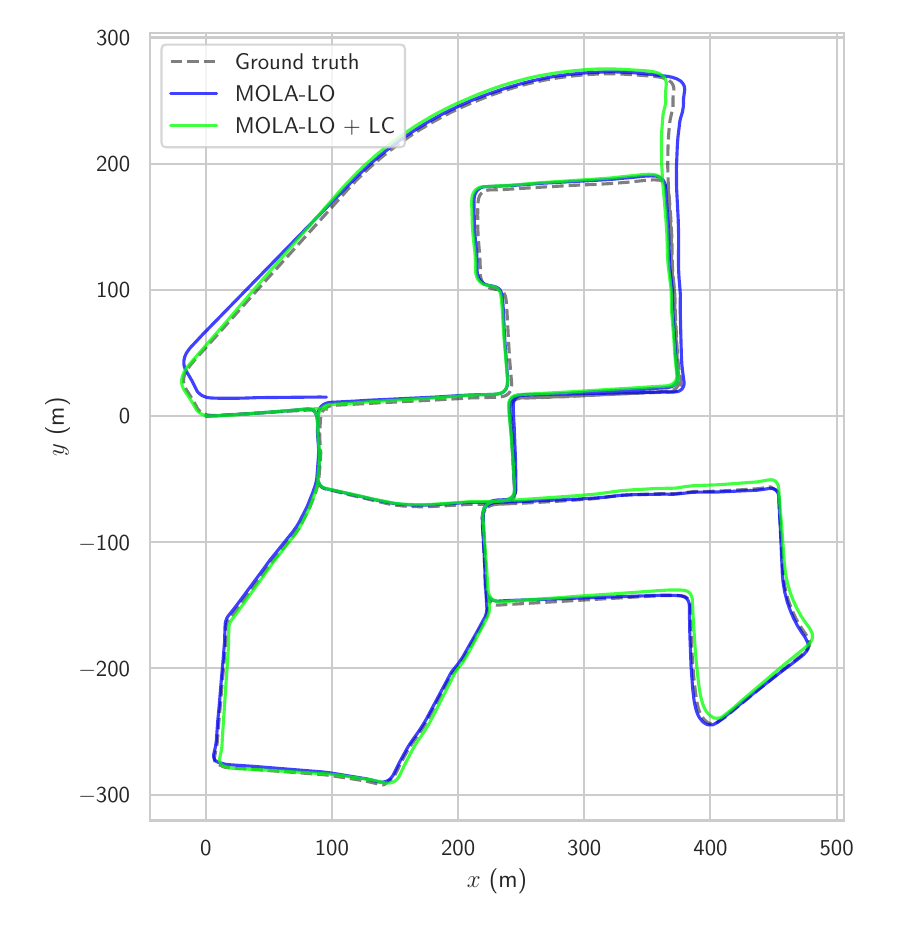}}
}
&
\multicolumn{2}{c}{
\subfigure[\texttt{05}]{\includegraphics[width=0.505\textwidth]{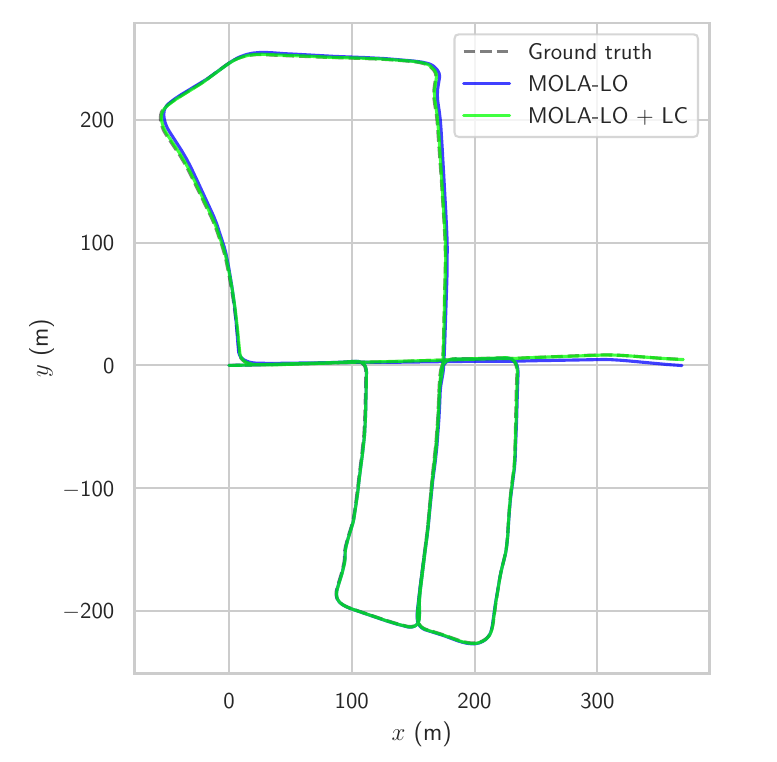}}
}
\\
\subfigure[\texttt{01}]{\includegraphics[width=0.24\textwidth]{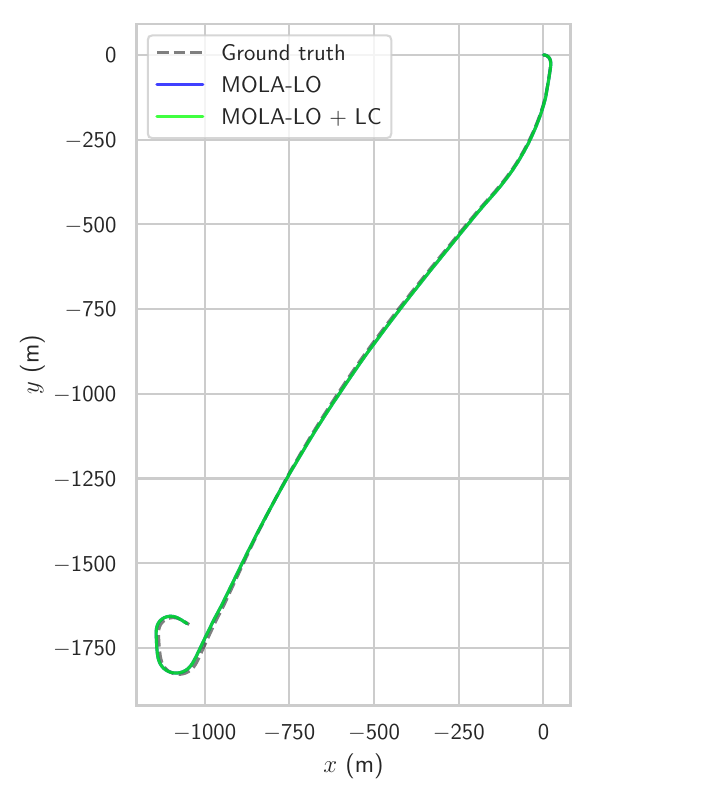}}
&
\subfigure[\texttt{03}]{\includegraphics[width=0.24\textwidth]{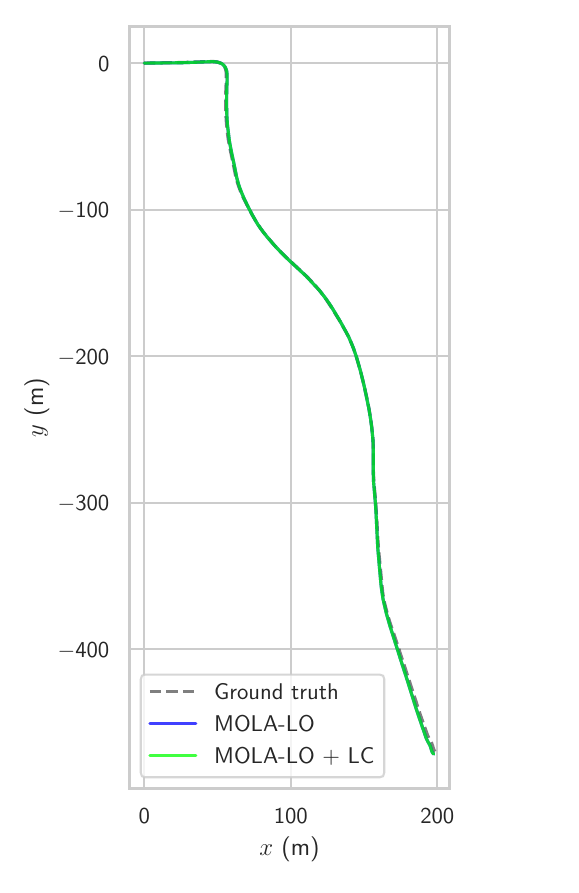}}
&
\subfigure[\texttt{08}]{\includegraphics[width=0.24\textwidth]{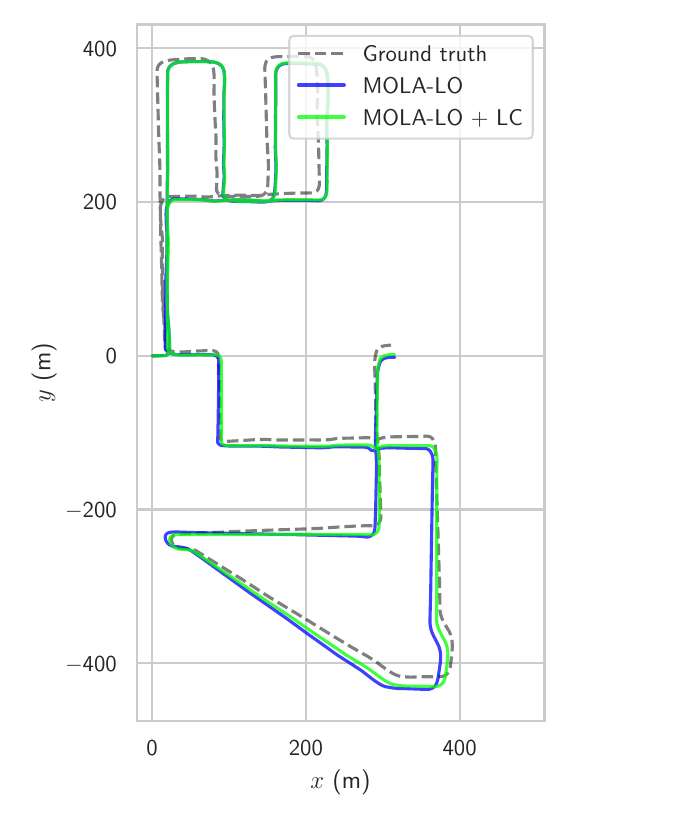}}
&
\subfigure[\texttt{10}]{\includegraphics[width=0.24\textwidth]{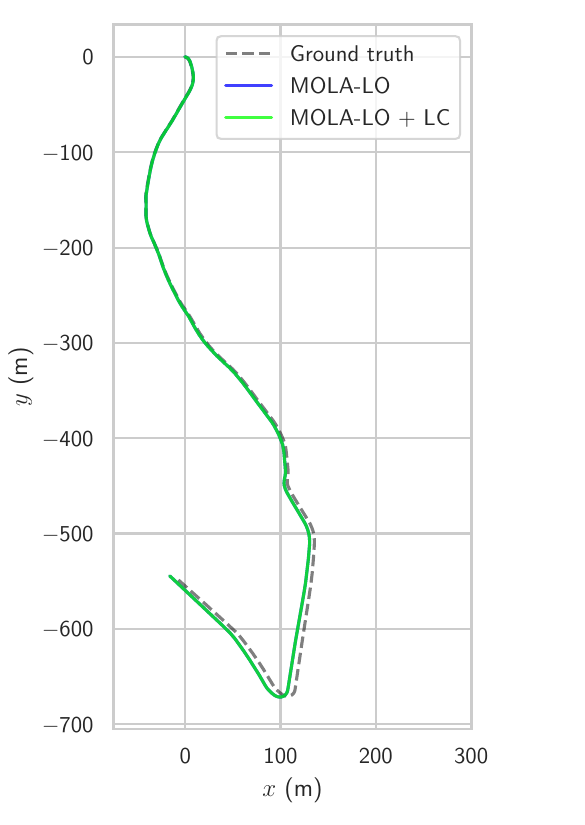}}
\\
\raisebox{1.8cm}{
\begin{tabular}{c}
\subfigure[\texttt{04}]{\includegraphics[width=0.24\textwidth]{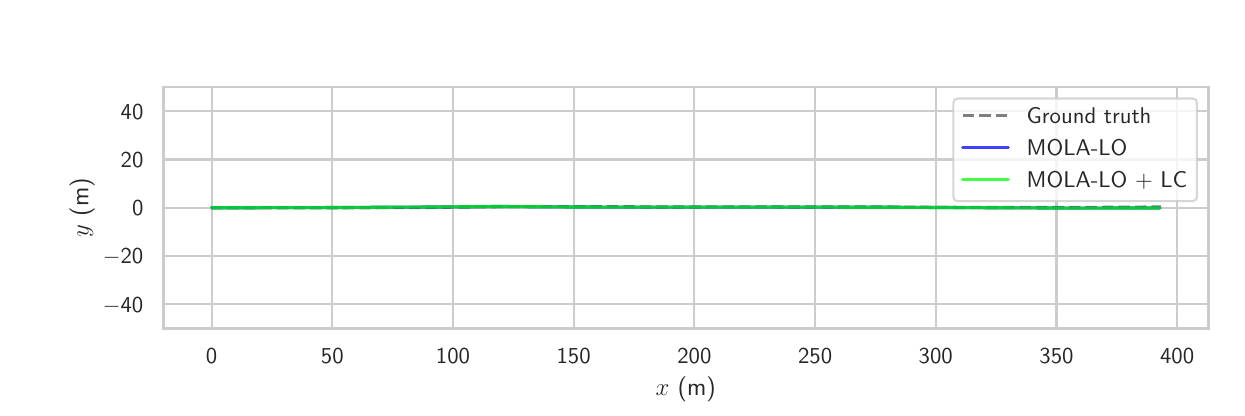}}
\\
\subfigure[\texttt{06}]{\includegraphics[width=0.24\textwidth]{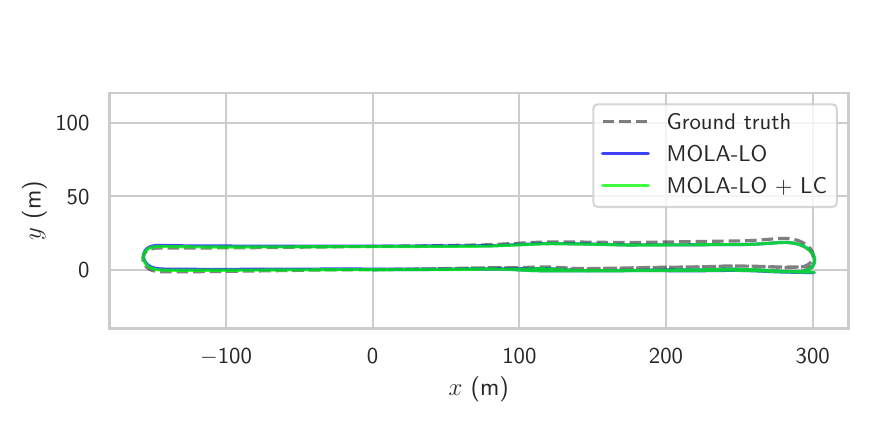}}
\end{tabular}}
&
\subfigure[\texttt{02}]{\includegraphics[width=0.24\textwidth]{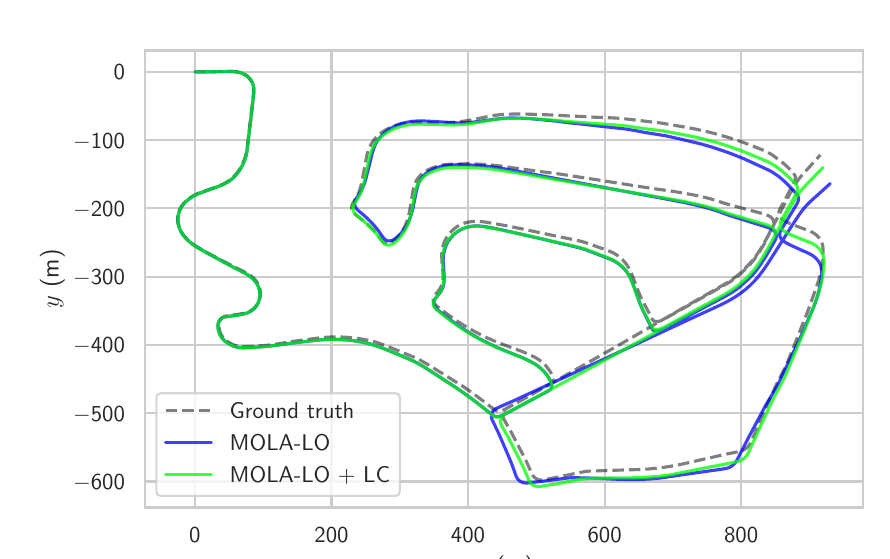}}
&
\subfigure[\texttt{07}]{\includegraphics[width=0.24\textwidth]{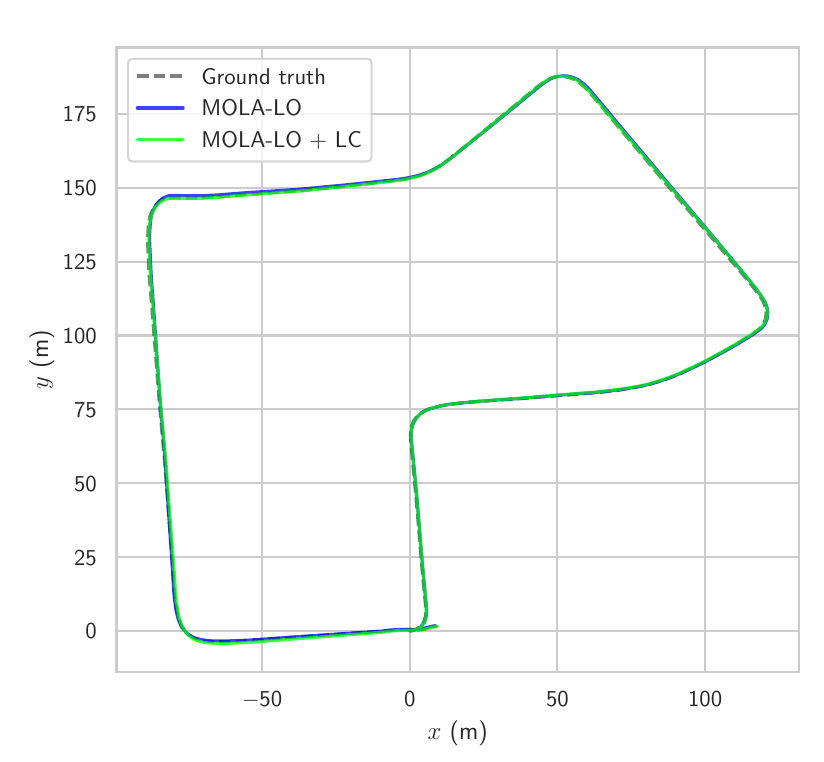}}
&
\subfigure[\texttt{09}]{\includegraphics[width=0.24\textwidth]{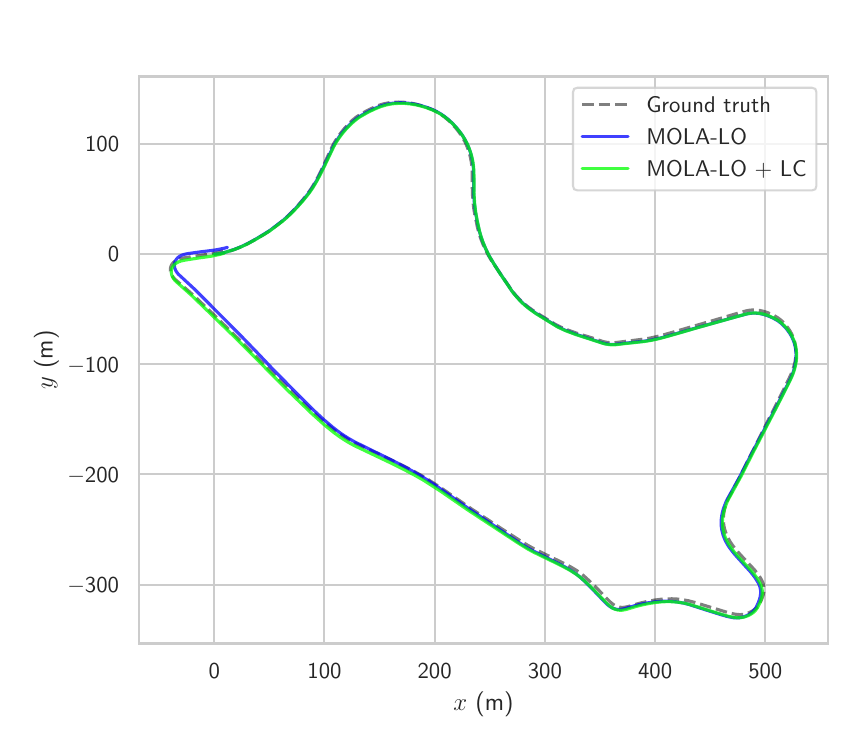}}
\\
\end{tabular}
}
\caption{Estimated trajectories for the proposed LiDAR odometry system ("MOLA-LO") applied to the \textbf{KITTI odometry dataset}, compared to ground truth. Trajectories denoted as ``MOLA-LO + LC'' include loop-closure. See discussion in Section~\ref{sect:kitti}.
}
\label{fig:kitti.paths}
\end{figure*}

Sample trajectories from our method are shown in \Figure{fig:mulran.paths}, compared to GT.
It can be seen that the output from loop closure is nearly indistinguishable from GT.
A quantitative benchmark is provided in Table~\ref{tab:mulran.metrics}, where all three metrics
(RTE, RRE, APE) are provided, together with average run times per LiDAR scan (measured on an Intel i7-8700 at 3.20GHz).
\REVIEW{A graphical summary of ATE metrics is also provided in Figure~\ref{fig:mulran.summary.ate}}.
All LO methods are \REVIEW{faster than} the sensor rate (10 Hz), hence all of them would be able to run onboard in real-time. 
An interesting observation is that there is not always a best performing method 
for a given sequence, since different methods may have the lowest values for one given metric only: 
LO methods would rank in a different order depending on the metric of interest. 
That said, for this dataset we can see \REVIEW{how our 3D-NDT configuration clearly outperforms the others, despite not being 
specifically tuned for this dataset as SiMpLE is}.
Our SLAM version (reflected as ``MOLA LO+LC'' in the table) achieves, obviously, better results than the pure LO methods, hence
their values are not marked in bold since the method is qualitatively different.

\begin{table*}
\caption{Relative translational error (RTE) (\%), relative rotational error (RRE) (d/hm, that is, degrees per 100~m) (as defined in \cite{geiger2013vision}), and absolute translational error (ATE) (rmse values in meters as reported by ``\texttt{evo\_ape -a}'')
for the training and evaluation sets of sequences in the \textbf{KITTI visual-LiDAR odometry dataset}.
Bold figures mark either the absolute best
for each sequence if it is pure odometry method (no loop closure), or the best ones for each category (with and without loop closure).
Average RTE is also shown for the whole training (00-10) and validating (11-21) KITTI sequence subsets, with the latter being publicly available 
on the KITTI odometry leaderboard website.
Starred sequence numbers are those having loop closures.
Results for IMLS-SLAM, MULLS, and CT-ICP2 have been taken from their respective publications (hence the missing RRE and ATE values).
The rest have been evaluated 
manually from result trajectories.
}
\resizebox{\textwidth}{!}{
\setlength\extrarowheight{1pt}
\begin{tabular}{|c||c|c|c|c|c|c|c|c|c|c|c||c|c|c|}
\hline
\textbf{Method}
& \texttt{00}$^{\star}$ & \texttt{01} & \texttt{02}$^{\star}$ & \texttt{03} & \texttt{04} & \texttt{05}$^{\star}$ & \texttt{06}$^{\star}$ & \texttt{07}$^{\star}$ & \texttt{08}$^{\star}$ & \texttt{09}$^{\star}$ & \texttt{10} 
& \TwoRows{Avr.}{00-10}
& \TwoRows{Avr.}{11-21}
& \TwoRows{Avr. time}{per frame}
\\
\hhline{|-#*{10}{-|}-#-|-|-|}
\TwoRows{IMLS-SLAM}{\cite{deschaud2018imls}} & 
\textbf{0.50}\%  &  0.82\% &  0.53\% & 0.68\% & \textbf{0.33\%} &  0.32\% &  0.33\% &  0.33\% &  \textbf{0.80\%}  & 0.55\% & {0.53\%} & \textbf{0.55}\% & 
\TwoRowsR{0.69\%}{0.18~$d/hm$}
& $\sim$1000~ms\\
\hhline{|-#*{10}{-|}-#-|-|-|}
\TwoRows{KISS-ICP}{\cite{vizzo2023kiss}} & %
\ThreeRows{0.51\%}{0.19~$d/hm$}{3.74~m} &
\ThreeRows{\textbf{0.72}\%}{0.11~$d/hm$}{3.72~m} &
\ThreeRows{\textbf{0.52}\%}{0.15~$d/hm$}{6.60~m} &
\ThreeRows{0.66\%}{0.16~$d/hm$}{\textbf{0.43}~m} &
\ThreeRows{0.36\%}{0.14~$d/hm$}{{0.27}~m} &
\ThreeRows{\textbf{0.30}\%}{0.14~$d/hm$}{1.47~m} &
\ThreeRows{\textbf{0.26}\%}{0.08~$d/hm$}{0.39~m} &
\ThreeRows{0.32\%}{\textbf{0.17}~$d/hm$}{\textbf{0.30}~m} &
\ThreeRows{0.82\%}{0.18~$d/hm$}{2.23~m} &
\ThreeRows{{0.49}\%}{0.13~$d/hm$}{1.45~m} &
\ThreeRows{0.57\%}{0.19~$d/hm$}{0.81~m} &
\textbf{0.55}\% & 
\TwoRowsR{\textbf{0.61}\%}{\textbf{0.17}~$d/hm$}
& \textbf{13} ms
\\
\hhline{|-#*{10}{-|}-#-|-|-|}
\TwoRows{SiMpLE (offline)}{\cite{bhandari2024minimal}} & 
\ThreeRows{0.51\%}{\textbf{0.18}~$d/hm$}{\textbf{3.67}~m} &
\ThreeRows{0.86\%}{\textbf{0.09}~$d/hm$}{\textbf{2.50}~m} &
\ThreeRows{0.53\%}{\textbf{0.13}~$d/hm$}{\textbf{4.74}~m} &
\ThreeRows{0.70\%}{\textbf{0.14}~$d/hm$}{0.45~m} &
\ThreeRows{0.40\%}{\textbf{0.10}~$d/hm$}{{0.27}~m} &
\ThreeRows{\textbf{0.30}\%}{\textbf{0.13}~$d/hm$}{\textbf{1.23}~m} &
\ThreeRows{0.28\%}{0.08~$d/hm$}{0.34~m} &
\ThreeRows{\textbf{0.30}\%}{0.18~$d/hm$}{0.38~m} &
\ThreeRows{\textbf{0.80}\%}{\textbf{0.16}~$d/hm$}{\textbf{1.97}~m} &
\ThreeRows{0.55\%}{\textbf{0.12}~$d/hm$}{\textbf{1.30}~m} &
\ThreeRows{{0.51}\%}{\textbf{0.15}~$d/hm$}{{0.76}~m} &
\textbf{0.55}\% & 0.62\% & 545~ms
\\
\hhline{|-#*{10}{-|}-#-|-|-|}
\TwoRows{SiMpLE (online)}{\cite{bhandari2024minimal}} &
\ThreeRows{0.95\%}{0.45~$d/hm$}{5.45~m} &
\ThreeRows{1.48\%}{0.36~$d/hm$}{12.25~m} &
\ThreeRows{1.07\%}{0.40~$d/hm$}{7.11~m} &
\ThreeRows{1.34\%}{0.41~$d/hm$}{0.59~m} &
\ThreeRows{0.80\%}{0.53~$d/hm$}{0.48~m} &
\ThreeRows{0.61\%}{0.41~$d/hm$}{1.95~m} &
\ThreeRows{0.59\%}{0.31~$d/hm$}{0.92~m} &
\ThreeRows{0.50\%}{0.57~$d/hm$}{0.81~m} &
\ThreeRows{1.13\%}{0.37~$d/hm$}{2.66~m} &
\ThreeRows{1.19\%}{0.41~$d/hm$}{2.15~m} &
\ThreeRows{1.43\%}{0.55~$d/hm$}{1.98~m} &
0.62\% & N/A & 126~ms
\\
\hhline{|-#*{10}{-|}-#-|-|-|}
\TwoRows{\textbf{MOLA-LO} (default)}{(ours)} &
\ThreeRows{0.51\%}{0.19~$d/hm$}{3.85~m} &
\ThreeRows{0.74\%}{0.14~$d/hm$}{5.08~m} &
\ThreeRows{\textbf{0.52}\%}{0.15~$d/hm$}{6.91~m} &
\ThreeRows{{0.63}\%}{0.17~$d/hm$}{\textbf{0.43}~m} &
\ThreeRows{0.37\%}{0.13~$d/hm$}{0.28~m} &
\ThreeRows{0.33\%}{0.15~$d/hm$}{1.34~m} &
\ThreeRows{0.29\%}{\textbf{0.07}~$d/hm$}{\textbf{0.29}~m} &
\ThreeRows{0.36\%}{0.18~$d/hm$}{0.37~m} &
\ThreeRows{0.81\%}{0.19~$d/hm$}{2.35~m} &
\ThreeRows{{0.49}\%}{0.14~$d/hm$}{1.83~m} &
\ThreeRows{0.57\%}{0.18~$d/hm$}{0.81~m} &
\textbf{0.55}\%
& 
\TwoRowsR{0.62\%}{\textbf{0.17}~$d/hm$}
& 23 ms\\
\hline
\TwoRows{\REVIEW{\textbf{MOLA-LO}} (\REVIEW{3D-NDT})}{(ours)} &
\ThreeRows{0.56\%}{0.21~$d/hm$}{3.74~m} &
\ThreeRows{0.87\%}{0.12~$d/hm$}{3.61~m} &
\ThreeRows{0.57\%}{0.17~$d/hm$}{8.34~m} &
\ThreeRows{\textbf{0.60}\%}{0.18~$d/hm$}{0.50~m} &
\ThreeRows{0.41\%}{0.12~$d/hm$}{\textbf{0.26}~m} &
\ThreeRows{0.35\%}{0.15~$d/hm$}{1.24~m} &
\ThreeRows{0.31\%}{0.08~$d/hm$}{0.35~m} &
\ThreeRows{0.33\%}{\textbf{0.17}~$d/hm$}{0.39~m} &
\ThreeRows{0.82\%}{0.20~$d/hm$}{2.36~m} &
\ThreeRows{\textbf{0.47}\%}{0.16~$d/hm$}{1.40~m} &
\ThreeRows{\textbf{0.50}\%}{0.21~$d/hm$}{\textbf{0.75}~m} &
0.58\% & N/A & 32~ms
\\
\hline
\TwoRows{\REVIEW{\textbf{MOLA-LO}} (\REVIEW{Horn's})}{(ours)} &
\ThreeRows{0.69\%}{0.29~$d/hm$}{4.88~m} &
\ThreeRows{0.73\%}{0.14~$d/hm$}{4.97~m} &
\ThreeRows{0.62\%}{0.21~$d/hm$}{9.17~m} &
\ThreeRows{0.64\%}{0.17~$d/hm$}{0.45~m} &
\ThreeRows{0.37\%}{0.14~$d/hm$}{0.28~m} &
\ThreeRows{0.42\%}{0.22~$d/hm$}{1.65~m} &
\ThreeRows{0.28\%}{0.08~$d/hm$}{0.31~m} &
\ThreeRows{0.47\%}{0.29~$d/hm$}{0.51~m} &
\ThreeRows{0.87\%}{0.20~$d/hm$}{2.57~m} &
\ThreeRows{0.68\%}{0.18~$d/hm$}{2.43~m} &
\ThreeRows{0.65\%}{0.22~$d/hm$}{0.97~m} &
0.65\% & N/A & 45~ms
\\
\hhline{|=#=|=|=|=|=|=|=|=|=|=|=#=|=|=|}
\TwoRows{MULLS (LC)}{\cite{pan2021mulls}} &  0.54\%  & \textbf{0.62\%} & 0.69\% & \textbf{0.61}\% & \textbf{0.35}\% &   
0.29\% & 0.29\% & \textbf{0.27}\% & 0.83\% & 0.51\% & 0.61\% & \textbf{0.52}\% & \TwoRowsR{0.65\%}{0.19~$d/hm$}
 & 100 ms 
\\
\hhline{|-#*{10}{-|}-#-|-|-|}
\TwoRows{CT-ICP2 (LC)}{\cite{dellenbach2022ct}} &
  \textbf{0.49\%}  &  0.76\% &  \textbf{0.52}\% & 
    0.72\% &   0.39\% &   \textbf{0.25\%} &  \textbf{0.27}\% &  0.31\% &  \textbf{0.81}\%  & \textbf{0.49}\% & {0.48\%} & 0.53\% & 
\TwoRowsR{\textbf{0.58}\%}{\textbf{0.12}~$d/hm$}
 & 60 ms\\
\hhline{|-#*{10}{-|}-#-|-|-|}
\TwoRows{\textbf{MOLA-LO (default)}}{\textbf{+ LC} (ours)} &
\ThreeRows{0.60\%}{0.17~$d/hm$}{0.81~m} &
\ThreeRows{0.74\%}{0.14~$d/hm$}{5.08~m} &
\ThreeRows{0.55\%}{0.13~$d/hm$}{2.90~m} &
\ThreeRows{0.63\%}{0.17~$d/hm$}{0.43~m} &
\ThreeRows{0.37\%}{0.13~$d/hm$}{0.28~m} &
\ThreeRows{0.34\%}{0.13~$d/hm$}{0.34~m} &
\ThreeRows{0.33\%}{0.13~$d/hm$}{0.34~m} &
\ThreeRows{0.39\%}{0.16~$d/hm$}{0.28~m} &
\ThreeRows{\textbf{0.81}\%}{0.17~$d/hm$}{2.05~m} &
\ThreeRows{\textbf{0.49}\%}{0.14~$d/hm$}{1.83~m} &
\ThreeRows{0.57\%}{0.18~$d/hm$}{0.81~m} &
{0.58\%} &
\TwoRowsR{0.66\%}{0.16~$d/hm$}
& \textbf{31} ms
\\
\hline
\end{tabular}
}
\label{tab:kitti.metrics}
\end{table*}

\subsection{KITTI-odometry dataset}
\label{sect:kitti}

The KITTI odometry dataset \citep{geiger2013vision} was the first widely spread benchmark
for LiDAR and visual odometry and SLAM methods, 
hence its popularity and importance.
It comprises 11 training sequences (numbered \texttt{00}-\texttt{10})
with public ground truth and other 11 evaluation sequences with undisclosed ground truth trajectories.
Available sensors in this driving dataset 
include 3D LiDAR (a Velodyne HDL-64E)
and two pairs of forward-facing cameras.

The trajectories estimated by our system \REVIEW{(with its default configuration)} for the 11 training sequences are shown in \Figure{fig:kitti.paths}
for the LO-only module and for the SLAM system (including loop closure), compared with ground truth.
Quantitative results are shown in Table~\ref{tab:kitti.metrics}
and graphically in \REVIEW{Figure~\ref{fig:kitti.rte.summary.graph}.
A larger number of alternative methods
are compared in this case due to the popularity of this particular dataset.}
Note that two configurations of SiMpLE \citep{bhandari2024minimal} have been included (``online'' and ``offline''),
with one of them being faster and the other more accurate. However, note that none would be able to 
run at sensor rate (10 Hz). Excepting IMLS-SLAM and SiMpLE, all others are fast enough to run in real-time
at the sensor rate or faster. 

\graphicspath{{./figs/}}

\begin{figure*}
\centering
\begin{tabular}[t]{ccc}
\subfigure[\texttt{00} (8.4~km long)]{\hspace*{-0.0cm}\includegraphics[width=0.325\textwidth]{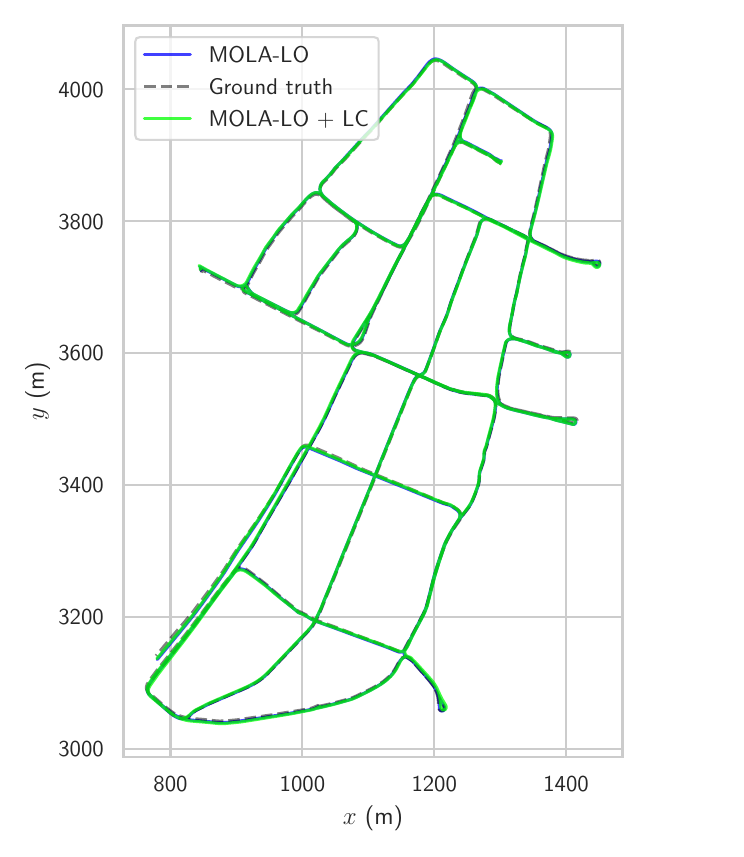}}
&
\subfigure[\texttt{03} (1.4~km long)]{\hspace*{-0.0cm}\includegraphics[width=0.325\textwidth]{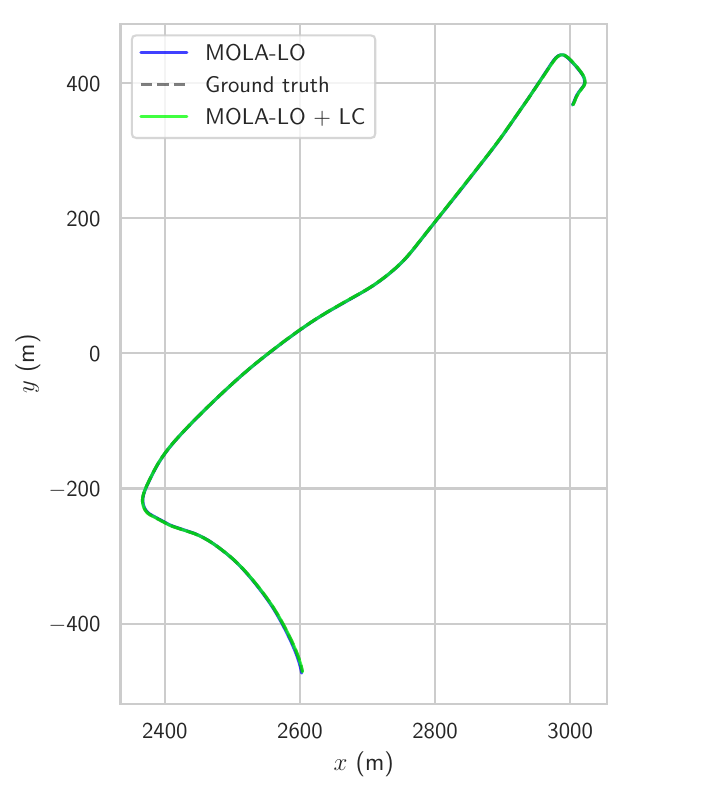}}
&
\subfigure[\texttt{04} (10.0~km long)]{\hspace*{-0.0cm}\includegraphics[width=0.325\textwidth]{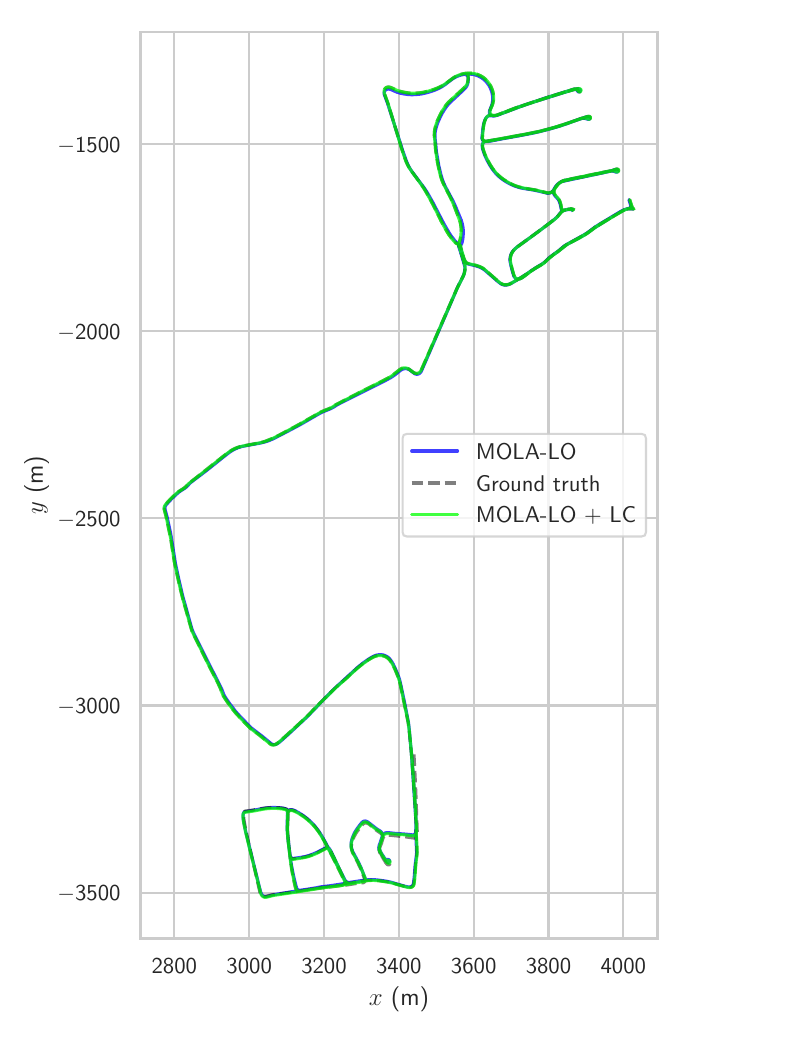}}
\\
\subfigure[\texttt{05} (4.7~km long)]{\hspace*{-0.0cm}\includegraphics[width=0.325\textwidth]{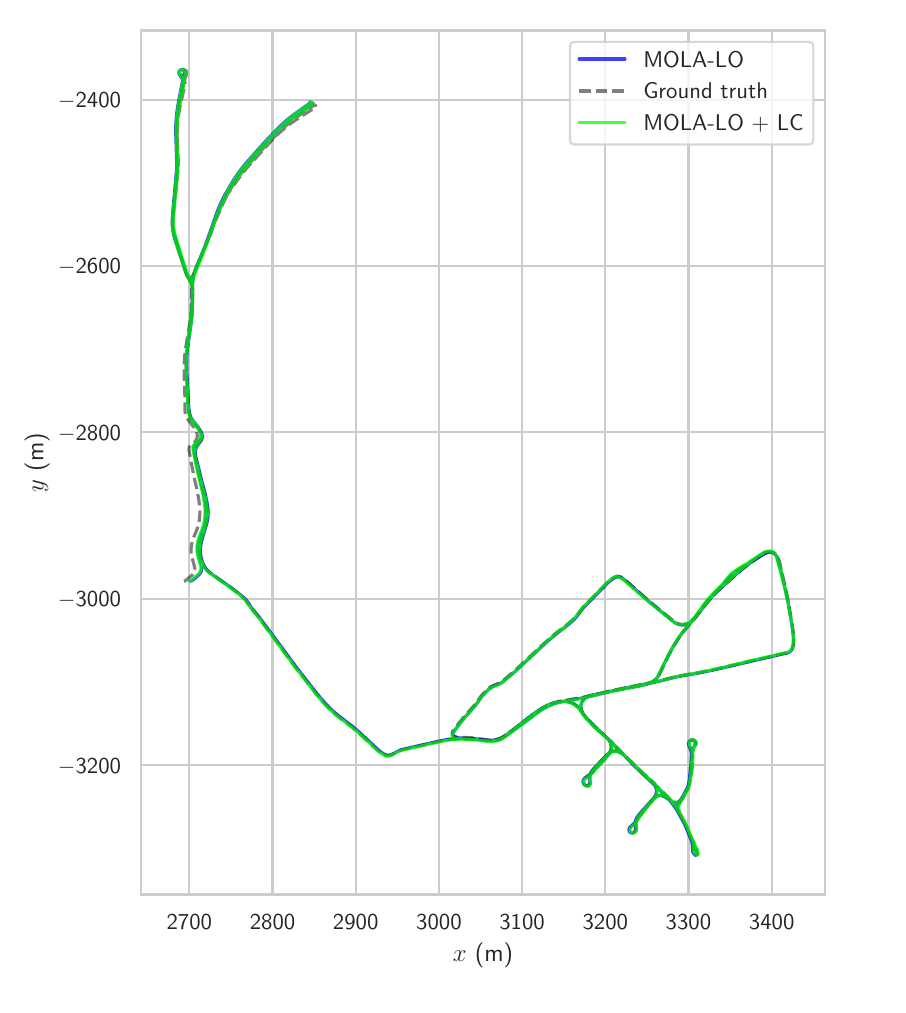}}
&
\subfigure[\texttt{06} (8.0~km long)]{\hspace*{-0.0cm}\includegraphics[width=0.325\textwidth]{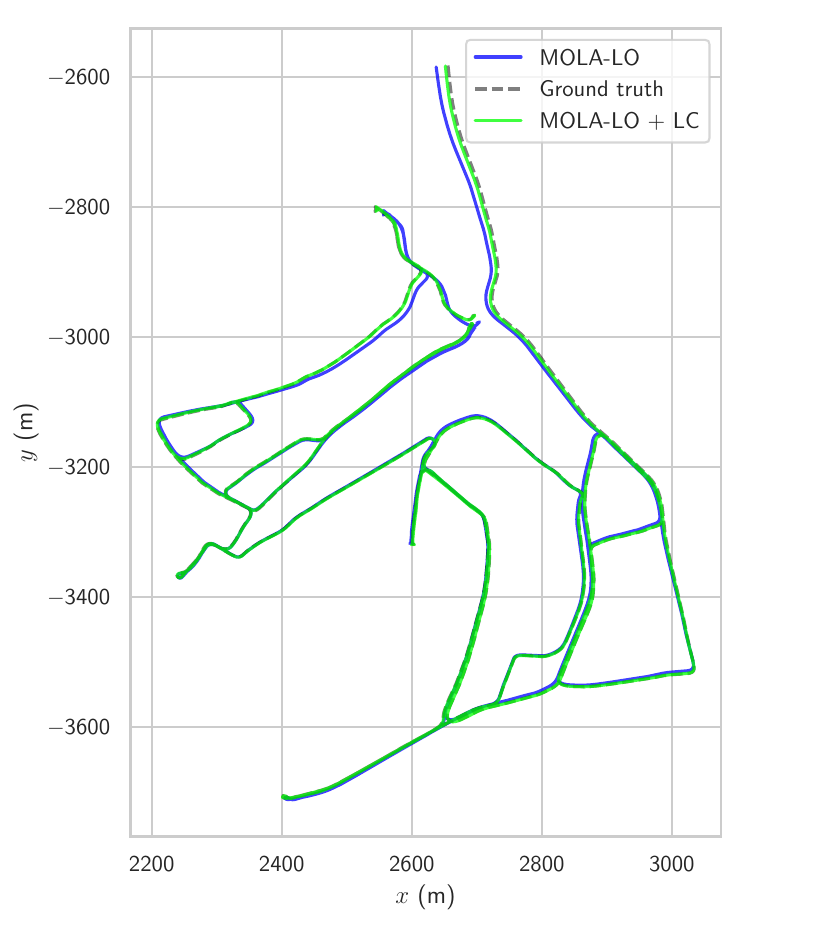}}
&
\subfigure[\texttt{07} (4.9~km long)]{\hspace*{-0.0cm}\includegraphics[width=0.325\textwidth]{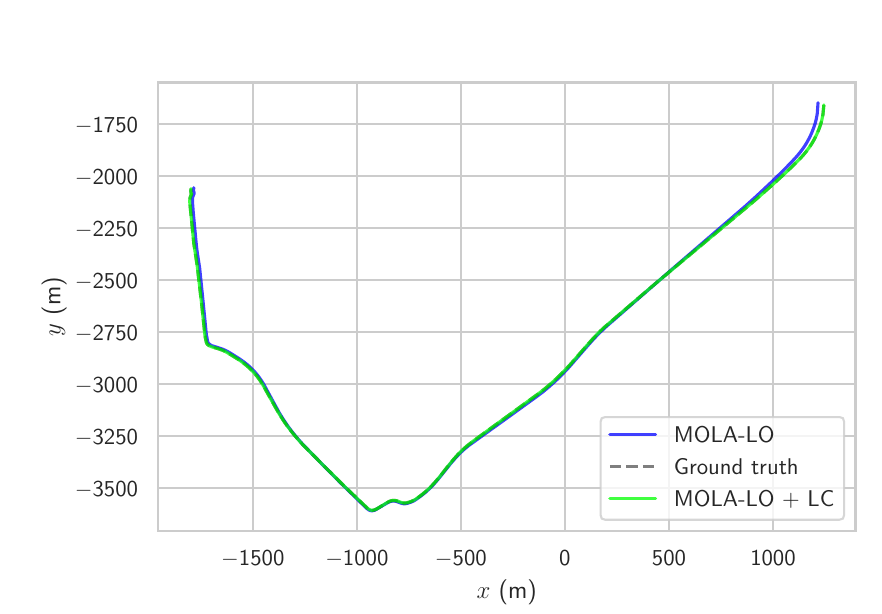}}
\\
\subfigure[\texttt{09} (10.6~km long)]{\hspace*{-0.0cm}\includegraphics[width=0.325\textwidth]{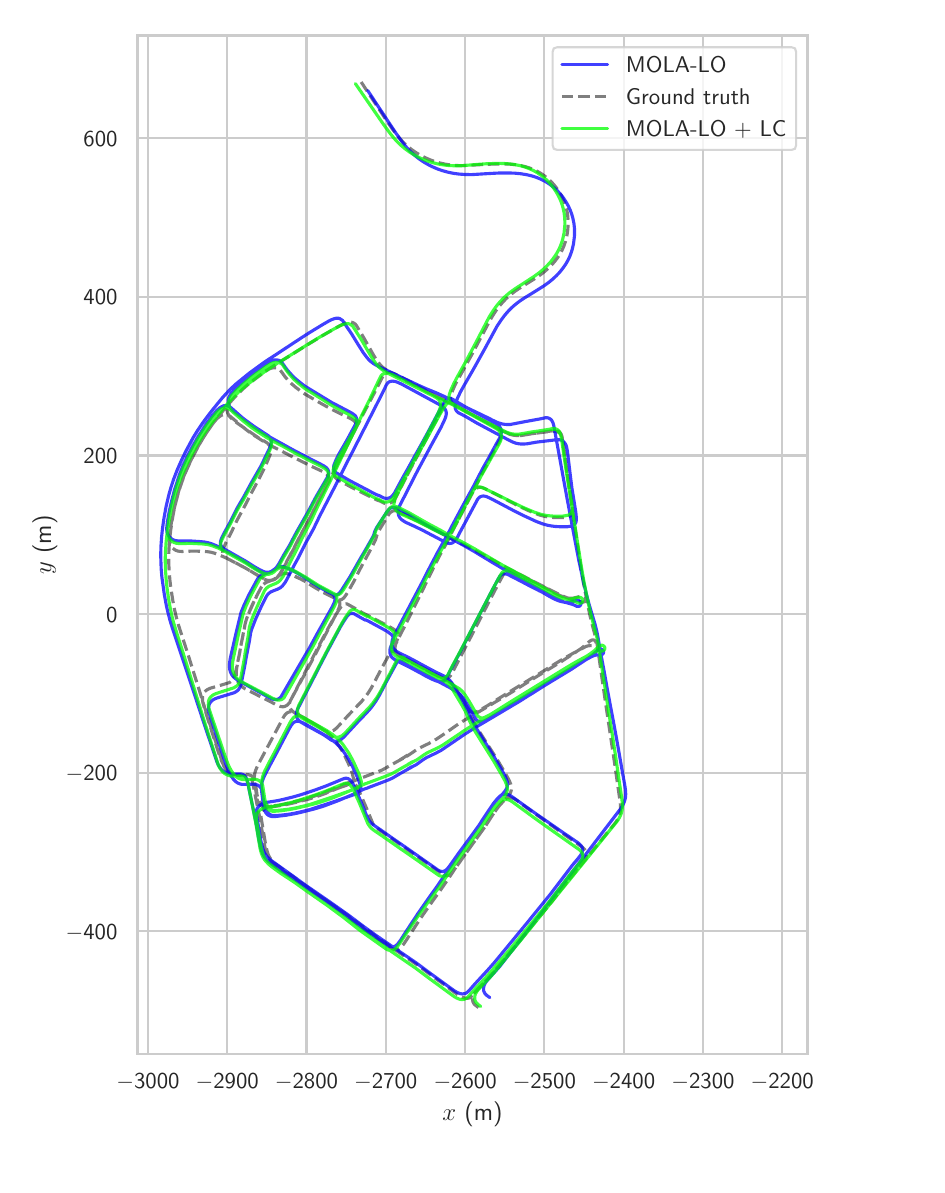}}
&
\subfigure[\texttt{10} (3.3~km long)]{\hspace*{-0.0cm}\includegraphics[width=0.325\textwidth]{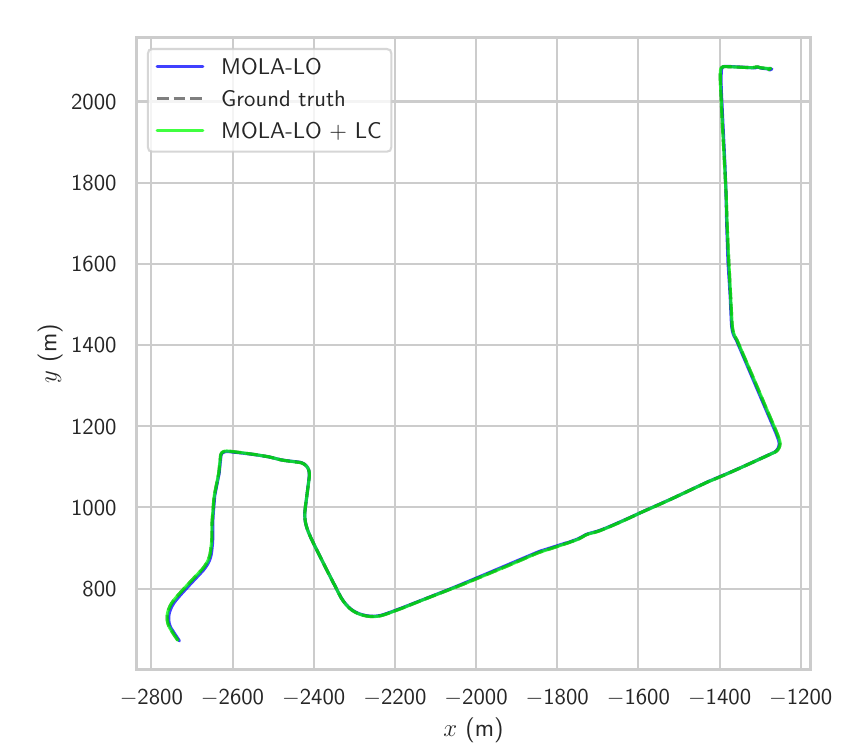}}
\end{tabular}
\caption{Estimated trajectories for the proposed LiDAR odometry system ("MOLA-LO") applied to the \textbf{KITTI-360 odometry dataset}, compared to ground truth. Trajectories denoted as ``MOLA-LO + LC'' include the post-processing loop-closure
stage. See discussion in Section~\ref{sect:kitti360}.}
\label{fig:kitti360.paths}
\end{figure*}

Regarding the accuracy of the estimated trajectories for LO methods, it can be seen in 
Table~\ref{tab:kitti.metrics} that the slower and more precise configuration of SiMpLE 
\REVIEW{is slightly better than the rest, although accuracy metrics for these SOTA methods are all quite similar,}
as demonstrated by an almost perfect tie in the average RTE metric used to evaluate all training ($\text{RTE}\sim 0.55\%$)
and all testing sequences ($\text{RTE}\sim 0.61-0.62\%$).

Note that all shown results for KISS-ICP, SiMpLE, and our system, have been run using the same MOLA 
KITTI dataset source (see Section~\ref{sect:dataset.sources}), which includes the $0.205^\circ$ vertical
angle correction of all original dataset scans due to a miscalibration; see \cite{deschaud2018imls}.
Another well-known issue with KITTI is the wrong ground truth in part of sequence \texttt{08}, which 
explains the relative elevated apparent errors of all methods in that sequence.

Regarding the results on the evaluation sequences (\texttt{11}-\texttt{21}), 
for which ground truth is not publicly available,
at the time of writing 
our system ranks 13\textsuperscript{rd} among all LiDAR-only methods (out of 140 total submissions),
and 6\textsuperscript{th} among those with an open source implementation,
according to the project website\footnote{See: \url{https://www.cvlibs.net/datasets/kitti/eval_odometry.php}}.

\subsection{KITTI-360 dataset}
\label{sect:kitti360}

This automotive dataset is a follow-up of the first KITTI dataset, including more sensors (fish-eye cameras, 2D LiDAR, and IMU)
and longer driving sequences in suburban scenarios \citep{Liao2022PAMI}.

\begin{figure*}
\centering
\resizebox{\textwidth}{!}
{
\begin{tabular}[t]{cc}
\subfigure[Trajectories]{\includegraphics[width=0.505\textwidth]{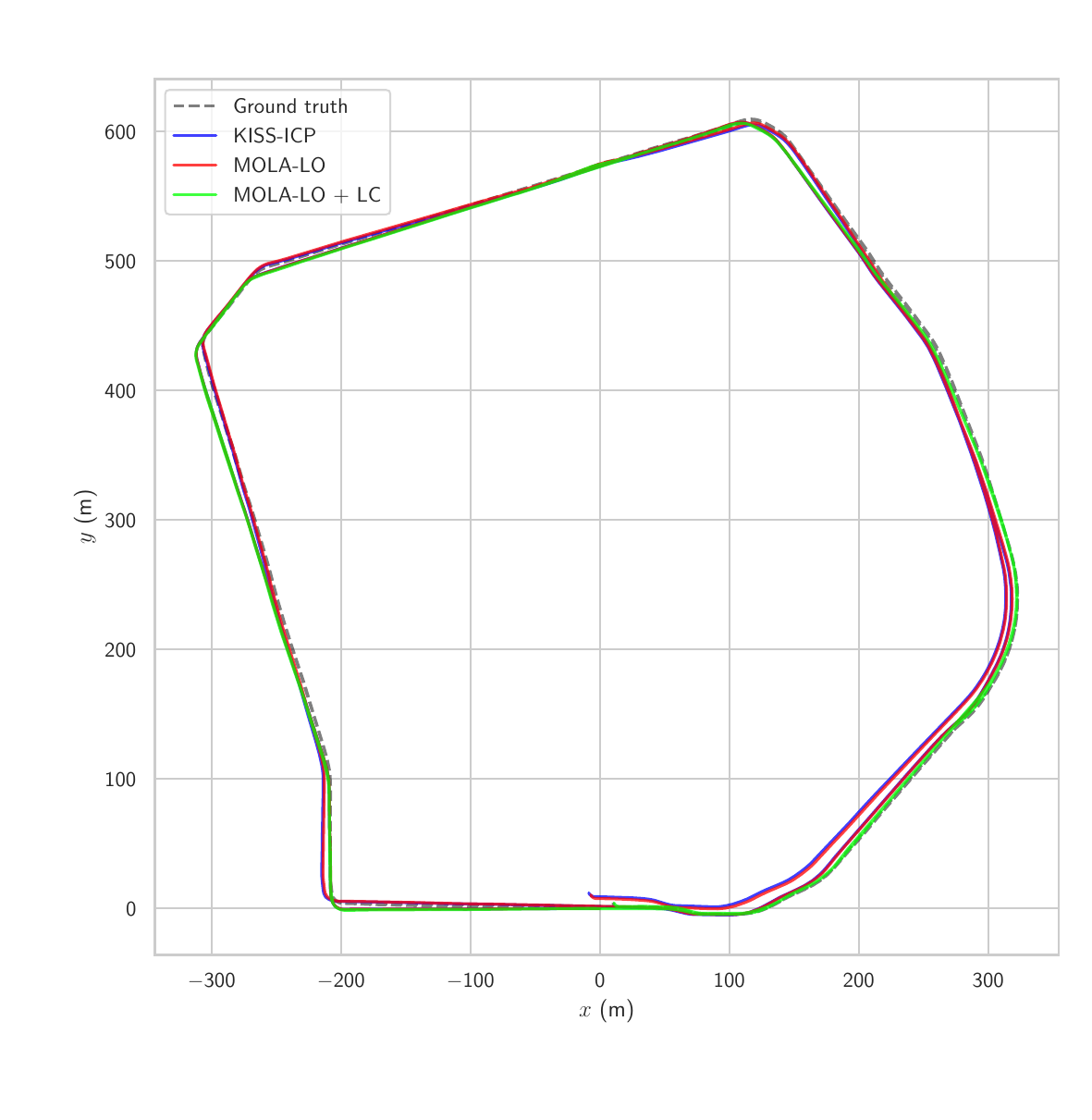}}
&
\subfigure[Trajectory components]{\includegraphics[width=0.505\textwidth]{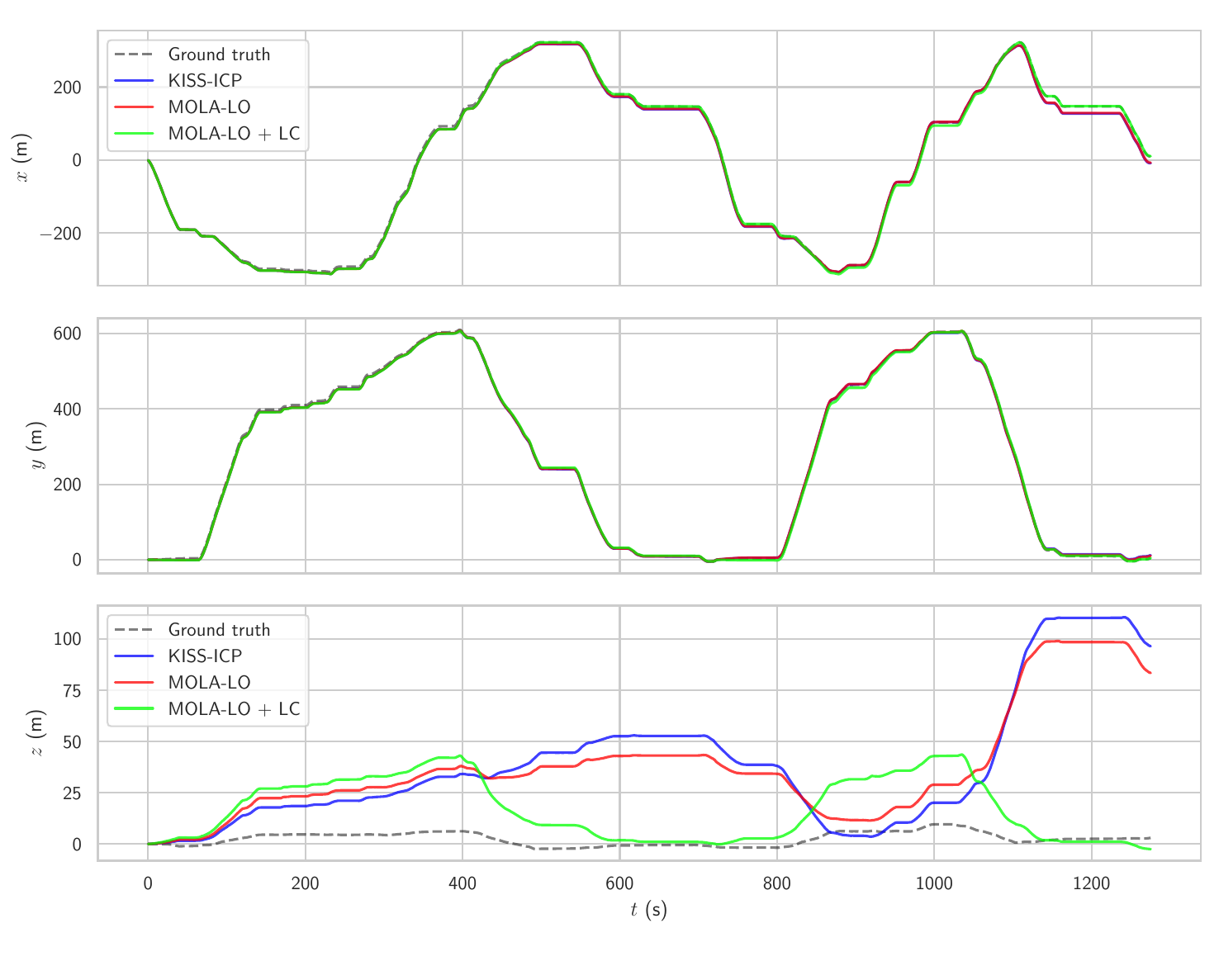}}
\end{tabular}
}
\caption{Estimated trajectories for the proposed LiDAR odometry system ("MOLA-LO") applied to the \textbf{Paris LuCo} dataset, compared to KISS-ICP, and to ground truth. Trajectories denoted as ``MOLA-LO + LC'' include loop-closure. See discussion in Section~\ref{sect:paris}.
}
\label{fig:paris-paths}
\end{figure*}

Trajectories estimated by our system are shown in \Figure{fig:kitti360.paths},
while quantitative performance results, compared to KISS-ICP as SOTA baseline, are summarized in Table~\ref{tab:kitti360.metrics}.
As can be seen in the table, where we evaluated the ATE metric, our system achieves better
accuracy than KISS-ICP in 6 out of 8 sequences. Regarding our SLAM method with loop closures, 
and given that most KITTI-360 sequences, except for \texttt{03}, \texttt{07}, and \texttt{10}, have multiple loop-closures, 
obtained ATE is further reduced.
This dataset also includes 4 evaluation sequences without public ground truth for benchmarking 
on the ``trajectory estimation'' public leaderboard website\footnote{See: \url{https://www.cvlibs.net/datasets/kitti-360/leaderboard_semantic_slam.php?task=trajectory}.}, 
where our SLAM solution ranks 3\textsuperscript{rd} out of 5 submissions at the time of writing.
In this leaderboard, more modern than the original KITTI, the metric that determines the classification
is APE, and the results available online are: CT-ICP2 with LiDAR SLAM in \citep{dellenbach2022ct} (0.50~m), 
SOFT2 with stereo visual SLAM in \citep{soft2} (0.70~m), our SLAM system (0.72~m), 
ORB-SLAM2 with stereo visual SLAM in \citep{Artal2017TR} (1.92~m), and SUMA++ with LiDAR SLAM in \citep{Chen2019IROS} (3.13~m).

\begin{table}[t]
\caption{Absolute translational error (ATE) (rmse values in meters as reported by ``\texttt{evo\_ape -a}'')
for the \textbf{KITTI-360} dataset.
Bold means best accuracy in each sequence and category.}
\centering
\resizebox{\columnwidth}{!}
{
\setlength\extrarowheight{1pt}
\begin{tabular}{|c||r|r|r|r|r|r|r|r|r|}
\hline
\textbf{Method}
& \texttt{00}
& \texttt{03}
& \texttt{04}
& \texttt{05}
& \texttt{06}
& \texttt{07}
& \texttt{09}
& \texttt{10}
\\
\hhline{|-|*{8}{-|}}
\TwoRows{KISS-ICP}{\cite{vizzo2023kiss}} & 
5.50~m & \textbf{0.62}~m & 5.85~m & 3.05~m & 10.45~m & \textbf{4.22}~m & 12.30~m & 5.90~m
\\ %
\hline
\TwoRows{MOLA-LO}{(ours)} & 
\textbf{2.81}~m & 0.72~m & \textbf{4.89}~m & \textbf{2.39}~m & \textbf{6.37}~m & 8.22~m & \textbf{10.78}~m & \textbf{1.75}~m
\\ %
\hhline{|*{9}{=|}}
\TwoRows{MOLA-LO + LC}{(ours)}
 & 
0.72~m & 0.72~m & 4.89~m & 1.02~m & 4.03~m & 8.22~m & 4.67~m & 1.75~m
\\
\hline
\end{tabular}
}
\label{tab:kitti360.metrics}
\end{table}

\subsection{ParisLuco dataset}
\label{sect:paris}

This automotive dataset was presented in \cite{deschaud2018imls} and comprises
one single continuous driving sequence with two loops around the Luxembourg Garden in Paris, while 
grabbing 3D LiDAR scans from a Velodyne HDL-32.
\Figure{fig:paris-paths}(a) shows the trajectories estimated by the reference SOTA method KISS-ICP, 
our LO method, and our full SLAM solution including loop closure detection, along with ground truth.
It can be seen that both LO methods perform very similarly, with the SLAM solution being, as expected,
much closer to ground truth. 
Looking at the quantitative ATE metrics, in Table~\ref{tab:paris.metrics}, it can be seen that both LO 
methods have similar accuracy, with KISS-ICP having a small advantage in this case.
It is interesting to analyze where these trajectory errors tend to accumulate. 
As it typically occurs with 3D LiDAR LO, errors accumulate faster
in the vertical direction (perpendicular to the ground plane), as can be clearly seen in the trajectory component 
plots in \Figure{fig:paris-paths}(b), where absolute horizontal errors are in the order of 1 meter while vertical errors 
reach a maximum of 100 meters. Loop closures, naturally, bound these vertical errors, reducing them to 
a maximum of $\sim 45~m$.

\begin{table}[t]
\caption{Absolute translational error (ATE) (rmse values in meters as reported by ``\texttt{evo\_ape -a}'')
for the \textbf{Paris LuCo} dataset, in which there is only one sequence.
Bold means best accuracy in each sequence and category.
}
\centering
{
\setlength\extrarowheight{1pt}
\begin{tabular}{|c||r|r|}
\hline
\textbf{Method}
& Paris LuCo
& \TwoRows{Avr. time}{per frame}
\\
\hhline{|-|*{2}{-|}}
\TwoRows{KISS-ICP}{\cite{vizzo2023kiss}} & 
21.2~m & 24~ms
\\ %
\hline
\TwoRows{MOLA-LO (default)}{(ours)} & 
22.1~m & 23~ms
\\ %
\hline
\TwoRows{\REVIEW{MOLA-LO (3D-NDT)}}{(ours)} & 
\REVIEW{\textbf{12.3~m}} & \REVIEW{88~ms}
\\ %
\hhline{|*{3}{=|}}
\TwoRows{MOLA-LO (default) + LC}{(ours)}
 & 
2.38~m & 50~ms
\\
\hline
\end{tabular}
}
\label{tab:paris.metrics}
\end{table}

\subsection{NTU VIRAL: two 3D LiDARs}
\label{sect:ntu}

This airborne dataset was presented in \cite{nguyen2022ntu}, 
and comprises 9 sequences (three in each of three scenarios)
of visual, inertial, and LiDAR data as a drone flights within
the range of a motion capture system that serves as accurate ground truth.
Two Velodyne VLP-16 scanners are installed in the drone, one in an horizontal
position (as typically placed in automotive datasets) and another one facing vertically
towards the ground.
Hence, this dataset is ideal to benchmark visual or LiDAR odometry algorithms, 
possibly making use of high-rate inertial data, in the challenging conditions of drone flight, 
where linear and angular accelerations tend to be more abrupt than in automotive datasets.
The dataset was used in recent works like \cite{nguyen2024eigen} to evaluate 
tightly-coupled LiDAR-inertial methods. 
Table~\ref{tab:ntu.viral.metrics} compares the performance of SOTA 
visual-LiDAR-inertial methods reported in \cite{nguyen2024eigen}
to our solution and KISS-ICP, although they do not make use of neither inertial data, nor images.
Loop-closure was not evaluated in this dataset since the range of motion is in the order of magnitude
of the LiDAR range, hence there are no loop closures.

\begin{table*}
\caption{Absolute translational error (ATE) (rmse values in meters as reported by ``\texttt{evo\_ape -a}'')
for the \textbf{NTU Viral} dataset. Bold means best accuracy in each sequence.}
\centering
\resizebox{\textwidth}{!}
{
\setlength\extrarowheight{1pt}
\begin{tabular}{|c||r|r|r|r|r|r|r|r|r|r|}
\hline
\textbf{Method}
& \multicolumn{3}{c|}{\texttt{EEE}}
& \multicolumn{3}{c|}{\texttt{NYA}}
& \multicolumn{3}{c|}{\texttt{SBS}}
& Avr. time
\\
\hhline{|~|*{9}{-|}~|}
~
& \texttt{01} & \texttt{02} & \texttt{03} 
& \texttt{01} & \texttt{02} & \texttt{03} 
& \texttt{01} & \texttt{02} & \texttt{03} 
& per frame
\\ %
\hline
\ThreeRowsC{LIOSAM}{(1 LiDAR + IMU)}{\cite{shan2020lio}} & 
0.075~m & 0.069~m & 0.101~m & 
0.076~m & 0.090~m & 0.137~m &
0.089~m & 0.083~m & 0.140~m &
$<100$~ms 
\\ %
\hline
\ThreeRowsC{VIRAL SLAM}{(2 LiDARs + IMU)}{\cite{nguyen2021viral}} & 
\textbf{0.060}~m & \textbf{0.058}~m & \textbf{0.037}~m &
\textbf{0.051}~m & \textbf{0.043}~m & \textbf{0.032}~m &
\textbf{0.048}~m & \textbf{0.062}~m & \textbf{0.054}~m &
$<100$~ms 
\\ %
\hline
\TwoRows{KISS-ICP (1 LiDAR)}{\cite{vizzo2023kiss}} & 
2.383~m & 1.586~m & 1.055~m & 0.359~m & $\times$ & 1.389~m & 1.353~m & 1.435~m & 1.037~m
&
31.0~ms
\\ %
\hline
\TwoRows{MOLA-LO (ours)}{(1 LiDAR)} & 
1.484~m & 1.639~m & 1.046~m & 0.746~m & 1.256~m & 0.424~m & 0.361~m & 1.015~m & 1.066~m
&
31.6~ms
\\
\hline
\TwoRows{MOLA-LO (ours)}{(2 LiDARs)} & 
0.780~m & 1.574~m & 0.725~m & 0.755~m & 0.595~m & 0.427~m & 0.344~m & 1.000~m & 0.914~m
&
40.0~ms
\\
\hline
\TwoRows{MOLA-LO (ours)}{(2 LiDARs + 2 maps)} & 
0.100~m & 0.362~m & 0.175~m & 0.220~m & 0.220~m & 0.121~m & 0.118~m & 0.123~m & 0.115~m
&
69.4~ms 
\\
\hline
\end{tabular}
}
\label{tab:ntu.viral.metrics}
\end{table*}

\begin{figure*}
\centering
{
\begin{tabular}{cc}
\subfigure[3D trajectory]{\includegraphics[width=0.28\textwidth]{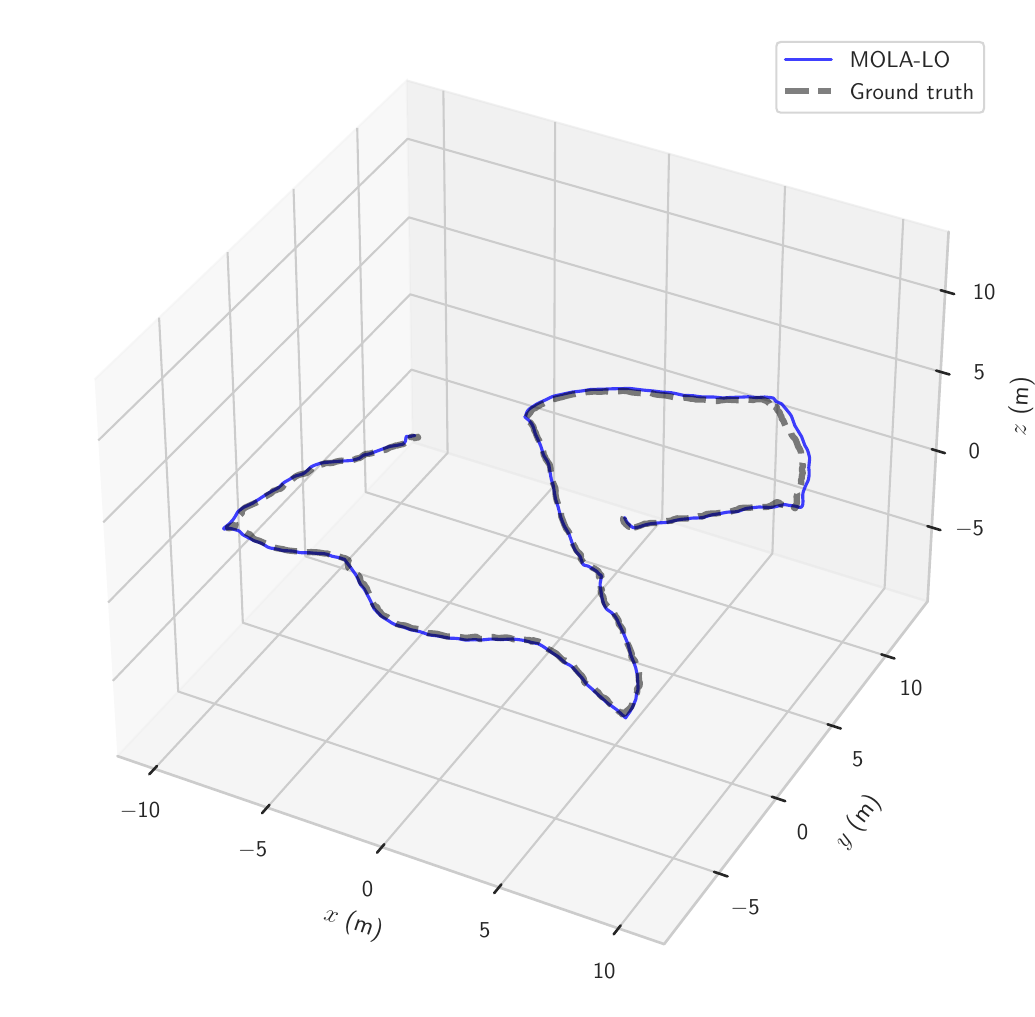}}
&
\subfigure[Trajectory components]{\includegraphics[width=0.70\textwidth]{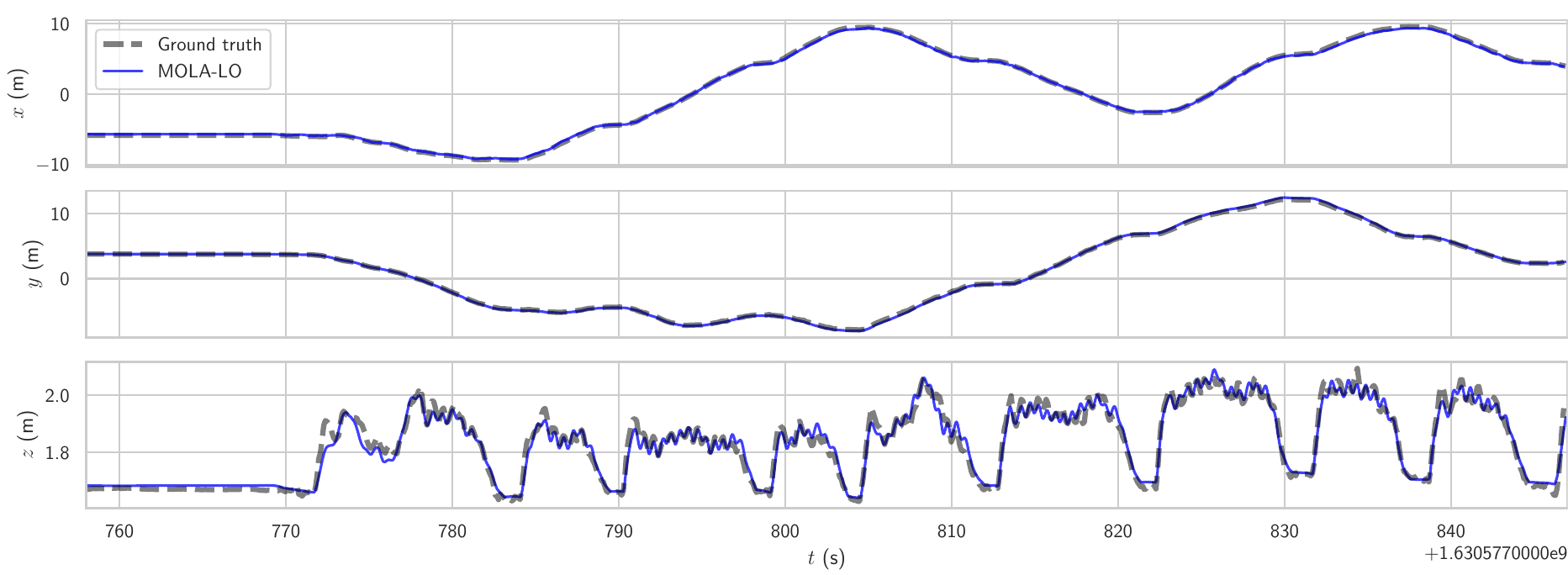}}
\end{tabular}
}
\caption{Estimated trajectories for the proposed LiDAR odometry system ("MOLA-LO") applied to the RPG Drone Testing Arena sequence of Drone Testing Area \textbf{HILTI-2021} dataset, compared to ground truth.
}
\label{fig:hilti2021.paths}
\end{figure*}

From the accuracy metrics, it is clear that methods using inertial sensors (LIOSAM and VIRAL SLAM)
have the best performance, although we believe that the smarter sampling of LiDAR scans 
using edge and plane key points \citep{nguyen2021miliom} plays a more significant role in achieving better accuracy than using an IMU; 
this topic would require further research.
In contrast, LO methods KISS-ICP and ours, using only one (horizontal) LiDAR, have trajectory errors at least one order of magnitude larger
than the aforementioned LIO methods, with KISS-ICP performing poorer in this case.
Analyzing the spatial trajectories, it becomes clear that vertical movements of the drone are where these two methods perform worst, which 
can be explained by the simplistic uniform sampling of point clouds (instead of searching for specific features), which makes hard to distinguish
vertical motions from remaining static. 
With this insight in mind, we provide an example of the flexibility of the present framework by adding two additional experiments evaluating MOLA-LO (the last two rows in Table~\ref{tab:ntu.viral.metrics}). 
First, we put at test the capability of our system to handle several LiDARs, by enabling both vertical and horizontal LiDARs as inputs, with the idea
of helping determining vertical motions.
Our system (recall Figure~\ref{fig:lo-module} and the default pipeline in Figure~\ref{fig:lidar3d.obs.pipeline})
then waits for two consecutive scans before populating the observation metric map, taking into account the time offset between 
both scans to accurately de-skew both in a spatially coherent way. The rest remains exactly the same for the ``MOLA-LO (2 LiDARs)'' experiment
shown in the table. As can be seen, using this vertical scanner indeed improves accuracy, but it is still far from LIO techniques.
Note that all these results for MOLA-LO use the same configuration than in all other automotive datasets, 
with an automatic determination of the minimum points range to be filtered to prevent observations of the vehicle body. 
In this particular dataset, the smaller size of the drone makes that these rules actually prevent using valid information about nearby objects,
but we did not manually tune this filter to ensure all datasets are run with the same configuration.
Then, in a second experiment we modified the default pipeline in Figure~\ref{fig:lidar3d.obs.pipeline} to create two local maps:
``near map'' and ``far map'', for nearby and distant points, respectively. 
ICP pipeline blocks are then configured to search for potential pairings between nearby LiDAR scan points
and the ``near map'', and between all LiDAR points and the ``far map'' (interested readers can check the details in the 
configuration YAML file \texttt{lidar3d-near-far.yaml} available online). This configuration corresponds to the last row in the table.
As can be seen, this method now has errors in the order of magnitude of LIO techniques, even surpassing LIOSAM on some particular sequences.
Finally, using a smarter sampling of feature points would probably further improve the results, but this point is left for future works.

\subsection{HILTI 2021 dataset}
\label{sect:hilti}

This dataset, presented in \cite{hilti2021}, comprises several challenging
sequences indoors and mixed indoor-outdoor scenarios recorded with 
an Ouster OS0-64 LiDAR. From the published sequences, only two of them have
full SE(3) ground truth. From those two, we selected the largest sequence (the Drone Testing Area)
for benchmarking our LO system. 
The resulting trajectories are illustrated in Figure~\ref{fig:hilti2021.paths}.
Quantitatively, we measured the rmse ATE for MOLA-LO against ground truth 
obtaining 0.18~m, slightly better than KISS-ICP \citep{vizzo2023kiss} which obtains an ATE of 0.21~m.
It is noteworthy that our system fails to converge 
for some of the other sequences in this dataset 
(whose ground truth is not publicly released),
where feature-extraction or multi-modality seem essential for robustly 
cope with scenarios like narrow, featureless indoor spaces.

\begin{table}
\caption{Absolute translational error (ATE) (rmse values in meters as reported by ``\texttt{evo\_ape -a}'')
for the \textbf{Voxgraph} dataset. Bold means best accuracy.
}
\centering
{
\setlength\extrarowheight{1pt}
\begin{tabular}{|c||r|}
\hline
\textbf{Method}
& \TwoRows{Voxgraph}{(Sequence: \texttt{t0})}
\\
\hhline{|-|*{1}{-|}}
\TwoRows{LOAM}{\cite{zhang2017low}} &  2.64~m \\  \hline
\TwoRows{Voxgraph}{\cite{reijgwart2019voxgraph}} &  0.83~m \\  \hline
\TwoRows{KISS-ICP}{\cite{vizzo2023kiss}} &  0.253~m \\  \hline
\TwoRows{MOLA-LO (default)}{(ours)} & 0.265~m \\ \hline
\TwoRows{\REVIEW{MOLA-LO (3D-NDT)}}{(ours)} & \textbf{0.250}~m \\ \hline
\end{tabular}
}
\label{tab:voxgraph.metrics}
\end{table}

\begin{figure}
\centering
\includegraphics[width=1.0\columnwidth]{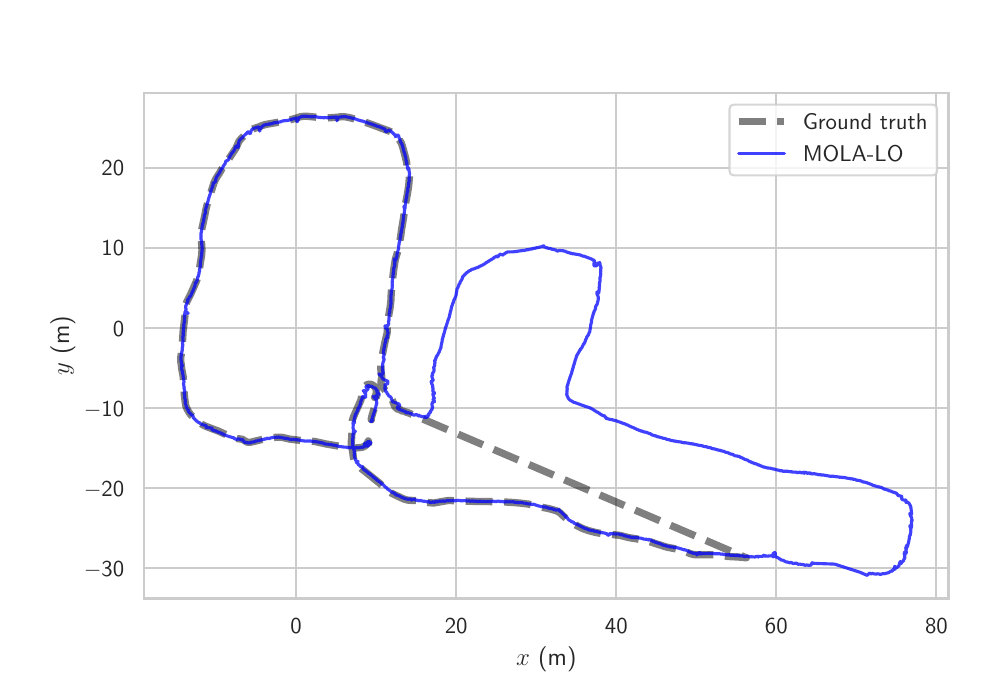}
\caption{Estimated trajectory by our LO system for the \textbf{VoxGraph} dataset, together with the (partial) ground truth. See discussion in Section~\ref{sect:voxgraph}}
\label{fig:voxgraph.path}
\end{figure}

\subsection{Voxgraph dataset}
\label{sect:voxgraph}

This aerial dataset, presented in \cite{reijgwart2019voxgraph}, 
comprises one sequence of a drone flight outdoors, equipped with an Ouster OS1-64 LiDAR
and featuring RTK-based ground truth.
We compare the ATE metric achieved by several LO SOTA methods in 
Table~\ref{tab:voxgraph.metrics}, where it can be seen that both KISS-ICP and ours achieve a significant 
better accuracy, with KISS-ICP having a small advantage in this case.
The estimated trajectory for our method and the partially-available ground truth are illustrated in 
Figure~\ref{fig:voxgraph.path}. Note that ATE metrics have been only evaluated for those segments of the trajectory
with corresponding ground truth. 
As in the HILTI Drone Testing Arena sequence (Section~\ref{sect:hilti}), loop closure is not required due to the small size
of the environment, hence we only evaluate LO methods.

\subsection{DARPA Subterranean dataset}
\label{sect:darpa}

We now evaluate our LO method against the dataset 
presented in \cite{tranzatto2022team} by Team Cerberus, 
winners of the subterranean 2021 DARPA challenge.
The public dataset comprises
3D LiDAR, vision, and IMU readings from 
four legged robots as they faced the final event of the challenge.
Robots in these sequences alternate periods of inactivity with walking as they
autonomously explore a series of cave-like scenarios. 
ANYmal C legged robots were used, equipped with one Velodyne VLP-16 each.
We only used LiDAR scans from these dataset, which contains four sequences, one per robot, 
and ground truth trajectories extracted by scan matching against a ground truth environment point cloud
from survey-quality scanners.

Since loop closures are not relevant in these scenarios, only LO methods were evaluated.
Two SOTA representative methods, KISS-ICP \citep{vizzo2023kiss} and SiMpLE \citep{bhandari2024minimal}, 
have been run side by side with our system by using MOLA wrapper modules to ensure identical conditions
for input data. KISS-ICP was run with its default parameters, just 
like system, run with its default configuration and pipelines (Section~\ref{sect:icp.pipeline.default}).
In turn, for SiMpLE we used 
its slowest and most accurate configuration (``offline''), already evaluted for KITTI in Section~\ref{sect:kitti}.

\begin{figure*}
\centering
\begin{tabular}{cc}
\subfigure[\texttt{anymal\_1} -- 3D trajectory]{\includegraphics[width=0.28\textwidth]{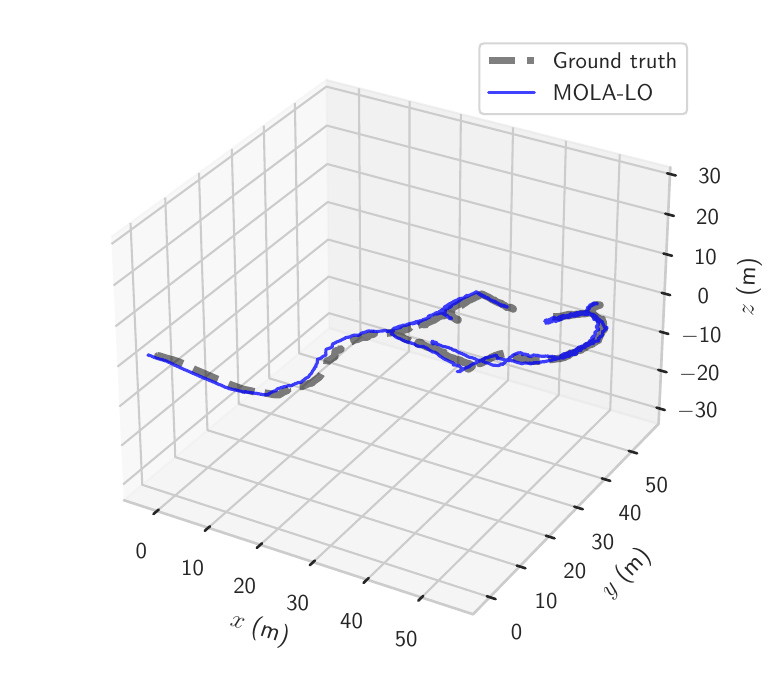}}
&
\subfigure[\texttt{anymal\_1} -- Trajectory components]{\includegraphics[width=0.66\textwidth]{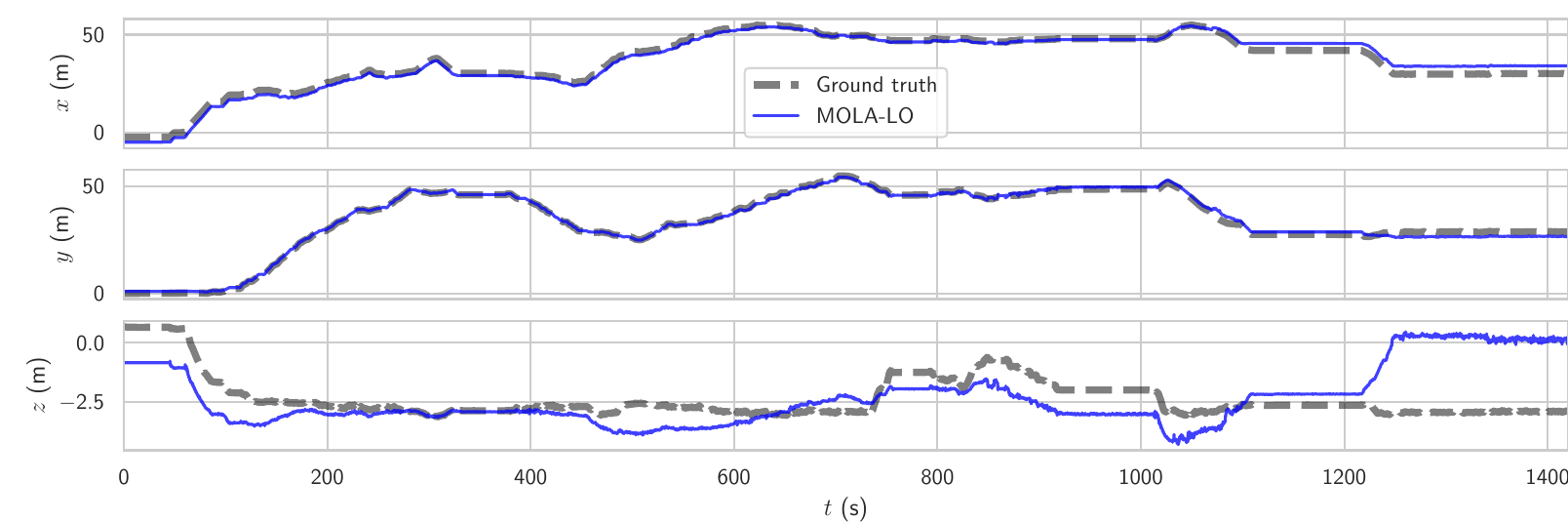}}
\\
\subfigure[\texttt{anymal\_2} -- 3D trajectory]{\includegraphics[width=0.28\textwidth]{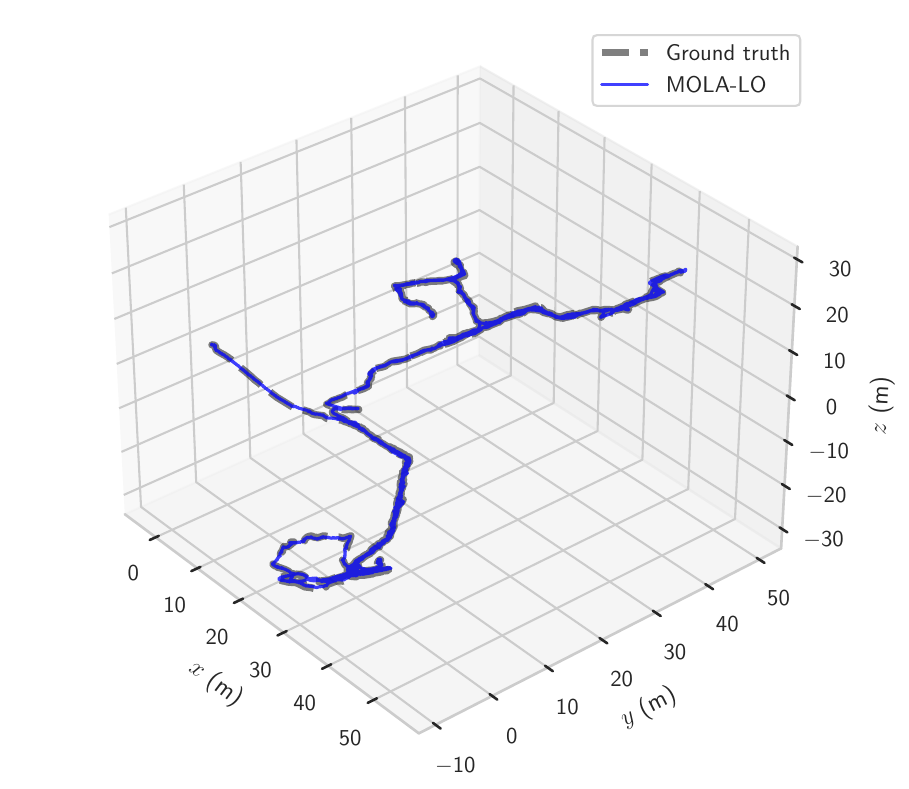}}
&
\subfigure[\texttt{anymal\_2} -- Trajectory components]{\includegraphics[width=0.66\textwidth]{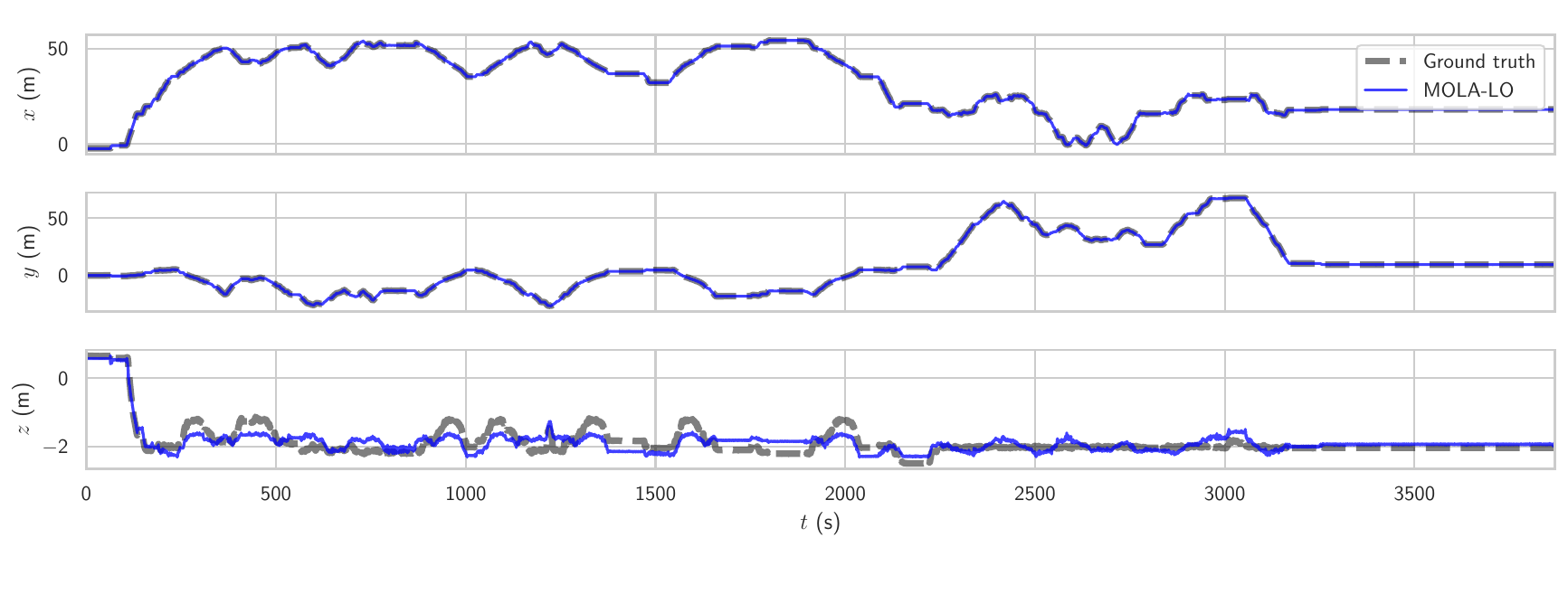}}
\\
\subfigure[\texttt{anymal\_3} -- 3D trajectory]{\includegraphics[width=0.28\textwidth]{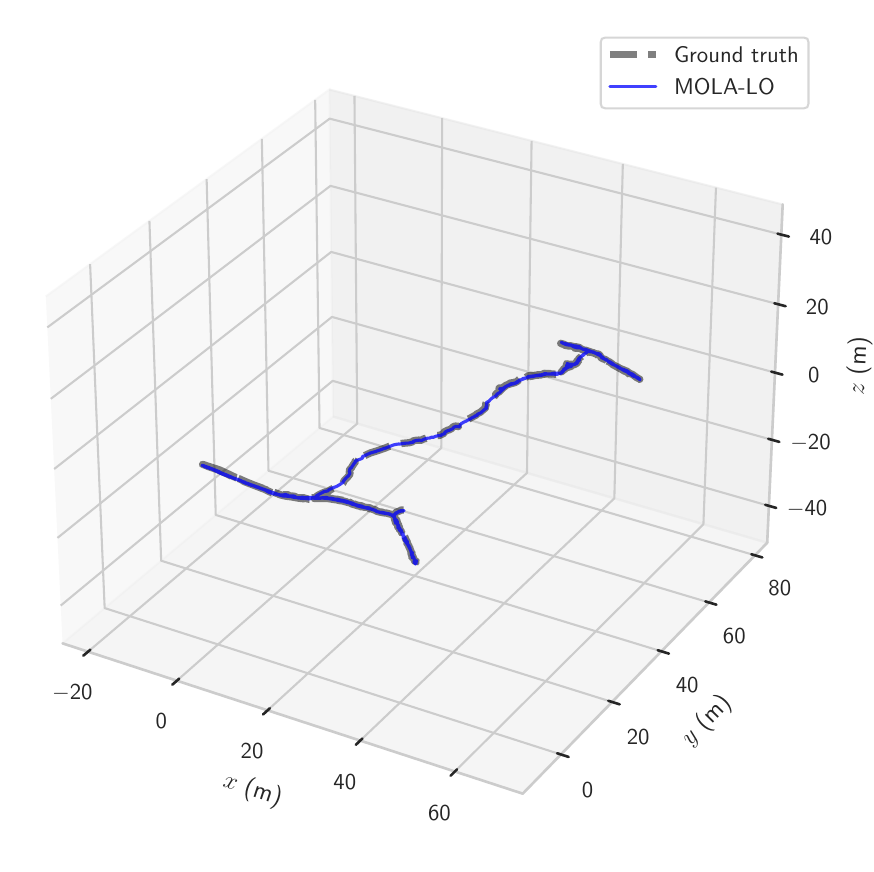}}
&
\subfigure[\texttt{anymal\_3} -- Trajectory components]{\includegraphics[width=0.66\textwidth]{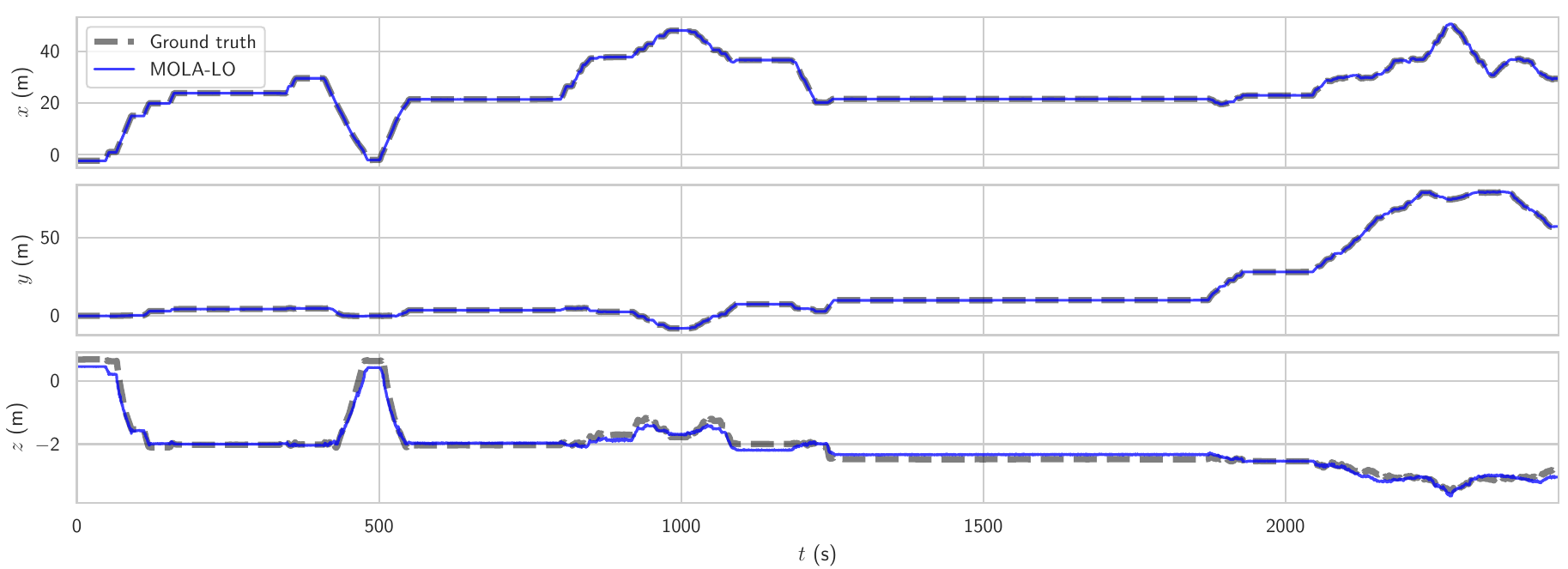}}
\\
\subfigure[\texttt{anymal\_4} -- 3D trajectory]{\includegraphics[width=0.28\textwidth]{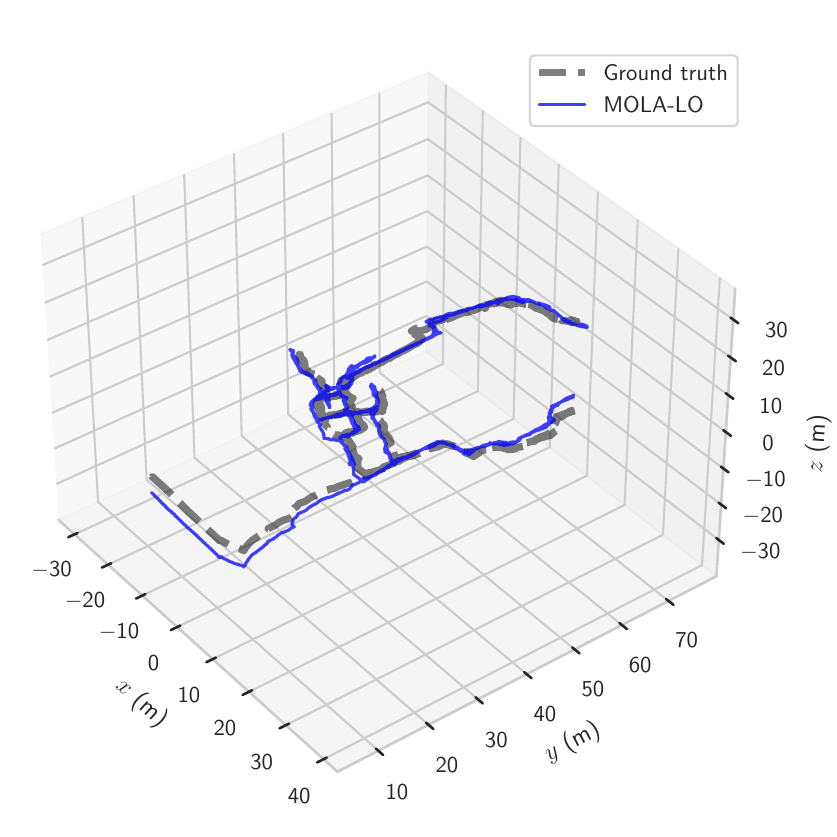}}
&
\subfigure[\texttt{anymal\_4} -- Trajectory components]{\includegraphics[width=0.66\textwidth]{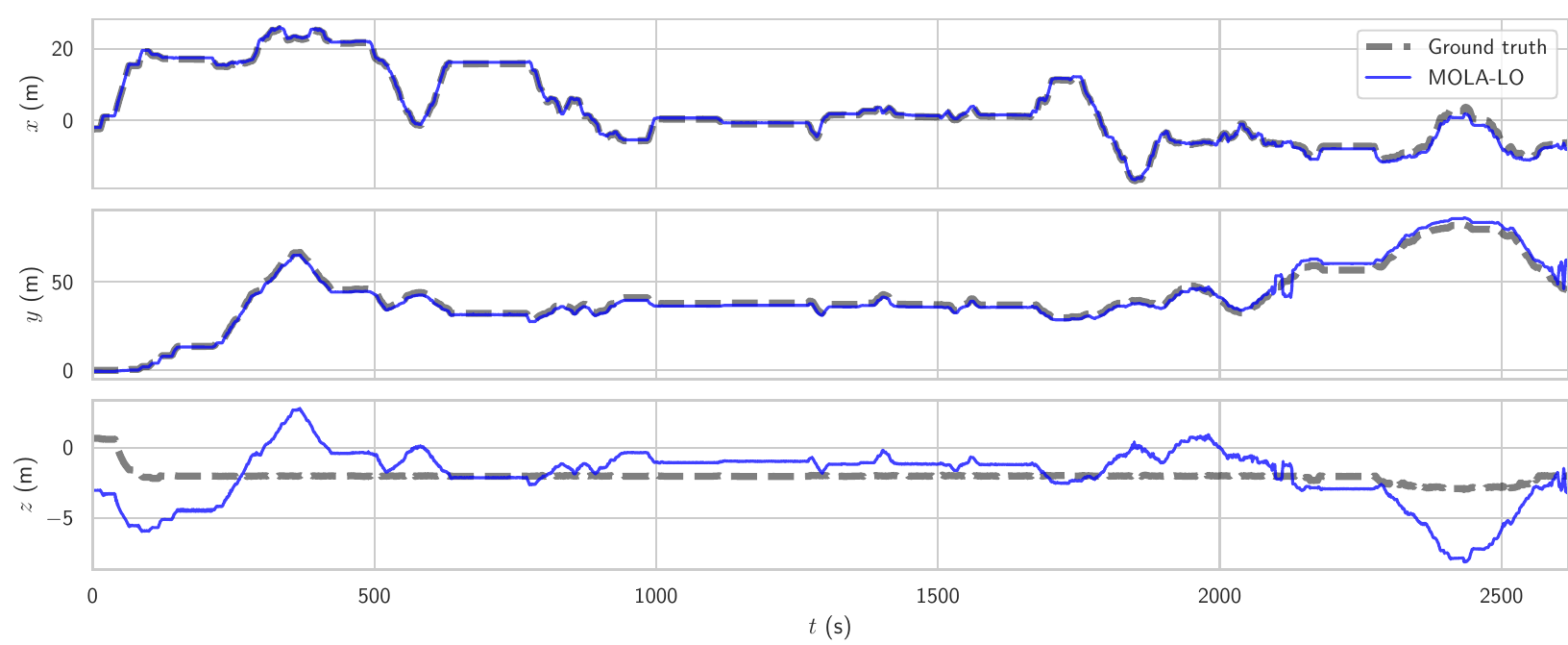}}
\end{tabular}

\caption{Estimated trajectories for the proposed LiDAR odometry system ("MOLA-LO") applied to the \textbf{DARPA Subterranean Final Event} dataset,
compared to ground truth. See discussion in Section~\ref{sect:darpa}.
}
\label{fig:darpa.paths}
\end{figure*}

The quantitative results are summarized in Table~\ref{tab:darpa.metrics}. KISS-ICP diverges in all sequences.
SiMpLE diverges in the \texttt{anymal\_4} sequence, achieving good accuracy in the others.
Our system does not diverge in any of the sequences, performs faster than SiMpLE, and achieves better ATE
trajectory metrics in 3 out of 4 sequences, demonstrating its robustness and viability for non-automotive, unstructured scenarios. 
As can be seen in the 3D trajectory views and component analysis in Figure~\ref{fig:darpa.paths},
our method reconstructs accurate trajectories, with the only weak point worth mentioning being 
the drift accumulated in the vertical component ($z$ axis), especially for sequences \texttt{anymal\_1} and \texttt{anymal\_4}; 
in fact, the largest part of RMSE ATE for those sequences comes from such $z$ component.
Smarter selection of down-sampled points would probably alleviate this drift, a topic that shall be studied in future works.

\begin{table*}
\caption{Absolute translational error (ATE) (rmse values in meters as reported by ``\texttt{evo\_ape -a}'')
for the \textbf{DARPA Subterranean Final Event} dataset.
Bold means best accuracy in each sequence.
Divergence is shown as $\times$.}
\centering
\begin{small}
{
\setlength\extrarowheight{1pt}
\begin{tabular}{|c||r|r|r|r|r|}
\hline
\textbf{Method}
& \texttt{anymal\_1}
& \texttt{anymal\_2}
& \texttt{anymal\_3}
& \texttt{anymal\_4}
& \textbf{Avr. time}
\\
\hline
{KISS-ICP}{ \cite{vizzo2023kiss}} & 
\TwoRows{$\times$}{(88.57~m)} & \TwoRows{$\times$}{(4918.14~m)} & \TwoRows{$\times$}{(2148.04~m)} & \TwoRows{$\times$}{(22.45~m)} &
1.4~ms
\\
\hline
{SiMpLE (offline)}{ \cite{bhandari2024minimal}} & 
\textbf{0.44}~m & 0.48~m & 0.28~m & \TwoRows{$\times$}{(18.46~m)} &
54.2~ms
\\
\hline
{MOLA-LO}{ (ours)} & 
2.80~m & \textbf{0.24}~m & \textbf{0.16}~m & \textbf{3.15}~m & 
13.1~ms
\\ %
\hline
\end{tabular}
}
\end{small}
\label{tab:darpa.metrics}
\end{table*}

\begin{figure*}
\centering
{
\begin{tabular}{ccc}
\subfigure[\texttt{01} (``short'')]{\includegraphics[width=0.32\textwidth]{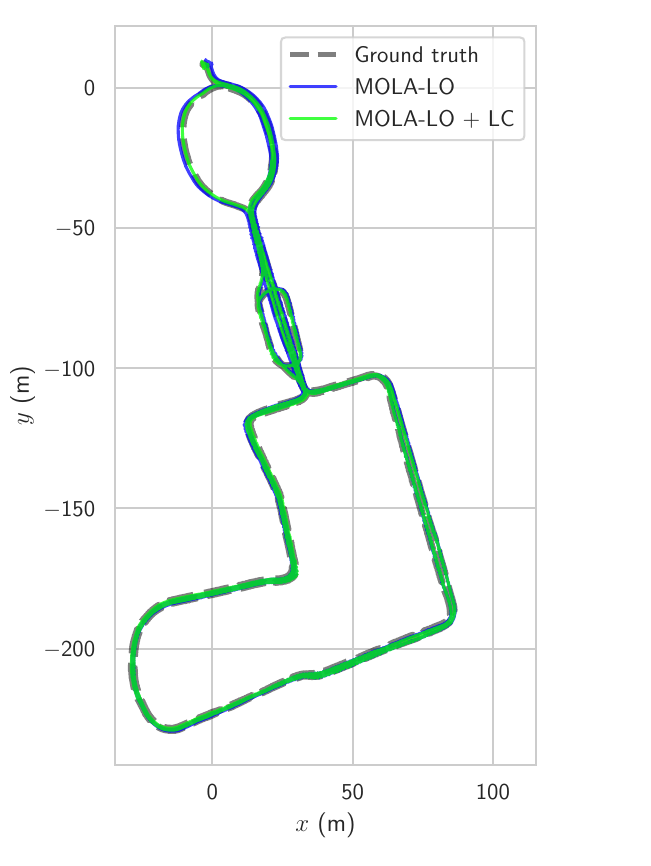}}
&
\subfigure[\texttt{02} (``long'')]{\includegraphics[width=0.32\textwidth]{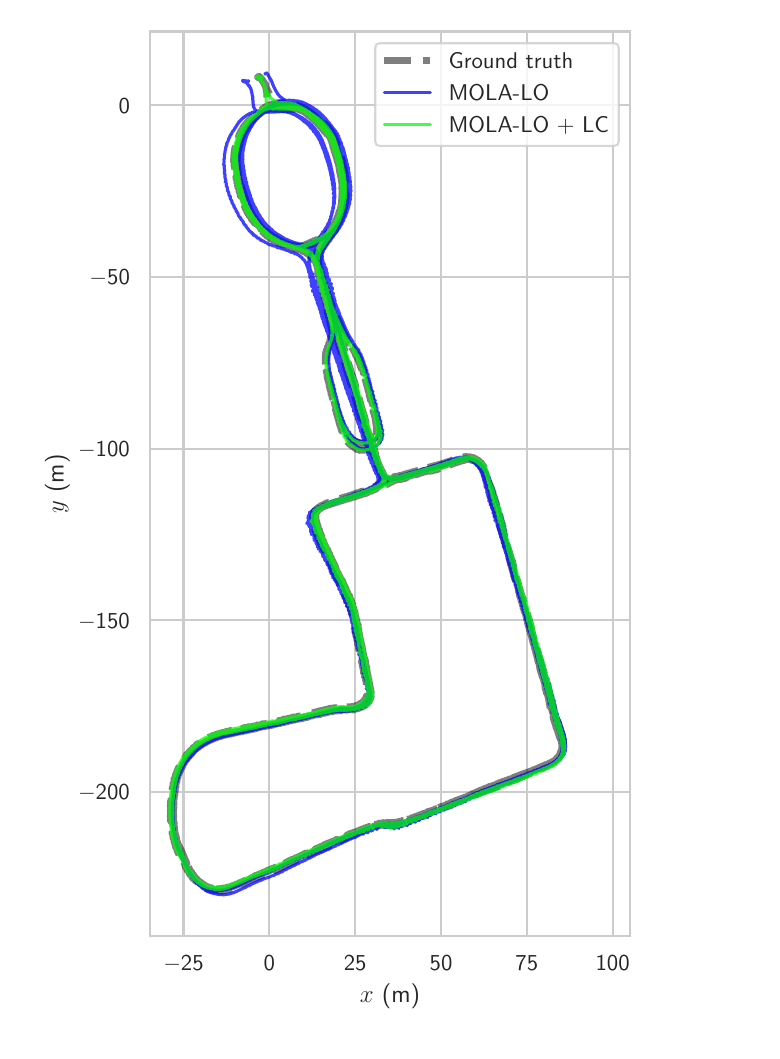}}
&
\subfigure[Cloister]{\includegraphics[width=0.32\textwidth]{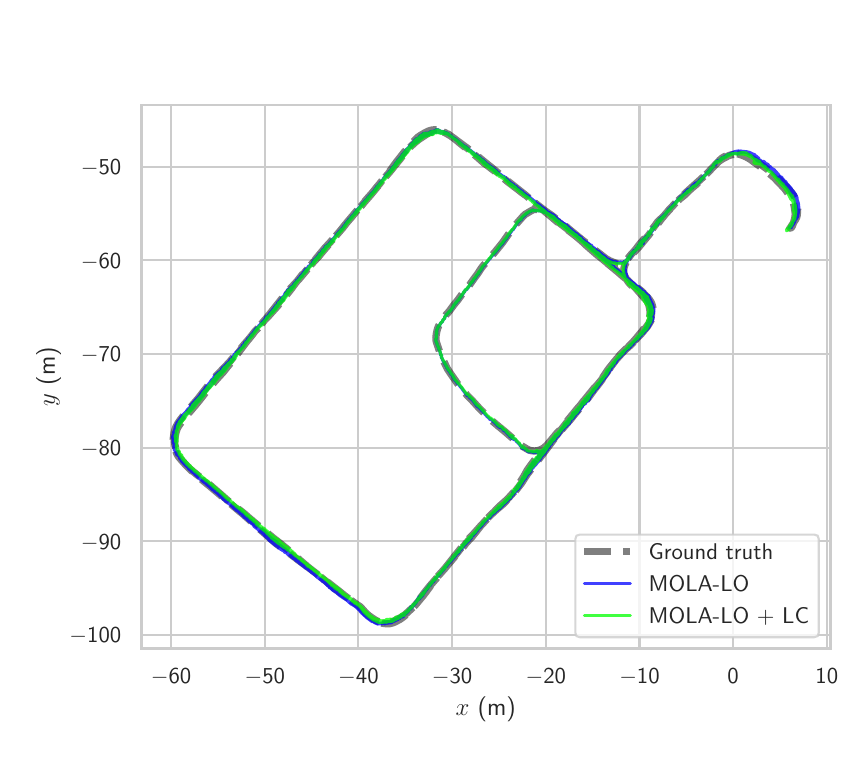}}
\end{tabular}
\begin{tabular}{cc}
\subfigure[Park]{\includegraphics[width=0.40\textwidth]{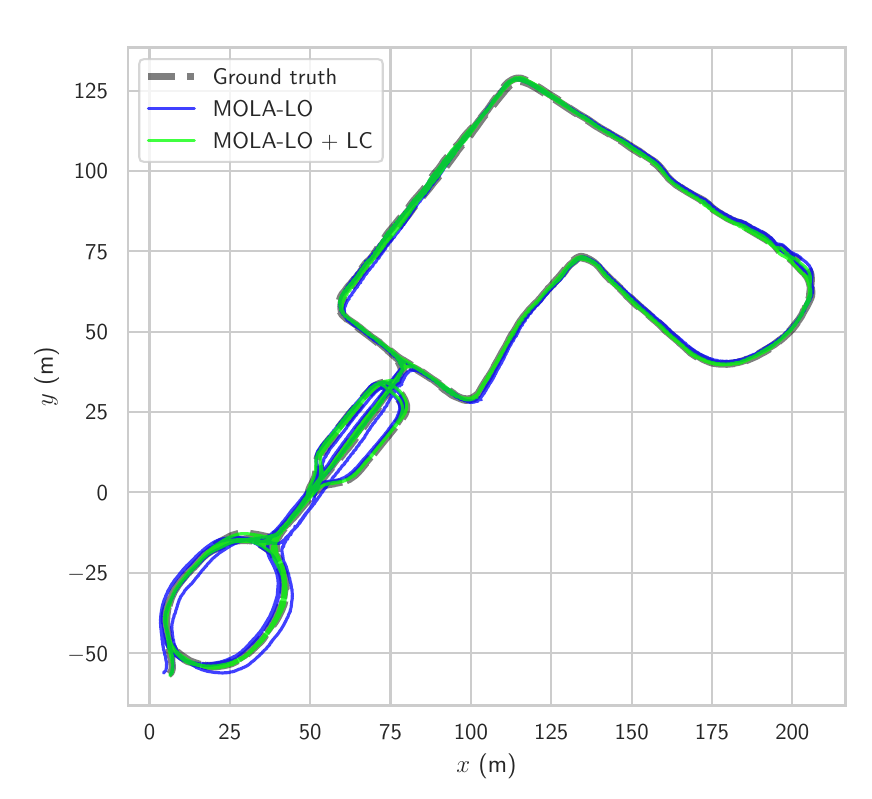}}
&
\subfigure[Stairs]{\includegraphics[width=0.40\textwidth]{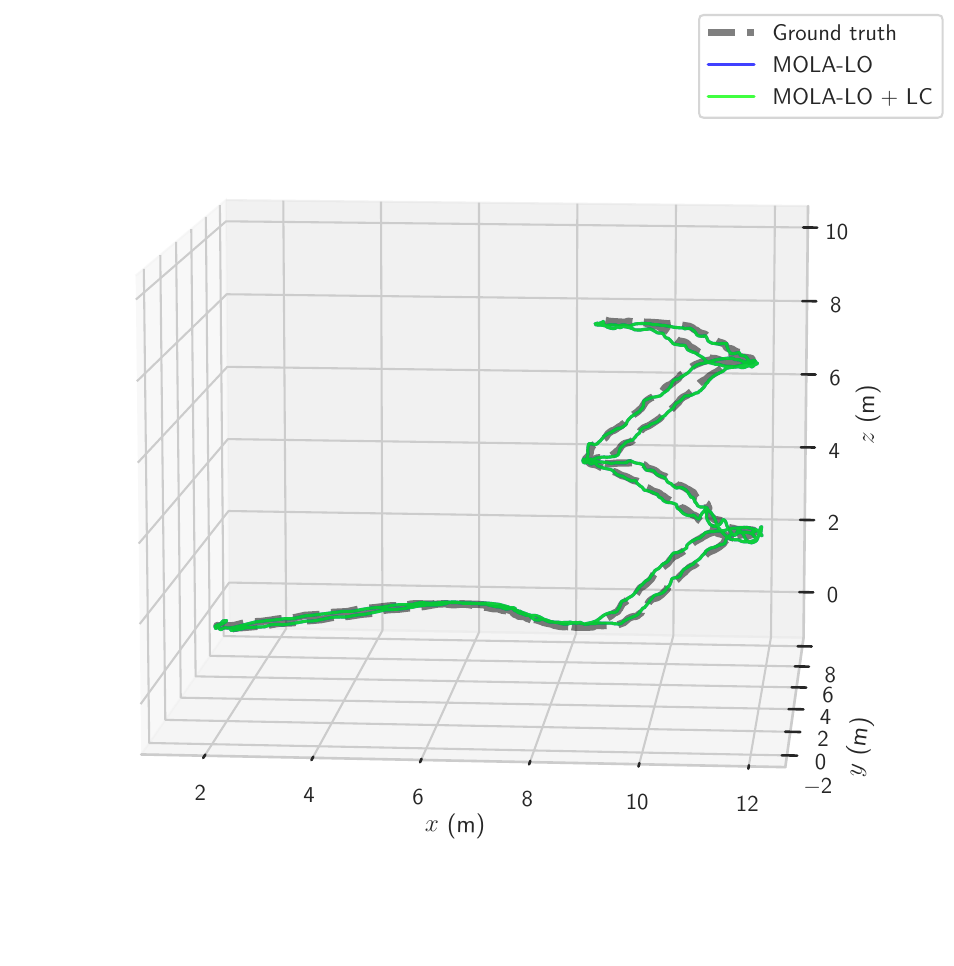}}
\end{tabular}
}
\caption{Estimated trajectories for the proposed LiDAR odometry system ("MOLA-LO") applied to the \textbf{NewerCollege dataset}, compared to ground truth. Trajectories denoted as ``MOLA-LO + LC'' include the post-processing loop-closure
stage discussed in section~\ref{sect:lc}.
Continues in Figure~\ref{fig:ncd.paths2}. See discussion in Section~\ref{sect:ncd}.
}
\label{fig:ncd.paths}
\end{figure*}

\begin{figure*}
\centering
{
\begin{tabular}{ccc}
\subfigure[Maths Easy]{\includegraphics[width=0.32\textwidth]{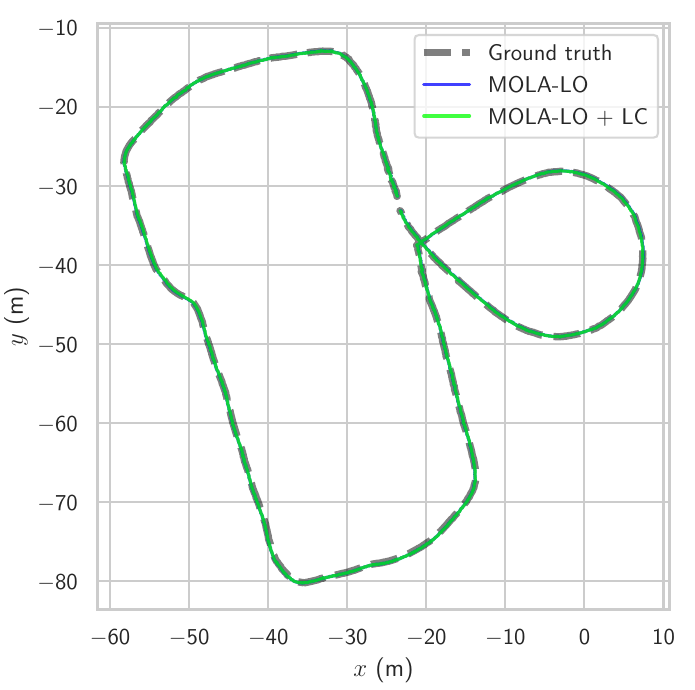}}
&
\subfigure[Maths Medium]{\includegraphics[width=0.32\textwidth]{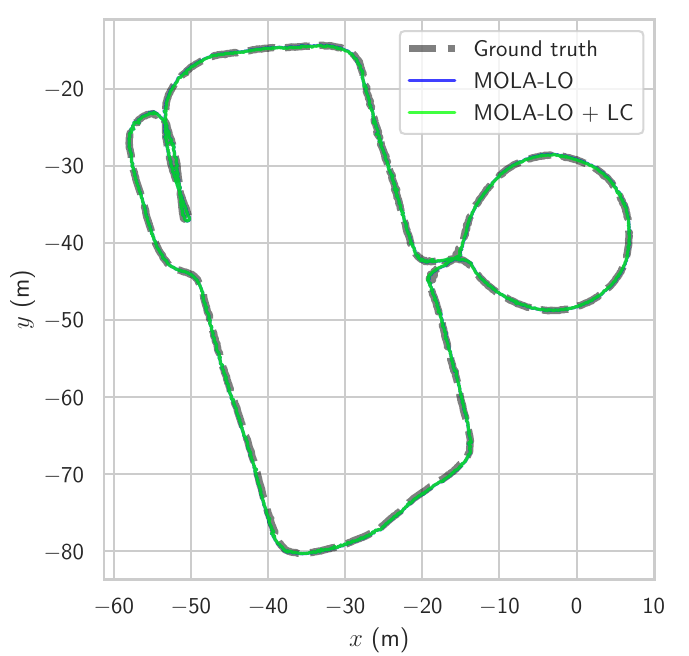}}
&
\subfigure[Maths Hard]{\includegraphics[width=0.32\textwidth]{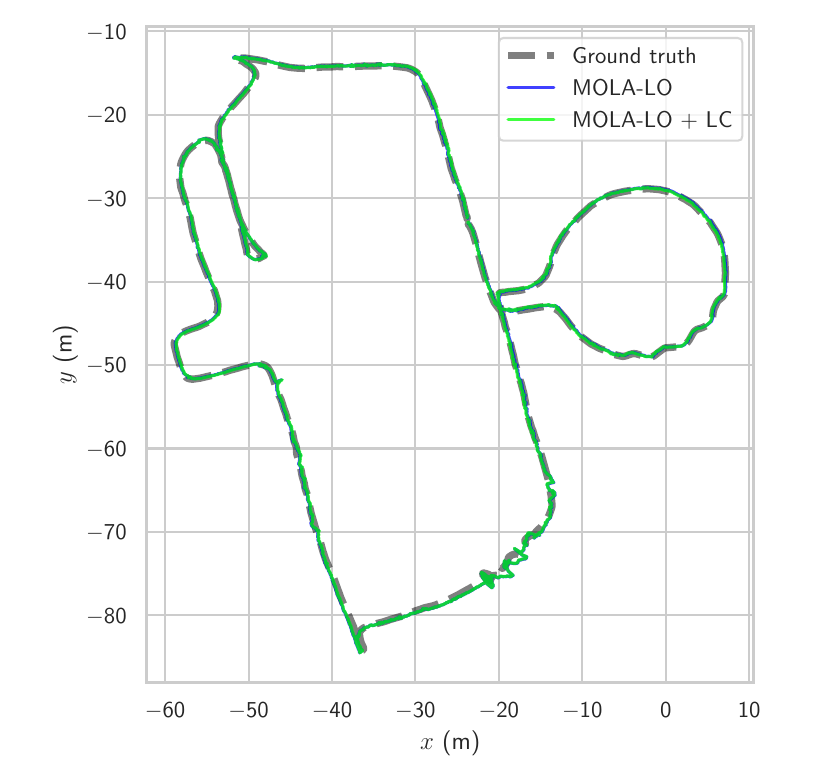}}
\end{tabular}
\begin{tabular}{ccc}
\subfigure[Quad Easy]{\includegraphics[width=0.32\textwidth]{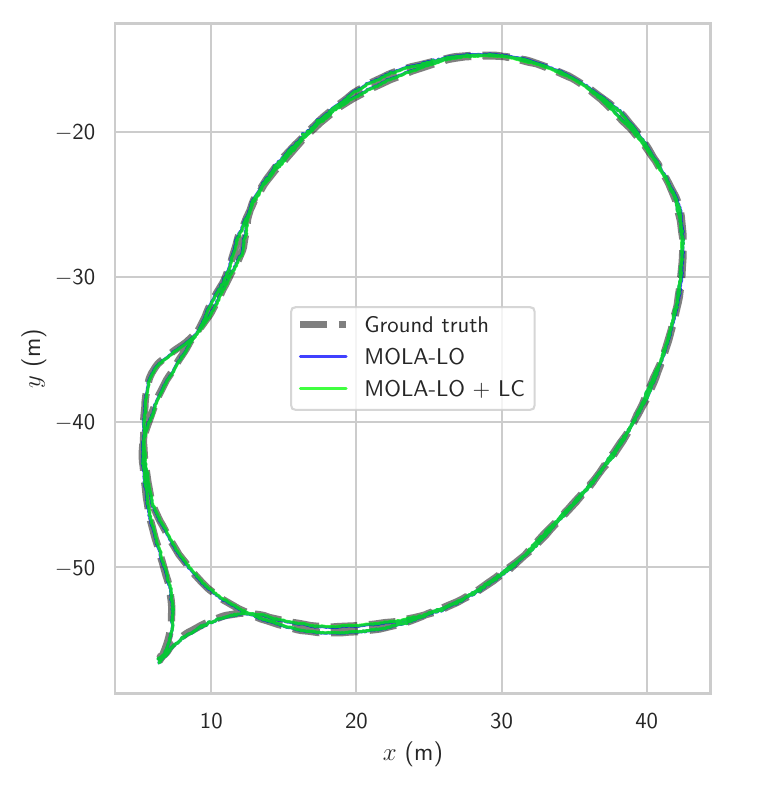}}
&
\subfigure[Quad Medium]{\includegraphics[width=0.32\textwidth]{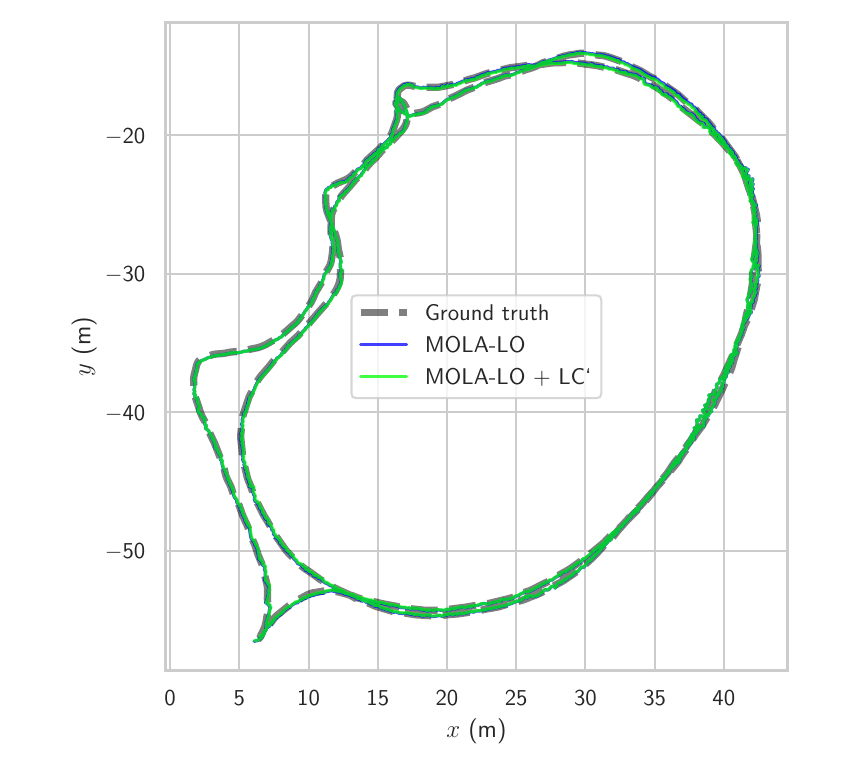}}
&
\subfigure[Quad Hard]{\includegraphics[width=0.32\textwidth]{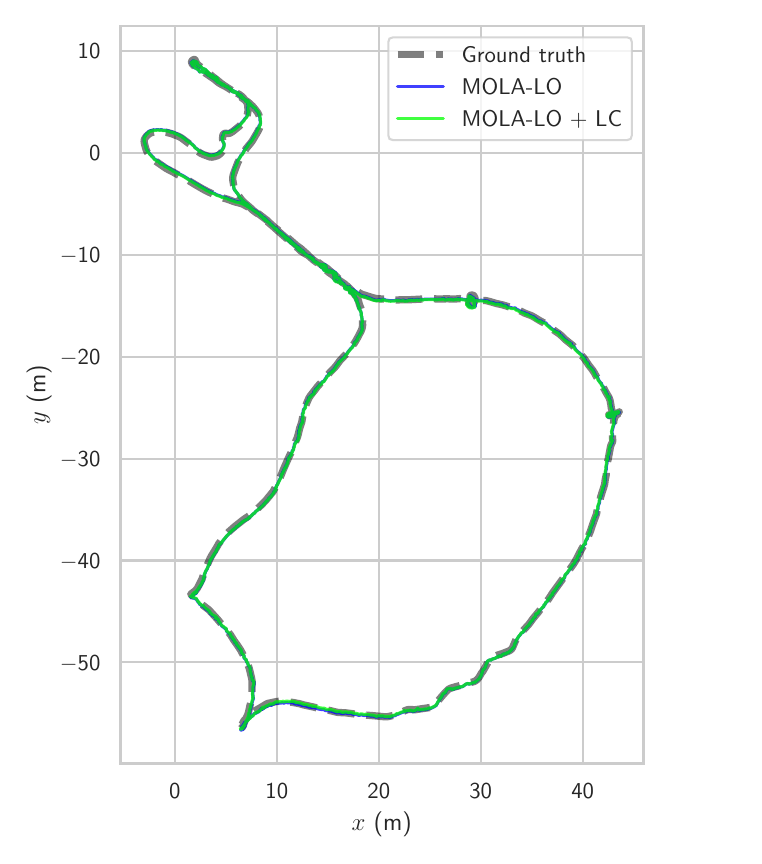}}
\end{tabular}
\begin{tabular}{ccc}
\subfigure[Underground Easy]{\includegraphics[width=0.32\textwidth]{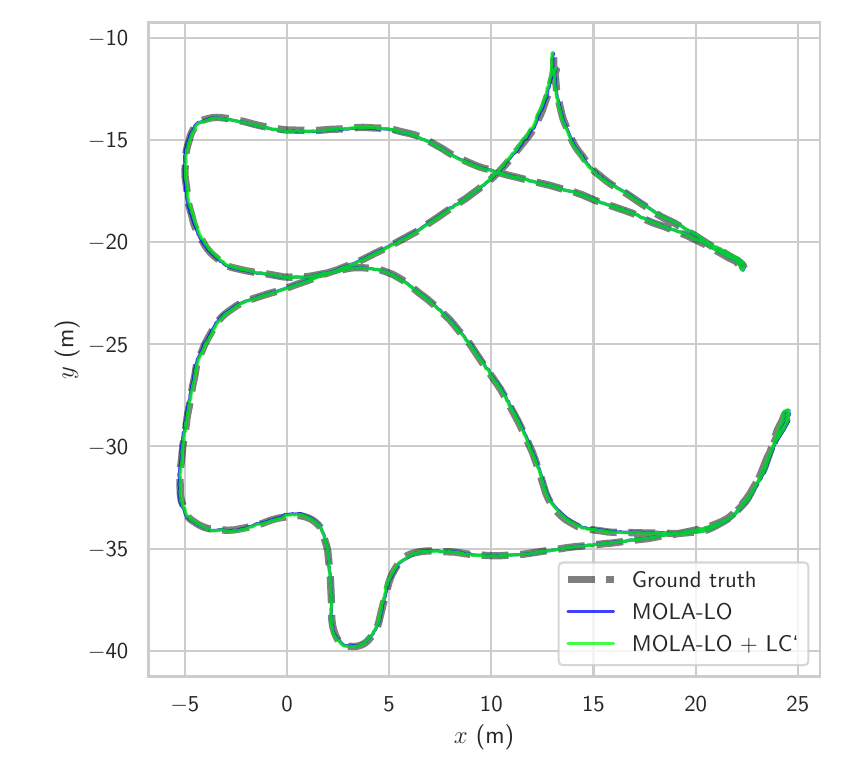}}
&
\subfigure[Underground Medium]{\includegraphics[width=0.32\textwidth]{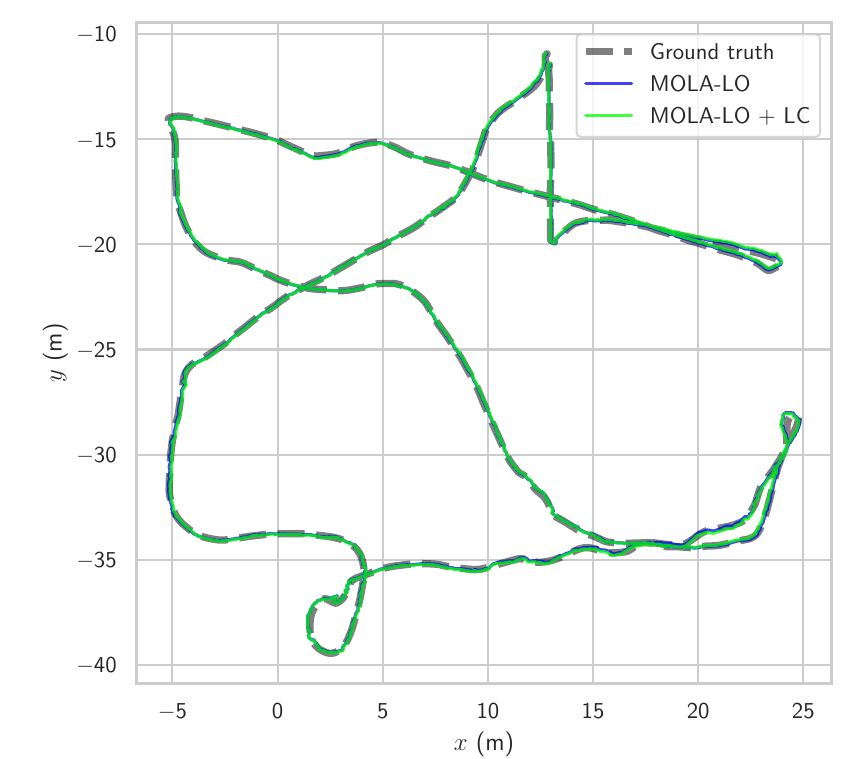}}
&
\subfigure[Underground Hard]{\includegraphics[width=0.32\textwidth]{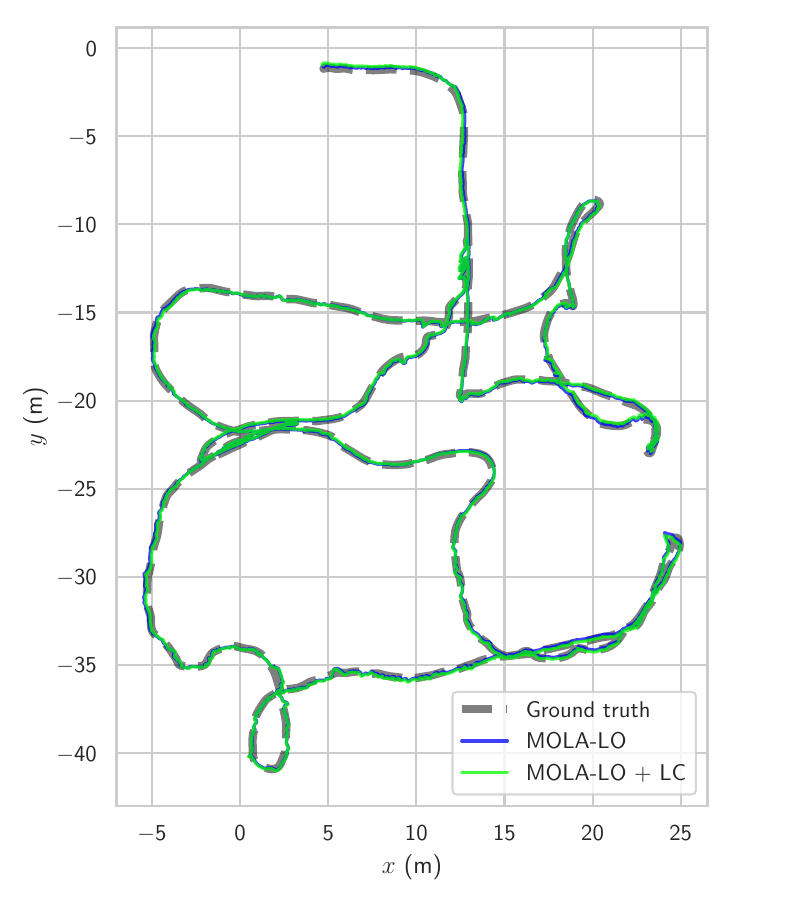}}
\end{tabular}
}
\caption{Continuation of Figure~\ref{fig:ncd.paths} with more results for the \textbf{Newer College dataset}. See discussion in Section~\ref{sect:ncd}.
}
\label{fig:ncd.paths2}
\end{figure*}

\begin{table*}
\caption{Absolute translational error (ATE) (rmse values in meters as reported by ``\texttt{evo\_ape -a}'')
for the \textbf{NewerCollege dataset}.
Bold means best accuracy in each sequence and category.
Divergence is shown as $\times$.
\REVIEW{The version of MOLA-LO that always updates the local map is discussed
in Section~\ref{sect:ablation.localmap.updates}.}
}
\centering
\resizebox{\textwidth}{!}
{
\setlength\extrarowheight{1pt}
\begin{tabular}{|c||r|r|r|r|r|r|r|r|r|r|r|r|r|r|}
\hline
\textbf{Method}
& \textbf{01}
& \textbf{02}
& \textbf{Cloister}
& \TwoRowsBf{Maths}{Easy}
& \TwoRowsBf{Maths}{Medium}
& \TwoRowsBf{Maths}{Hard}
& \textbf{Park}
& \TwoRowsBf{Quad}{Easy}
& \TwoRowsBf{Quad}{Medium}
& \TwoRowsBf{Quad}{Hard}
& \TwoRowsBf{Undergr.}{Easy}
& \TwoRowsBf{Undergr.}{Medium}
& \TwoRowsBf{Undergr.}{Hard}
& \textbf{Stairs}
\\
\hhline{|-|*{14}{-|}}
\TwoRows{KISS-ICP}{\cite{vizzo2023kiss}} & 
\textbf{0.61}~m & 1.78~m & 0.95~m & 0.07~m & \textbf{0.12}~m & 
\TwoRows{$\times$}{(29.6~m)} & {1.54}~m & 0.10~m & 0.19~m & 0.65~m & 0.26~m & 0.45~m & \TwoRows{$\times$}{(7.98~m)} & \TwoRows{$\times$}{(3407~m)}
\\
\hline
\TwoRows{MOLA-LO}{(ours)} & 
{0.68}~m & \textbf{0.54}~m & \textbf{0.12}~m & \textbf{0.06}~m & \textbf{0.12}~m & \textbf{0.11}~m & \textbf{0.90}~m & \textbf{0.08}~m & \textbf{0.08}~m & \textbf{0.12}~m & \textbf{0.07}~m & \textbf{0.08}~m & \textbf{0.09}~m & \textbf{0.11}~m
\\ %
\hline
\ThreeRowsC{\REVIEW{MOLA-LO}}{\REVIEW{(Always updates}}{\REVIEW{local map)}} & 
\TwoRows{$\times$}{(5.11~m)} & \TwoRows{$\times$}{(84.37~m)} & 0.39~m & 0.09~m & \TwoRows{$\times$}{(4.93~m)} & 0.78~m & 1.38~m & 0.08~m & 2.99~m & 0.10~m & 0.08~m & 0.46~m & \TwoRows{$\times$}{(8.06~m)} & 0.34~m \\
\hhline{|*{15}{=|}}
\TwoRows{MOLA-LO + LC}{(ours)} &
0.31~m & 0.40~m & 0.19~m & 0.09~m & 0.13~m & 0.31~m & 0.31~m & 0.09~m & 0.12~m & 0.16~m & 0.13~m & 0.13~m & 0.25~m & 0.42~m 
\\
\hline
\end{tabular}
}
\label{tab:ncd.metrics}
\end{table*}

\begin{figure*}
\centering
\includegraphics[width=0.99\textwidth]{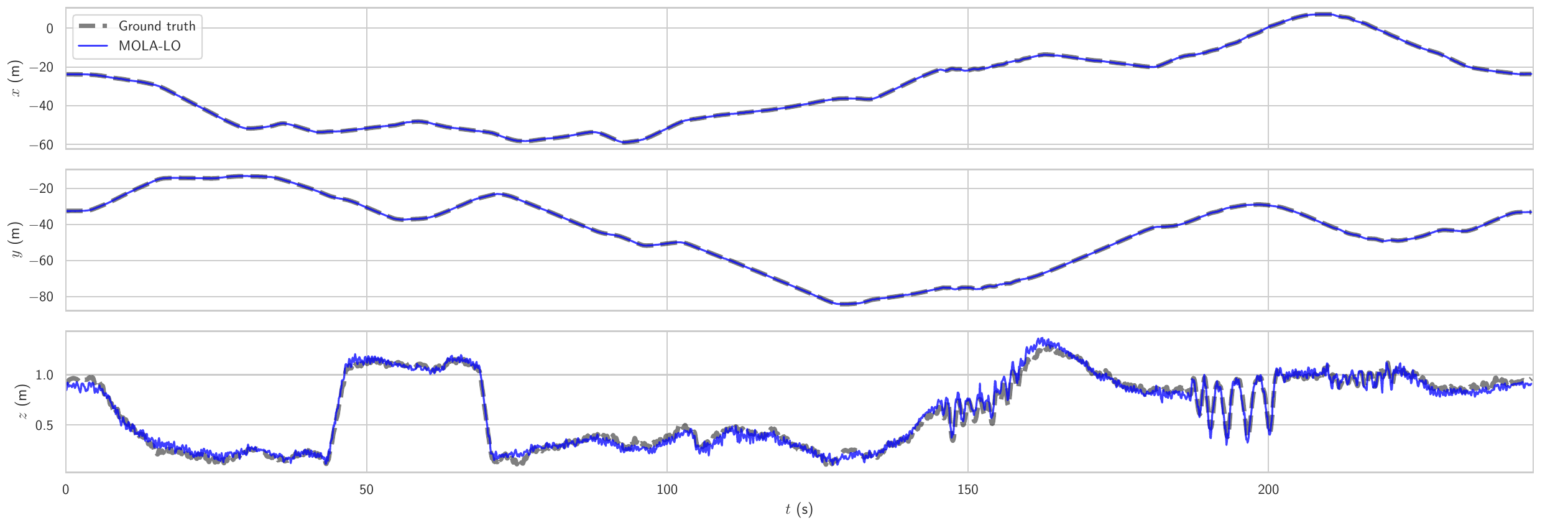}
\includegraphics[width=0.99\textwidth]{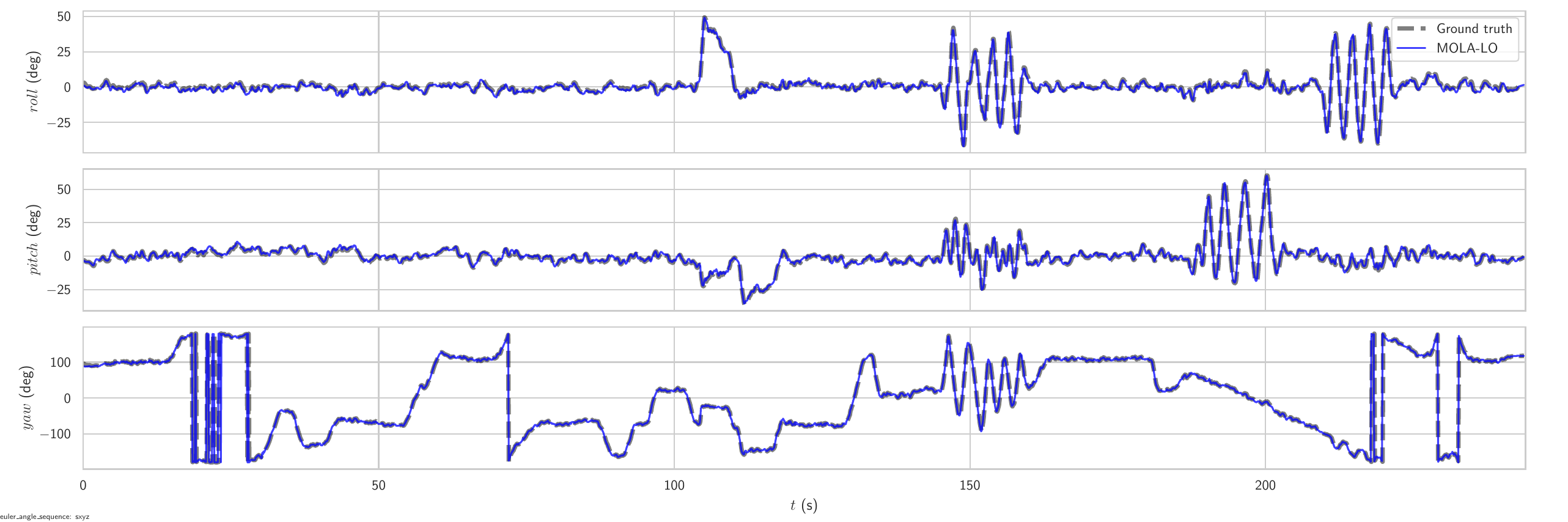}
\caption{Detailed view of the Maths Hard (MH) sequence (\textbf{Newer College dataset}) in Figure~\ref{fig:ncd.paths2}. 
Observe the highly dynamic rotations imposed to the sensor in all six degrees of freedom with the intention of making this sequence a challenge for LO methods, and how our method keeps its estimation close to ground truth at all times. See discussion in Section~\ref{sect:ncd}.}
\label{fig:ncd.mh.details}
\end{figure*}

\subsection{Newer College dataset}
\label{sect:ncd}

This dataset consists of several sequences acquired in New College (Oxford) with 
a handheld device comprising a 3D LiDAR, a high-frequency IMU, and cameras, 
with the intention of becoming a benchmark for visual-inertial odometry
and LiDAR-inertial odometry techniques.
The first two sequences (\texttt{01} and \texttt{02}) used an Ouster OS0-64 
and were presented in \cite{ramezani2020newer},
with walks covering the college Great Quad, Garden Quad, and the Bowling Green, 
long enough to justify the use of a full SLAM solution with loop closures.
Later on, 12 additional sequences using an Ouster OS1-128 sensor were presented 
in \cite{zhang2021multi}, with more diverse scenarios and difficulty levels, 
organized by intentionally-introduced high dynamic translations and rotations
to create challenging conditions.

Our LO method has been evaluated against the SOTA method KISS-ICP \citep{vizzo2023kiss},
with quantitative results summarized in Table~\ref{tab:ncd.metrics}, 
where it can be seen that our system has a better accuracy in 11 out of the 14 sequences,
8 of them with ATE reductions larger than 50\%.
More over, KISS-ICP diverges in 3 of the hardest sequences (``Maths Hard'' (MH), ``Underground Hard'' (UH), ``Stairs'' (ST))
while MOLA-LO achieves ATE better than 0.12~m in all of them. 
It is worth mentioning that none of the two benchmarked LO methods
make use of the IMU data. 
Other works, like \cite{pfreundschuh2024coinlio},
have recently reported even better accuracy 
in novel methods exploiting the intensity channel of LiDAR points together with inertial measurements,
but for the sake of fair benchmarking we limit the present comparison to methods relying solely on
3D points.

Trajectories obtained by our LO method, and by the loop-closure post-processing stage, 
are illustrated in Figures~\ref{fig:ncd.paths}--\ref{fig:ncd.paths2}, together with ground truth.
In order to understand why the sequences labeled as ``hard'' are harder to cope with,
we show the translational ($x$,$y$,$z$) and orientation (yaw, pitch, roll) components
of the trajectory estimated for one such sequence (``Maths hard''), along with ground truth,
in Figure~\ref{fig:ncd.mh.details}. 
Observe how, in time steps corresponding to $t\approx 140~s$, $t\approx 180~s$, and $t\approx 225~s$, 
the sensor is rotated at a high speed in different combinations of orientations and translations. 
The introduction of the joint estimation of the velocity vector within the ICP loop itself
(see Section~\ref{sect:icp.local.map.update})
has revealed essential to cope with these sudden disruptions of the 
constant velocity assumption.

\subsection{UAL VLP-16 campus dataset}
\label{sect:ual.campus}

\begin{figure*}
\centering
{
\begin{tabular}[t]{c}
\subfigure[Trajectories]{\includegraphics[width=0.80\textwidth]{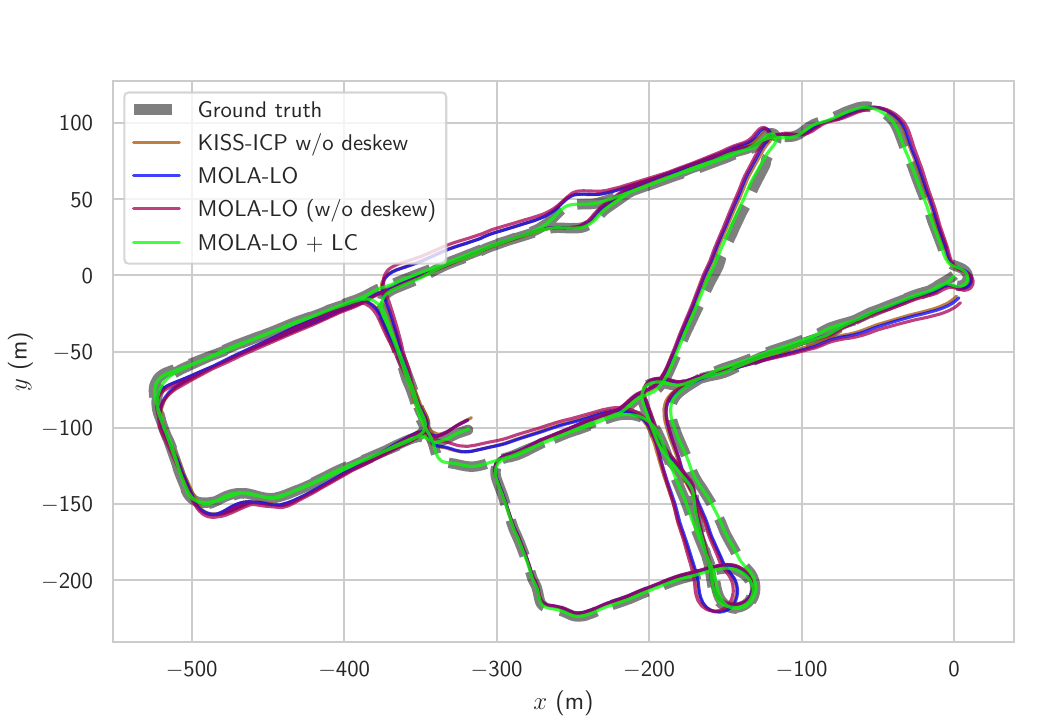}}
\\
\subfigure[Trajectory components]{\includegraphics[width=0.95\textwidth]{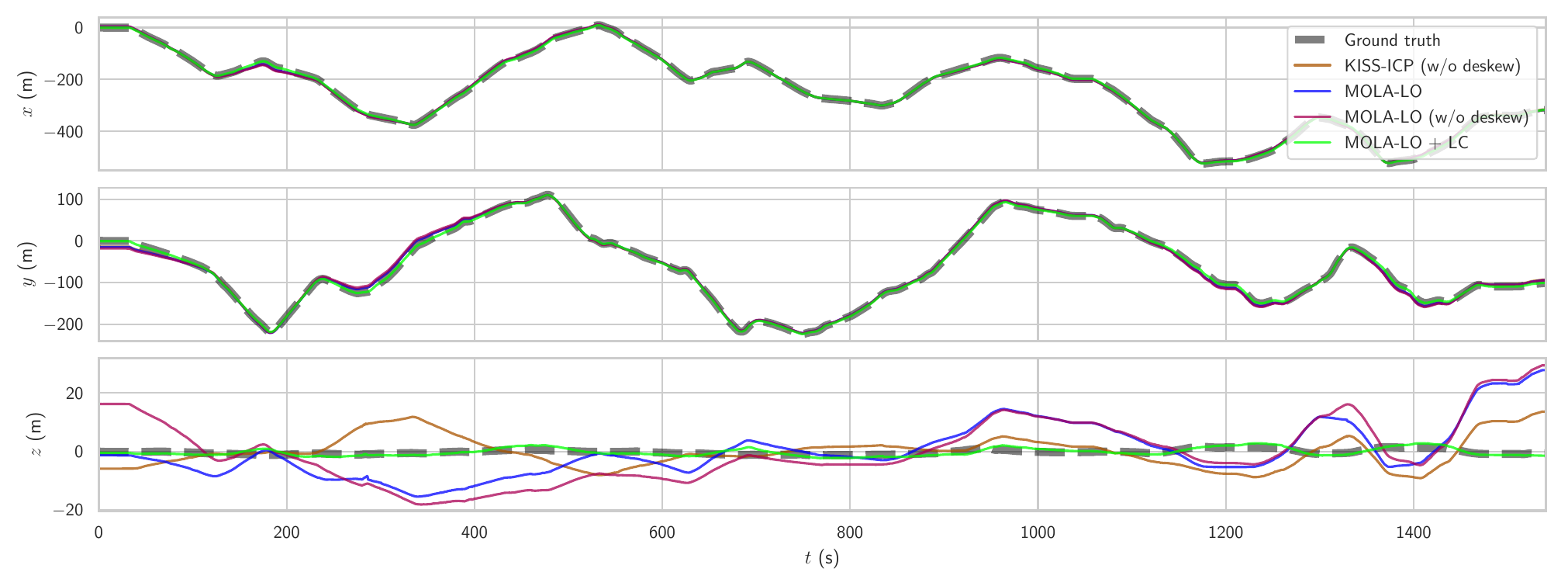}}
\end{tabular}
}
\caption{Estimated trajectories for the proposed LiDAR odometry system ("MOLA-LO") applied to the \textbf{UAL VLP-16 Campus} dataset, compared to ground truth. Trajectories denoted as ``MOLA-LO + LC'' include the post-processing loop-closure. See discussion in Section~\ref{sect:ual.campus}.
}
\label{fig:ual.vlp16-paths}
\end{figure*}

Finally, the proposed LO system, the full SLAM system including loop closure,
and the SOTA method KISS-ICP \citep{vizzo2023kiss} have been benchmarked
in the automotive dataset reported in \cite{blanco2019benchmarking}, 
where an electric vehicle is driven around the University of Almeria campus
while grabbing scans from a Velodyne VLP-16. Ground truth trajectory is available
from RTK GNSS 3D positioning. Estimated trajectories are shown in \Figure{fig:ual.vlp16-paths}, 
where it can be seen that, as already explained above, the largest part of the obtained ATE values
is caused by accumulation of errors in the vehicle altitude above the initial ground level.
Such error is drastically reduced by handling loop closures, as can be seen in \Figure{fig:ual.vlp16-paths}(b).
Regarding a quantitative analysis of errors, they are available in Table~\ref{tab:ual.vlp16.metrics},
with our system exhibiting $\sim 30\%$ less ATE than the baseline method (KISS-ICP), 
while still being able to run LO $\sim 6$ times faster than the sensor rate.

\begin{table}
\caption{Absolute translational error (ATE) (rmse values in meters as reported by ``\texttt{evo\_ape -a}'')
for the \textbf{UAL Campus} dataset, in which there is only one sequence.
Bold means best accuracy in each sequence and category.
}
\centering
{
\setlength\extrarowheight{1pt}
\begin{tabular}{|c||r|r|}
\hline
\textbf{Method}
& UAL Campus
& \TwoRows{Avr. time}{per frame}
\\
\hline
\TwoRows{KISS-ICP (w/o deskew)}{\cite{vizzo2023kiss}} & 
13.21~m & 12.0~ms
\\ %
\hline
\TwoRows{MOLA-LO (w/o deskew)}{(ours)} & 
\textbf{7.97}~m & 14.7~ms
\\ %
\hhline{|*{3}{=|}}
\TwoRows{KISS-ICP}{\cite{vizzo2023kiss}} & 
10.06~m & 11.2~ms
\\ %
\hline
\TwoRows{MOLA-LO}{(ours)} & 
\textbf{6.65}~m & 14.5~ms
\\ %
\hhline{|*{3}{=|}}
\TwoRows{MOLA-LO + LC}{(ours)}
 & 
\textbf{1.00}~m & 21.3~ms
\\
\hline
\end{tabular}
}
\label{tab:ual.vlp16.metrics}
\end{table}

\section{Qualitative demonstrations}
\label{sect:results.qualitative}

After experimentally measuring the accuracy of our proposal for several public 3D LiDAR 
datasets with known ground truth,
we now provide evidence of other qualitative features illustrating different claims of this work.
\REVIEW{Further claims will be also analyzed later on in a \emph{quantitative} way in Section~\ref{sect:ablation} with ablation studies.}

\subsection{2D LiDAR configuration}
\label{sect:2dlidar}

\begin{figure*}
\centering
\includegraphics[width=0.99\textwidth]{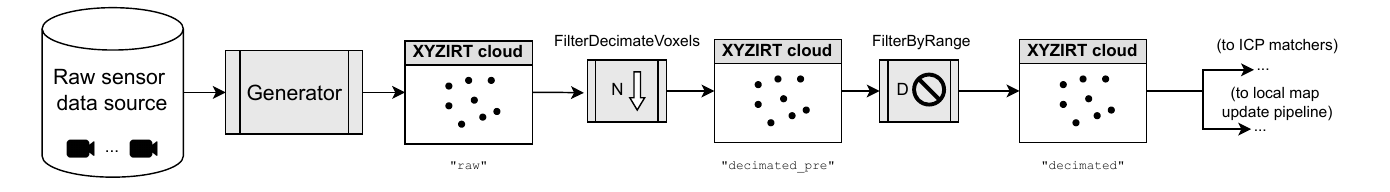}
\caption{Contents of the ``observation pipeline'' blocks in Figure~\ref{fig:lo-module} for the ``2D LiDAR'' LO system configuration.
Refer to discussion on pipelines in general in Section~\ref{sect:pipelines} and to Section~\ref{sect:2dlidar} for this particular pipeline.
}
\label{fig:lidar2d.obs.pipeline}
\end{figure*}

The flexibility of our framework to cope with different sensors is here illustrated
with an alternative configuration, dubbed \texttt{lidar-2d}, suitable
for mapping with 2D range finders. 
While all experiments shown so far comprise building a 3D point cloud local map
from 3D scans, just two configuration changes enable
the architecture in Figure~\ref{fig:lo-module} to build maps from 2D sensors.
First, the local map generator (block \#5 in Figure~\ref{fig:lo-module})
is set to create an occupancy voxel map.
A plain 2D grid map would also work but, as discussed in Section~\ref{sect:metric.maps},
the sparse data structure of voxel maps avoids the need for frequent memory reallocations
as the robot explores. 
Secondly, the observation pipeline (block \#3 in Figure~\ref{fig:lo-module})
is simplified to that shown in \Figure{fig:lidar2d.obs.pipeline}, that is,
there is no need to down-sample the raw sensor points twice (for registration and for map update)
due to the reduced size of 2D scans.

\begin{figure*}
\centering
\subfigure[\texttt{fr079} final map]{\includegraphics[width=0.85\textwidth]{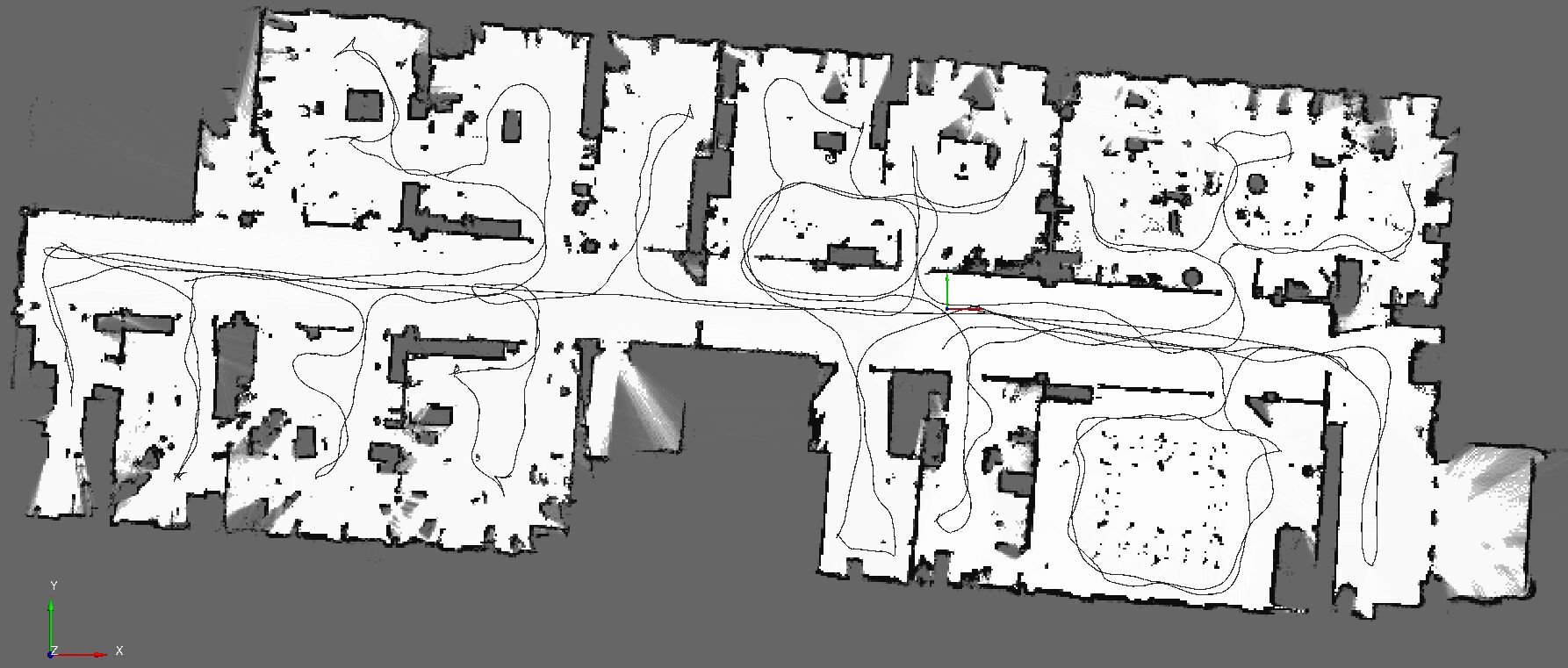}}
\subfigure[Málaga CS faculty final map]{\includegraphics[width=0.95\textwidth]{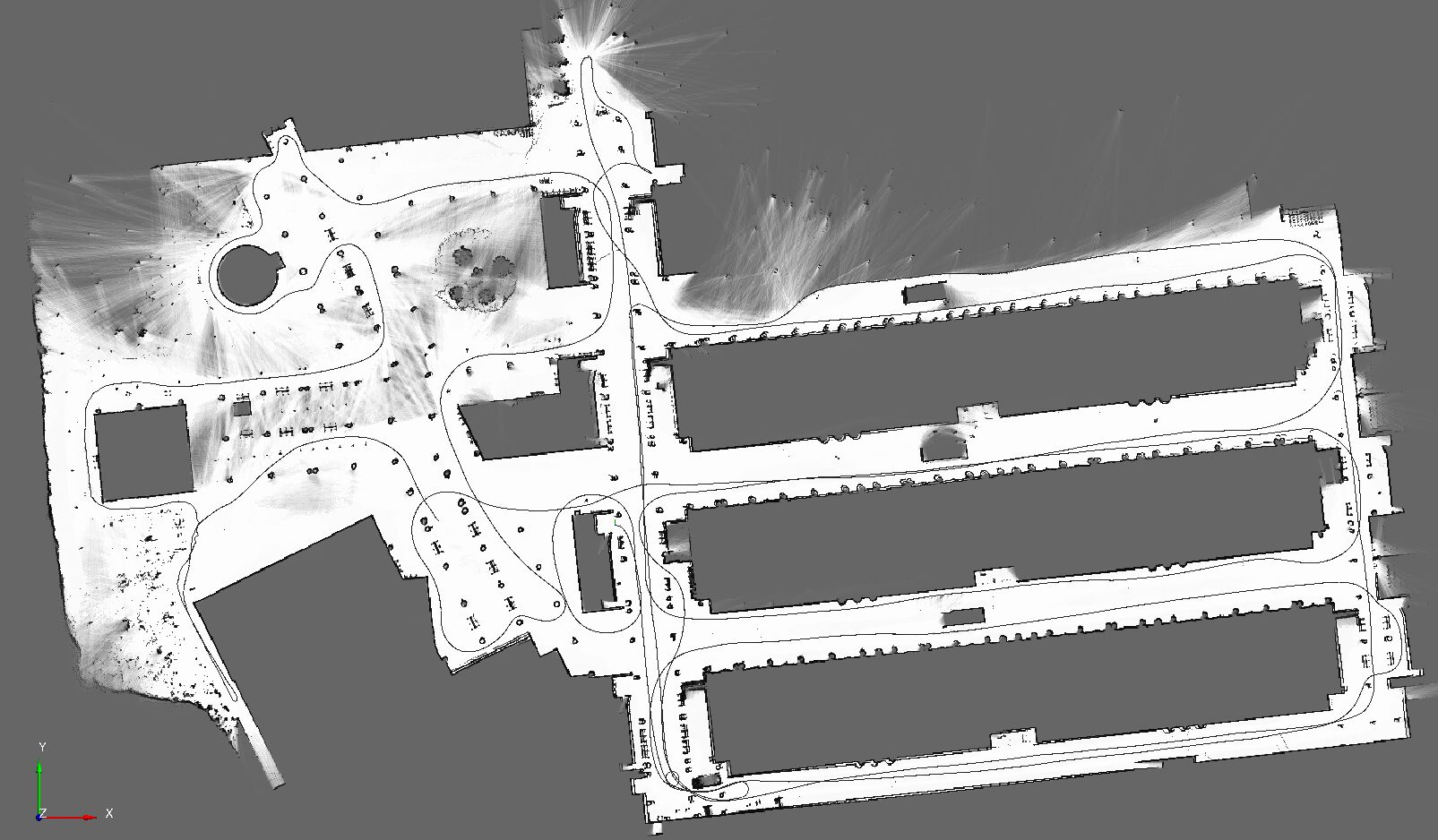}}
\caption{Results for the 2D LiDAR configuration. Refer to discussion in Section~\ref{sect:2dlidar}.
}
\label{fig:lidar2d.results}
\end{figure*}

This alternative configuration has been validated with the datasets exposed below, 
all of them from wheeled robots equipped with encoders from which incremental odometry is taken
as an input for the kinematic state prediction module (Section~\ref{sect:navstate.fuse}).
First, we applied the LO system to the Freiburg building 079 (\texttt{fr079}) dataset \citep{Radish}, 
recorded by Cyrill Stachniss in 2010 with a Pioneer2 mobile robot equipped with a SICK LMS range finder.
The resulting voxel map (which is only populated at one fixed height) is illustrated in \Figure{fig:lidar2d.results}(a).
Our method took an average of 10.2~ms to process each scan.
Note that loop closures were not required here.

Secondly, we have applied the SLAM solution (LO plus loop-closure detection)
to the larger Málaga CS faculty dataset, comprising a 1.9~km trajectory of a robotic wheelchair 
equipped with encoders and a 2D SICK LMS range finder. It was recorded in 2006 and 
is available for download in \cite{blanco2024uma2d}. 
The resulting consistent global map is shown in \Figure{fig:lidar2d.results}(b).
As with the former dataset, there is no ground truth for quantitative quality assessment.
Our system takes an average of 68.8~ms per scan, including loop closure detection and global optimization.

\begin{figure*}
\centering
\subfigure[DCC01]{\includegraphics[width=0.5105\textwidth]{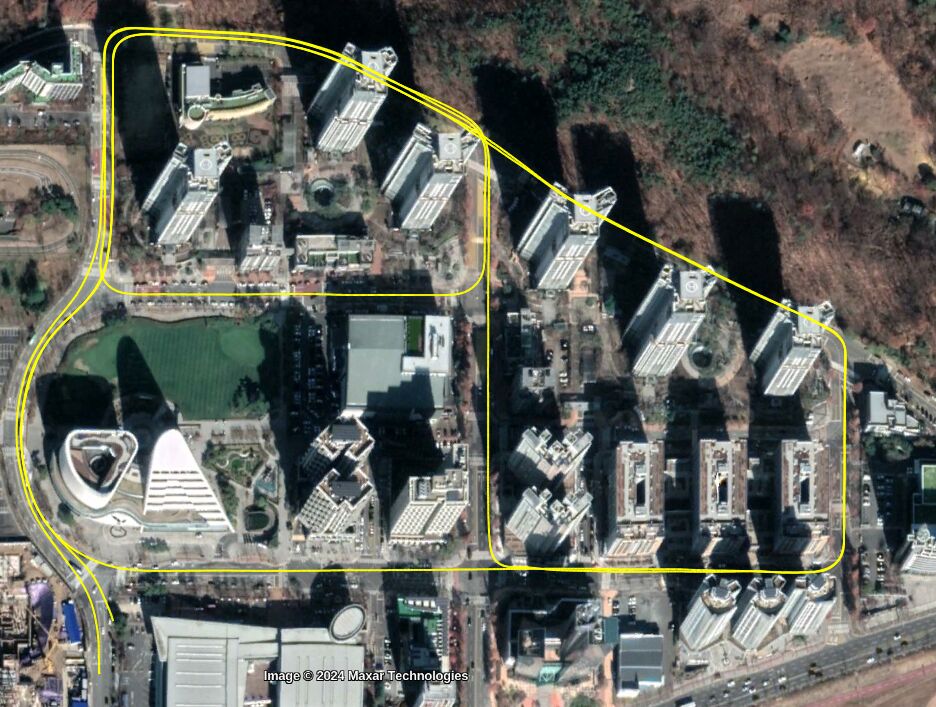}}
\subfigure[Riverside01]{\includegraphics[width=0.4695\textwidth]{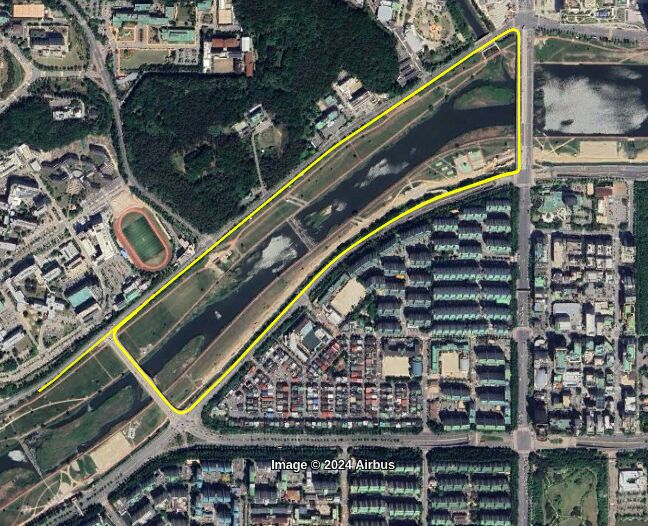}}
\subfigure[KAIST03]{\includegraphics[width=0.58\textwidth]{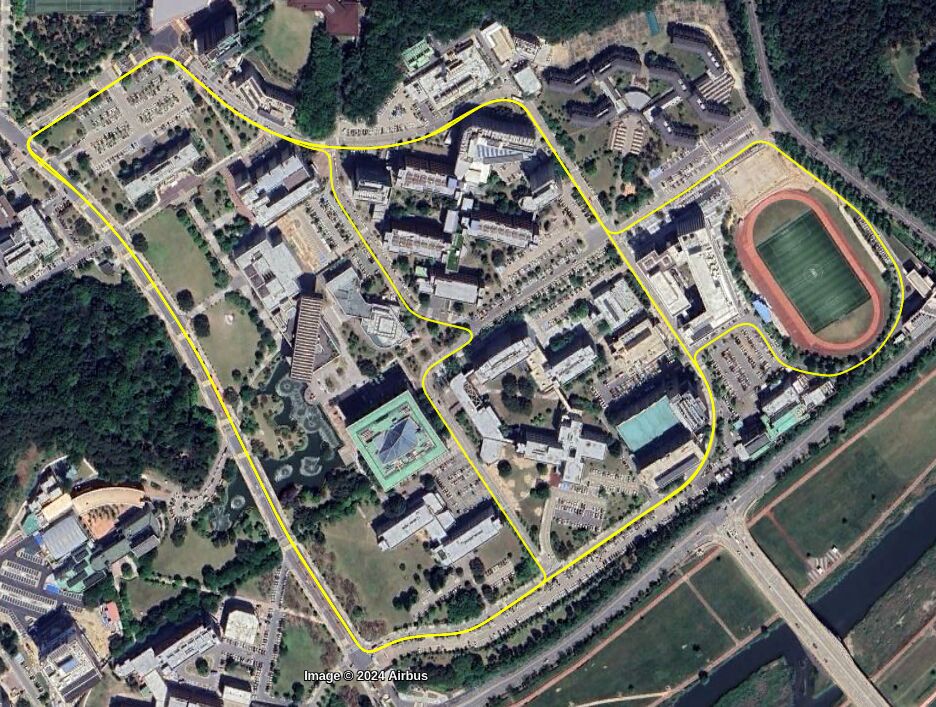}}
\subfigure[Sejong02]{\includegraphics[width=0.41\textwidth]{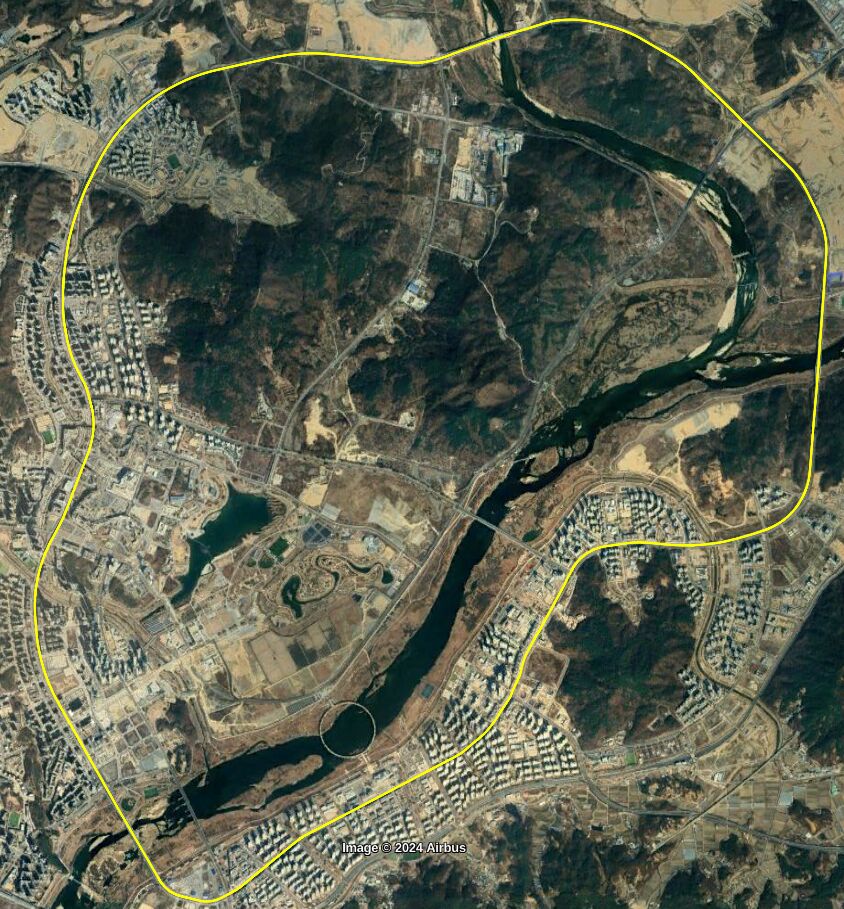}}
\caption{Illustrative results for metric map georeferenciation for sequences estimated by the SLAM solution
and automatically georeferenced as described in Section~\ref{sect:results.geo} from consumer-grade GNSS data
(Screenshot from Google Earth v7.3.6.9326, images by: Maxar Technologies, 2024; Airbus, 2024).)
}
\label{fig:mulran.georef}
\end{figure*}

\begin{figure*}
\centering
\subfigure[Forest parcel \#2]{\includegraphics[width=0.48\textwidth]{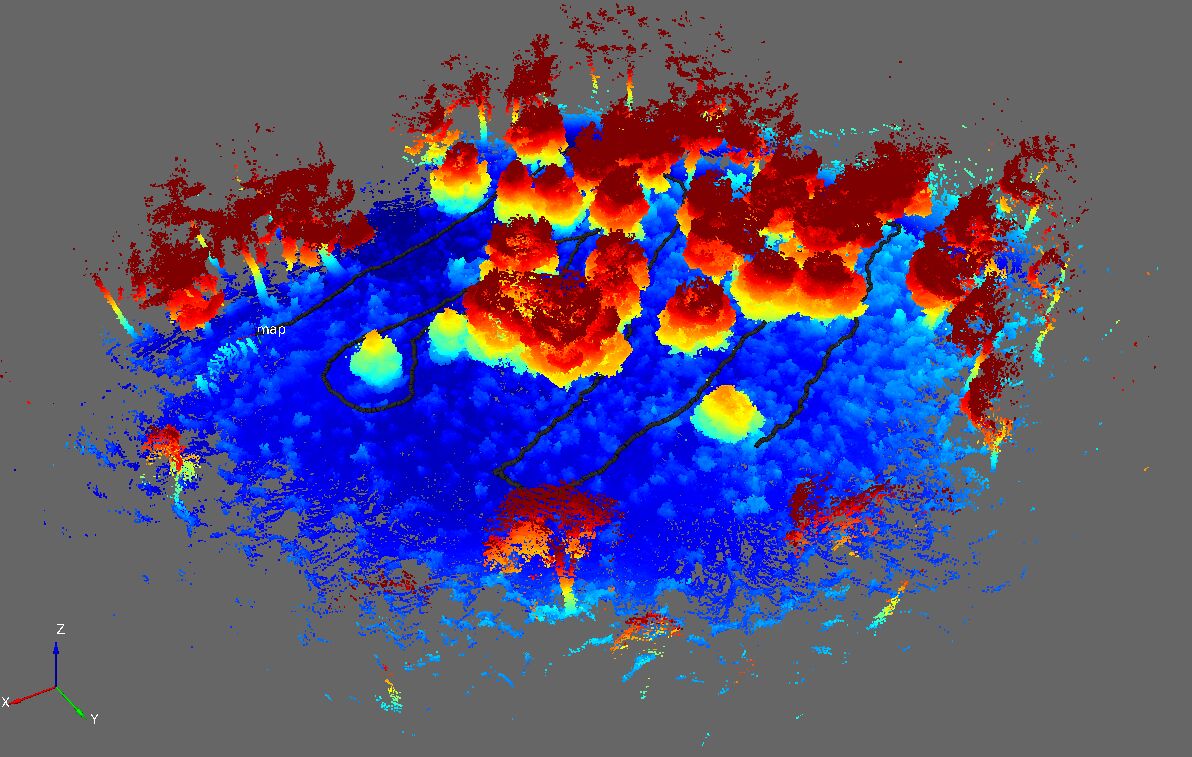}}
\subfigure[Forest parcel \#3]{\includegraphics[width=0.43\textwidth]{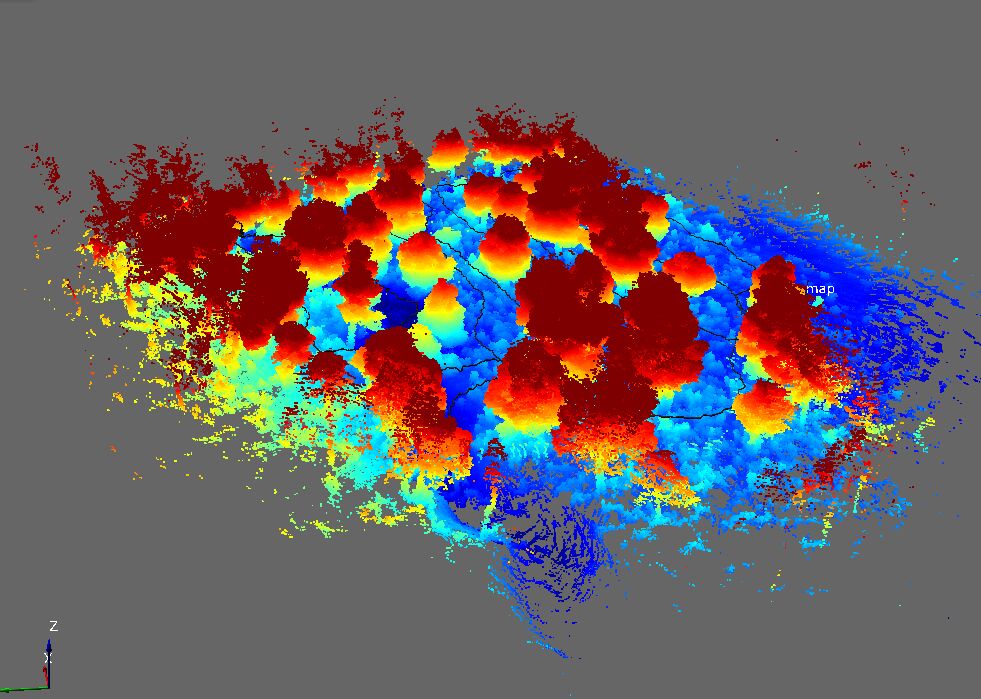}}
\caption{Final 3D maps from datasets collected in Almería forests with a LiDAR backpack. See discussion in Section~\ref{sect:backpack}.}
\label{fig:backpack.bosques}
\end{figure*}

\subsection{Georeferencing metric maps}
\label{sect:results.geo}

Finding the geodetic coordinates of metric maps is possible from a low-cost GNSS receiver
given observations enough, following the method described in Section~\ref{sect:sm.georef}.
In this section we show experimental results in georeferencing maps, 
based on the MulRan dataset (see Section~\ref{sect:mulran}) since it includes both, accurate ground truth,
and readings from a low-cost GNSS device.

To illustrate georeferenciation, we took estimated trajectories in the local map frame, 
convert them to the ENU frame (refer to Figure~\ref{fig:georef}(a)), then to Earth geocentric coordinates, 
and finally to geodetic datums for the WGS84 standard geoid.
The estimated paths for a subset of four MulRan dataset sequences were exported
to KML (Keyhole Markup Language) and displayed on Google Earth for visual inspection. 
Figure~\ref{fig:mulran.georef} shows the final results, automatically generated by tools in the presented
framework.

\begin{figure*}
\centering
\includegraphics[width=1\textwidth]{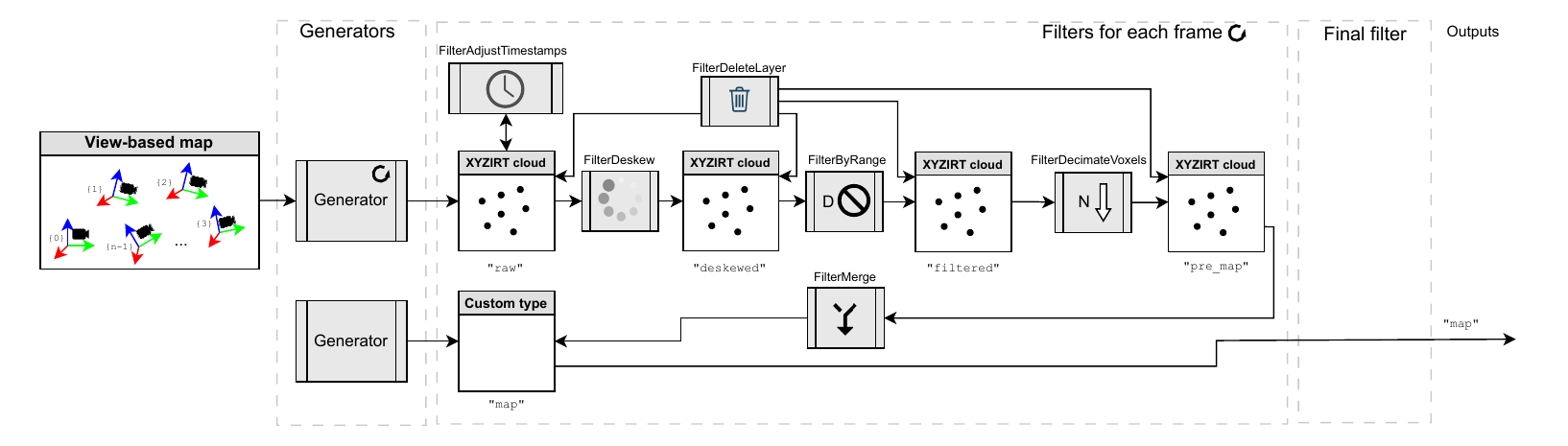}
\caption{Configuration for building metric maps of arbitrary types from view-based maps using the application \texttt{sm2mm}.
The type of the final map is solely determined by the ``Generator'' block on the bottom, without any other required change in the rest.
Refer to discussion in Section~\ref{sect:build.diff.maps}.
}
\label{fig:sm2mm.pipeline.dem}
\end{figure*}

\begin{figure*}
\centering
\subfigure[Occupancy voxel map]{\includegraphics[width=0.44\textwidth]{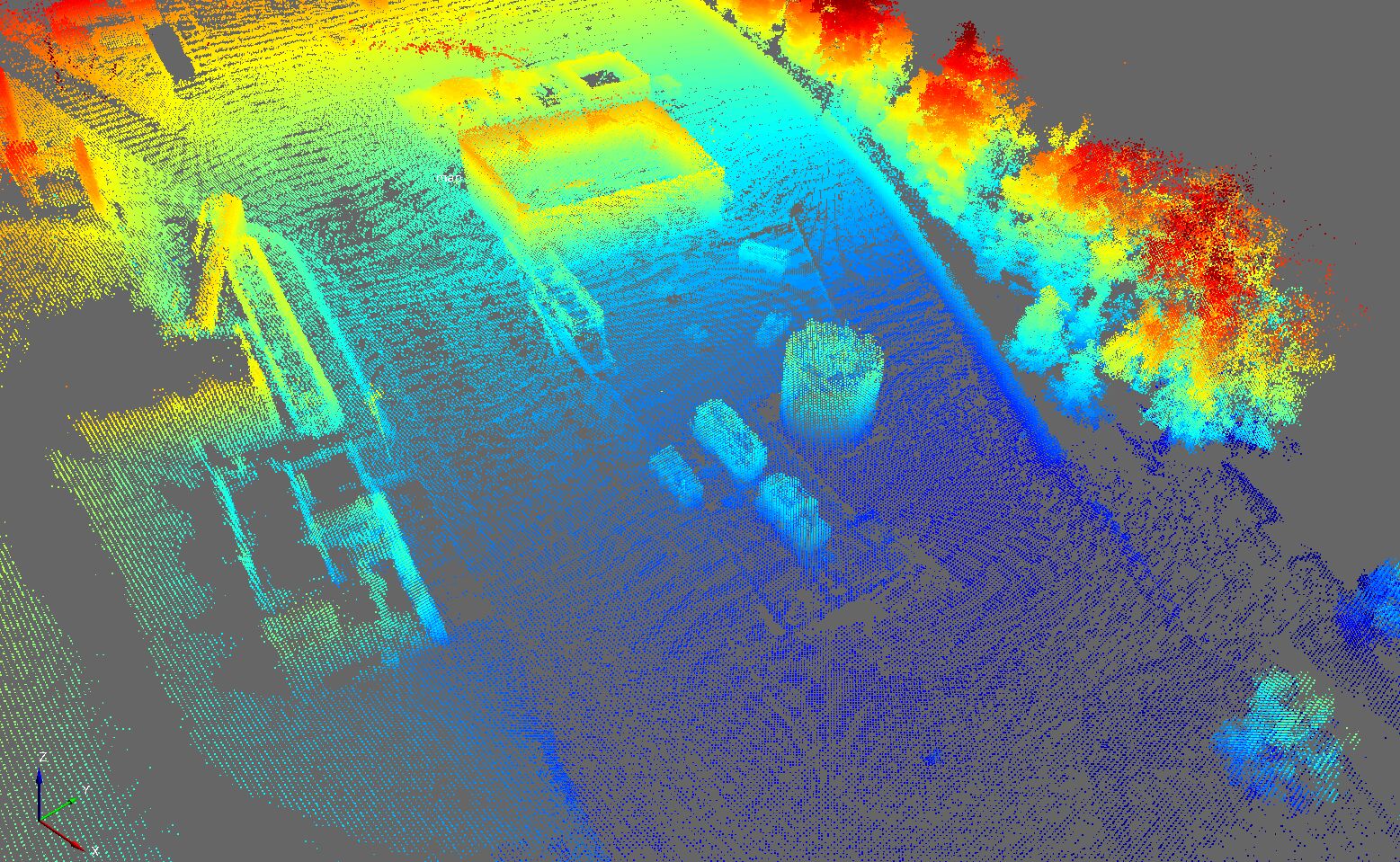} }
\subfigure[Digital terrain elevation model]{\includegraphics[width=0.44\textwidth]{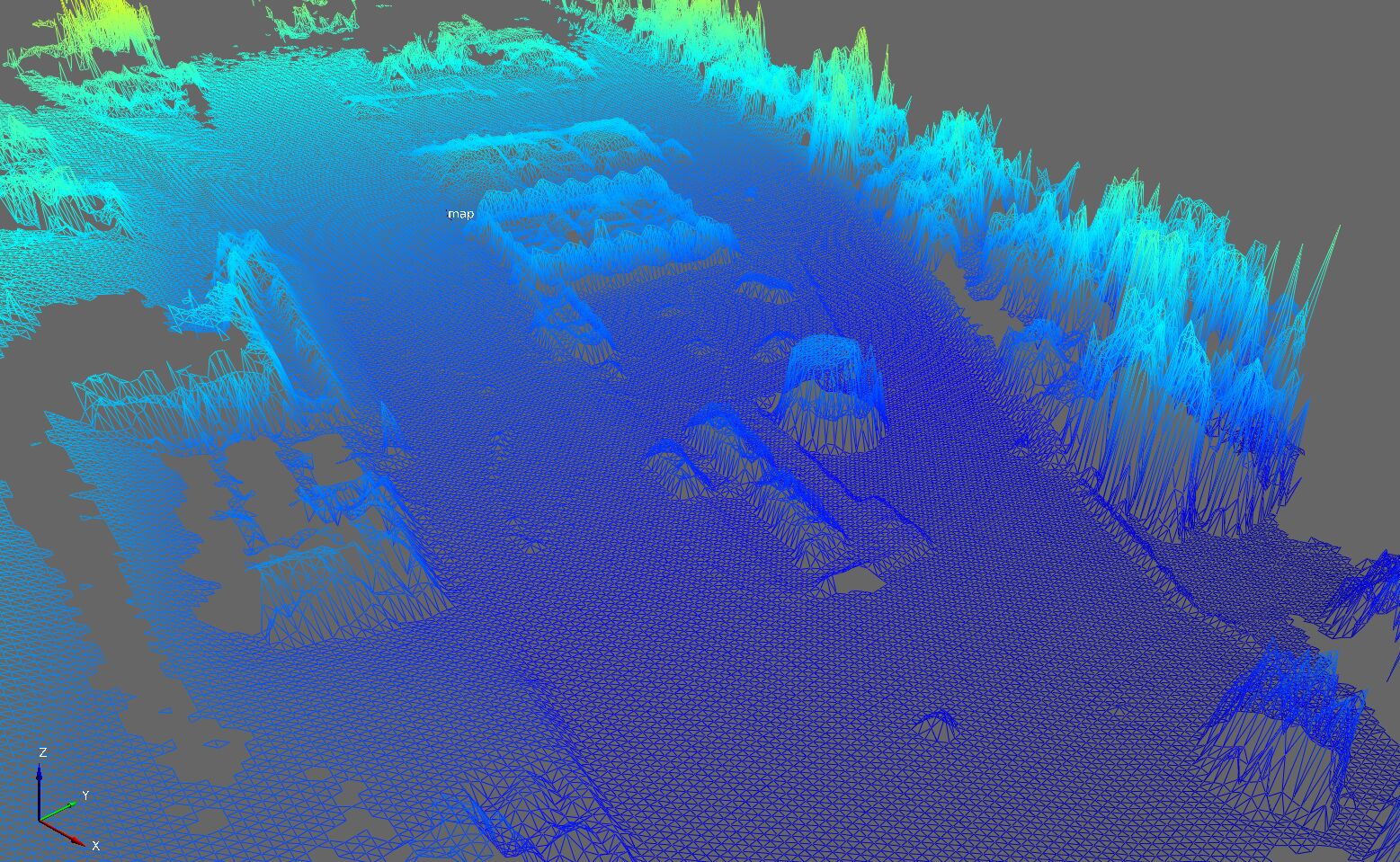} }
\caption{Example final maps built from the Voxgraph dataset. The voxel map in (a) has been visualized 
as one point per voxel to favor a clearer visualization.
Refer to discussion in Section~\ref{sect:build.diff.maps}.
}
\label{fig:sm2mm.examples.dem}
\end{figure*}

\subsection{Use as a backpack mapping system}
\label{sect:backpack}
As detailed in \cite{aguilar2024lidar}, the proposed LO system has been used
for mapping forest areas from datasets
grabbed with a backpack kit comprising an Ouster OS0-32 as the unique sensor.
Note that the system configuration was not modified with respect to all tested
datasets in Section~\ref{sect:results.quant} and, although these experiments
did not have ground truth for quantitative assessment, the final metric maps
are consistent, as illustrated in \Figure{fig:backpack.bosques} with views of 2 out of the 6 tested sequences.
This experiment shows that the LO system is able to cope with walking patterns in uneven terrains while
mapping unstructured, natural environments.

\subsection{Building different metric maps}
\label{sect:build.diff.maps}

One of the claims of this work is that different robotic tasks are better performed using different map representations, 
thus the mapping framework should be able to provide such flexibility.
To illustrate this feature, we show next how easy is to create metric maps of different types,
apart of the commonly-used point clouds.
Focusing on the post-processing stage, once mapping is done and there is a view-based map from 
the LO or SLAM system, it was explained in Section~\ref{sect:sm2mm} how such map 
can be converted into metric maps. In particular, \Figure{fig:metric.map.pipeline.example.b}
illustrated how to build several map layers by accumulating 3D LiDAR scans into a global map.

Here, we use a modification of such pipeline, shown in \Figure{fig:sm2mm.pipeline.dem}, where only one final map layer exists
of a custom type which will be changed between experiments.
The final map obtained from the Voxgraph dataset (Section~\ref{sect:voxgraph}) was fed into that pipeline
with two map types: an occupancy voxel map, and a 2D digital elevation model (DEM).
The resulting maps are illustrated in \Figure{fig:sm2mm.examples.dem}(a)--(b).
Changing a few lines in a configuration YAML file is all that it takes to build one map type or the other.

\begin{figure*}
\centering
\subfigure[KAIST01 - Localization error map]{\includegraphics[width=0.45\textwidth]{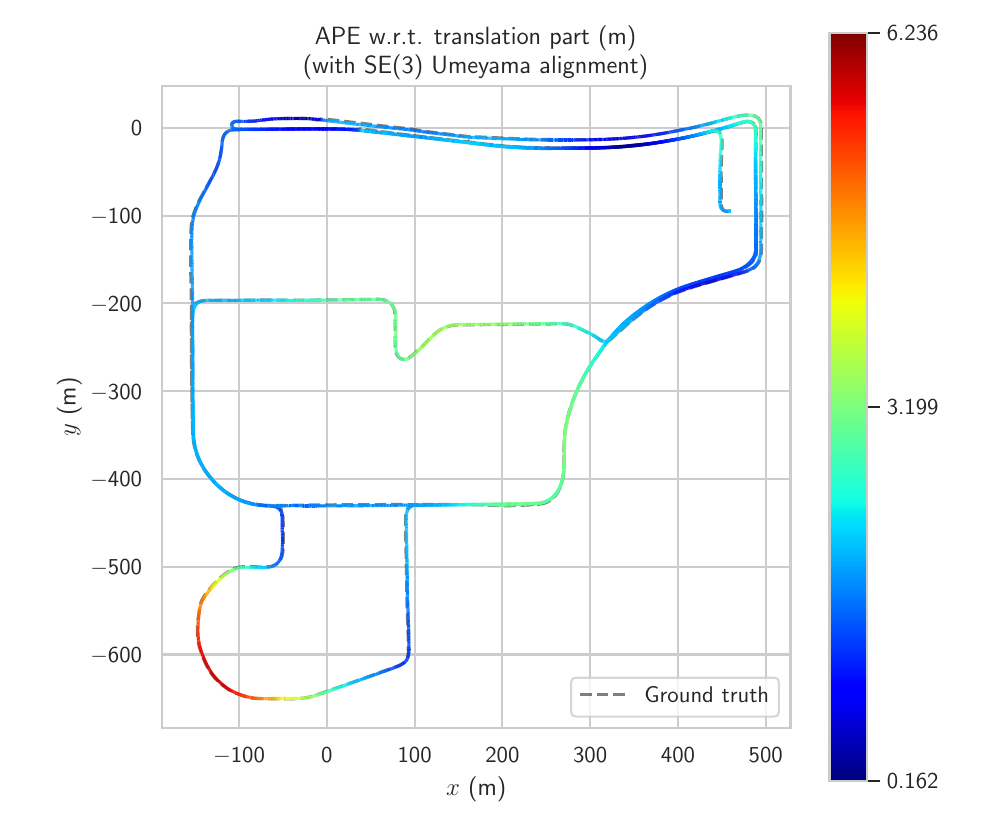} }
\subfigure[KAIST01 - Localization error plots]{\includegraphics[width=0.45\textwidth]{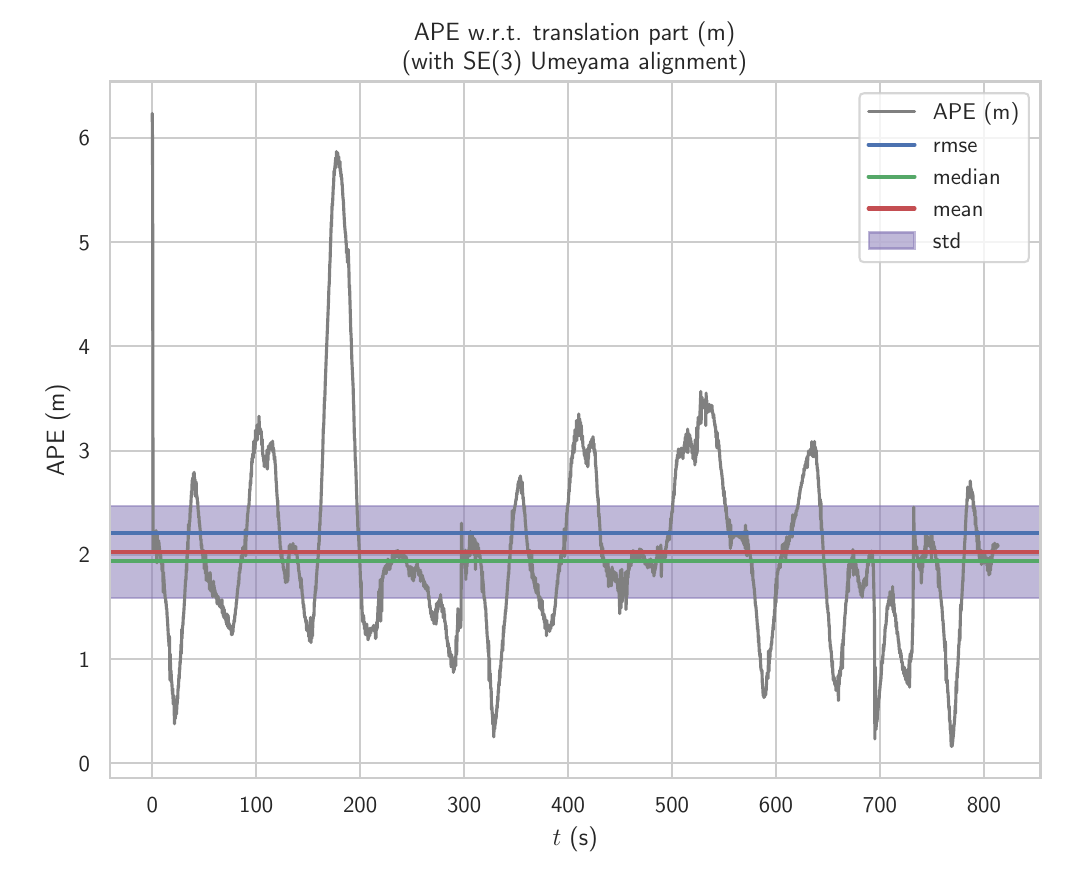} }
\\
\subfigure[KAIST03 - Localization error map]{\includegraphics[width=0.45\textwidth]{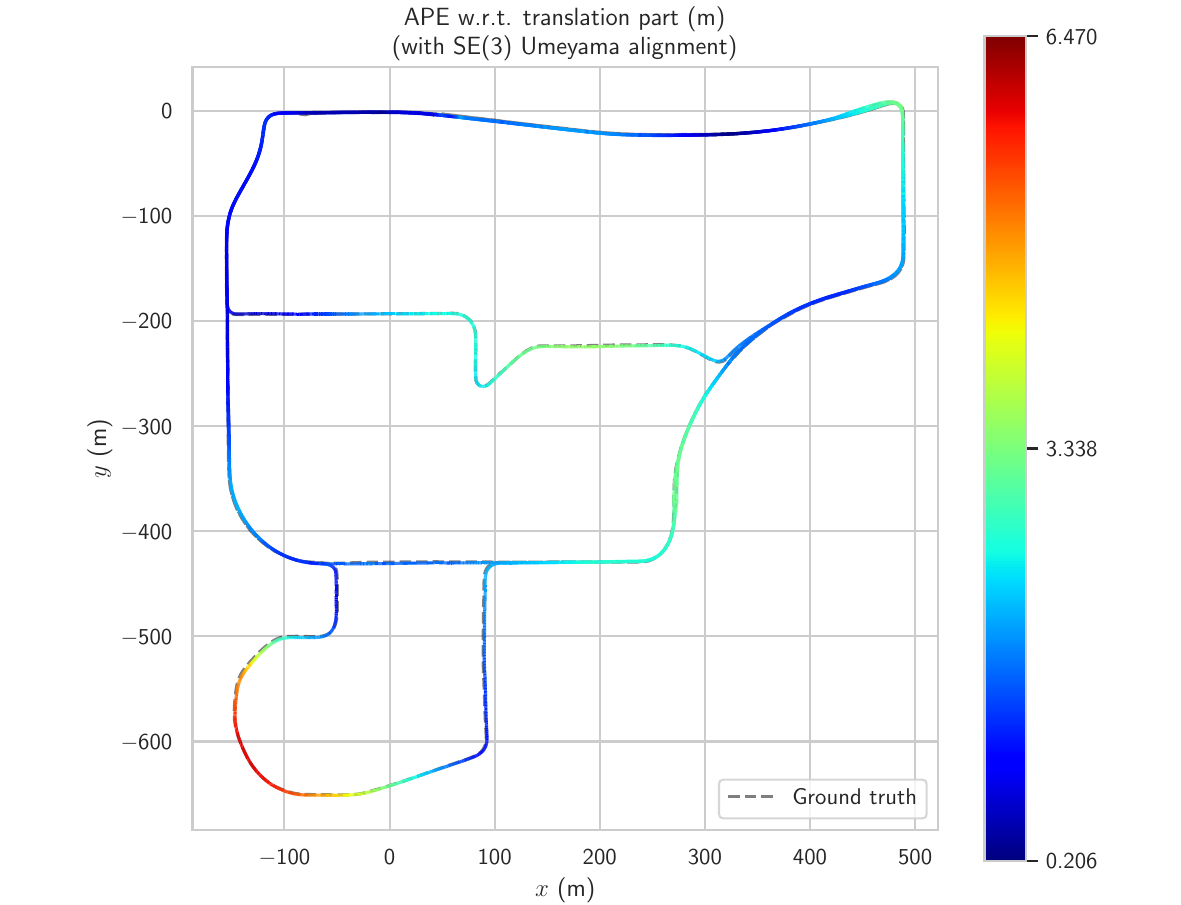} }
\subfigure[KAIST03 - Localization error plots]{\includegraphics[width=0.45\textwidth]{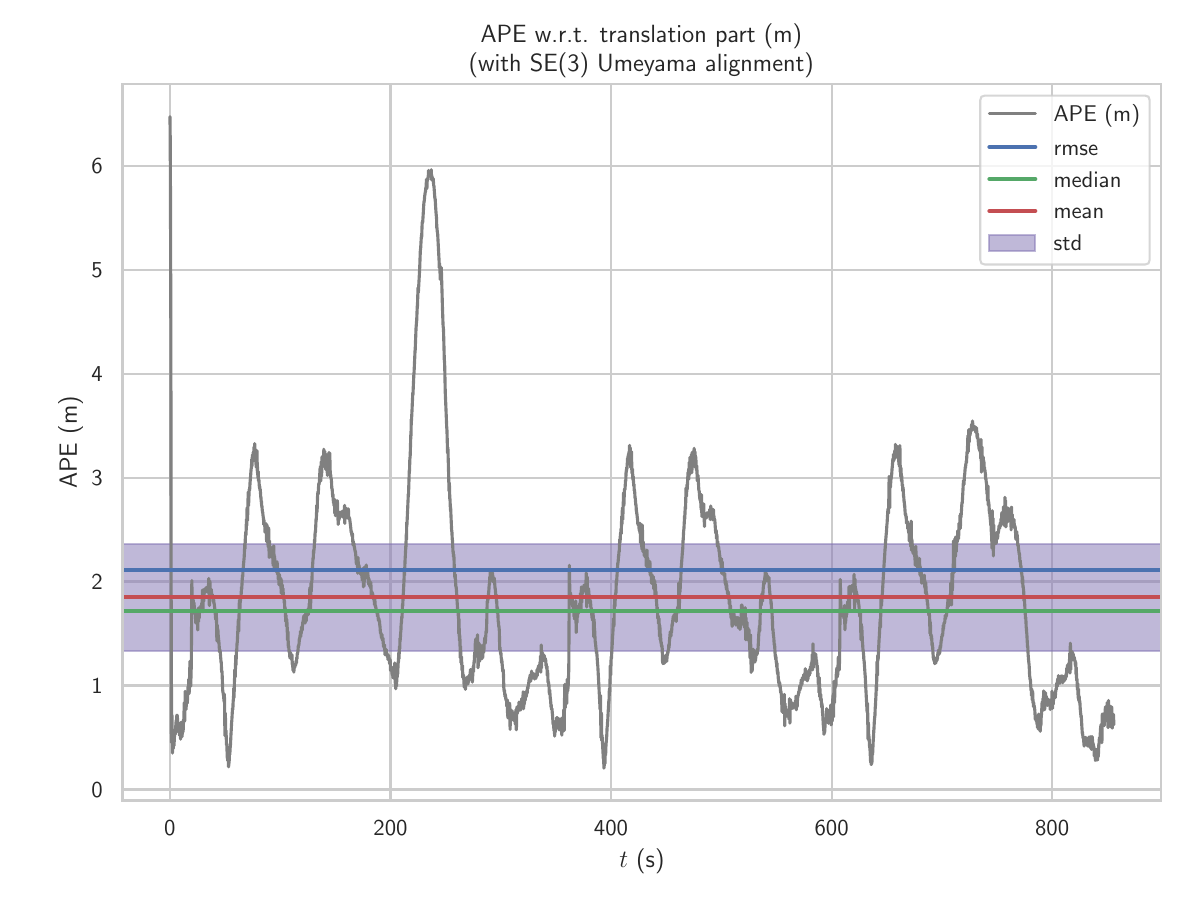} }
\caption{Results of the \textbf{localization} experiments performed with \textbf{MulRan dataset}
sequences KAIST01 and KAIST03 by using as reference the map built from KAIST02.
Refer to discussion in Section~\ref{sect:results.localiz}.
}
\label{fig:results.localiz.kaist}
\end{figure*}

\section{Localization experiments}
\label{sect:results.localiz}

Localization without mapping can be performed in the present framework in two ways:
(i) with particle filtering (as in \cite{blanco2019benchmarking}) using the \texttt{mrpt\_pf\_localization} software package, 
or (ii) directly with the LO pipeline introduced in Section~\ref{sect:lidar.odometry}, by disabling map updates. 
In both cases, the reference map can be given from converting a view-based map from a mapping session into 
a metric map, as described in Section~\ref{sect:sm2mm}.
Robots moving slowly and equipped with wheels encoders may be good fits for the particle filtering approach. For faster vehicles and drones,
the LO pipeline is probably a better option due to the more accurate initial predictions for each time step given by the kinematic state prediction module 
(Section~\ref{sect:lidar.odometry}).

In this section, we provide experimental results for localization using the MulRan dataset, which perfectly
fits this task since three driving sequences exist for each location. Therefore, a metric map from the 
SLAM output for one such sequence (\texttt{KAIST02}) is built and used as reference map for localizing 
the other two sequences of the same location (\texttt{KAIST01} and \texttt{KAIST03}). 
The reference global map was built with the metric map building pipeline
in Figure~\ref{fig:metric.map.pipeline.example.a}, using a point cloud as global map. 
In particular, from all point cloud maps described in Section~\ref{sect:metric.maps}, 
we selected the one with hashed voxels as underlying data structure, 
with a 2~m resolution and a maximum of 20 points per voxel.
Localization errors have been estimated using \texttt{evo\_ape} \citep{grupp2017evo} between the estimated trajectories
and ground truth, as illustrated in \Figure{fig:results.localiz.kaist}. 
In this case, the initial approximated pose was given manually for each sequence, with an error of several meters, hence the error peak at the beginning of each sequence.
Automatically re-localizing from GNSS readings
is possible as a more practical alternative, 
but at the time of writing, this feature is only implemented in the particle filter-based solution.
The localization RMSE with respect to ground truth in these two
experiments were 2.21~m and 2.11~m, respectively. 
Note, however, that virtually all this error
is attributable to accumulated residual drift in the map used as reference, 
since 2.13~m is precisely the RMSE ATE of the full SLAM solution for sequence \texttt{KAIST02}
(refer to row ``MOLA-LO+LC'' in Table~\ref{tab:mulran.metrics}).
Basically, localization accuracy is limited by how accurate the reference map is.
Note that the tested sequences include dynamic objects (e.g. vehicle overtaking), starts and stops due to traffic, etc.

\section{\REVIEW{Ablation studies}}
\label{sect:ablation}

\begin{figure*}
\centering
\subfigure[Mulran (automotive)]{\includegraphics[width=0.46\textwidth]{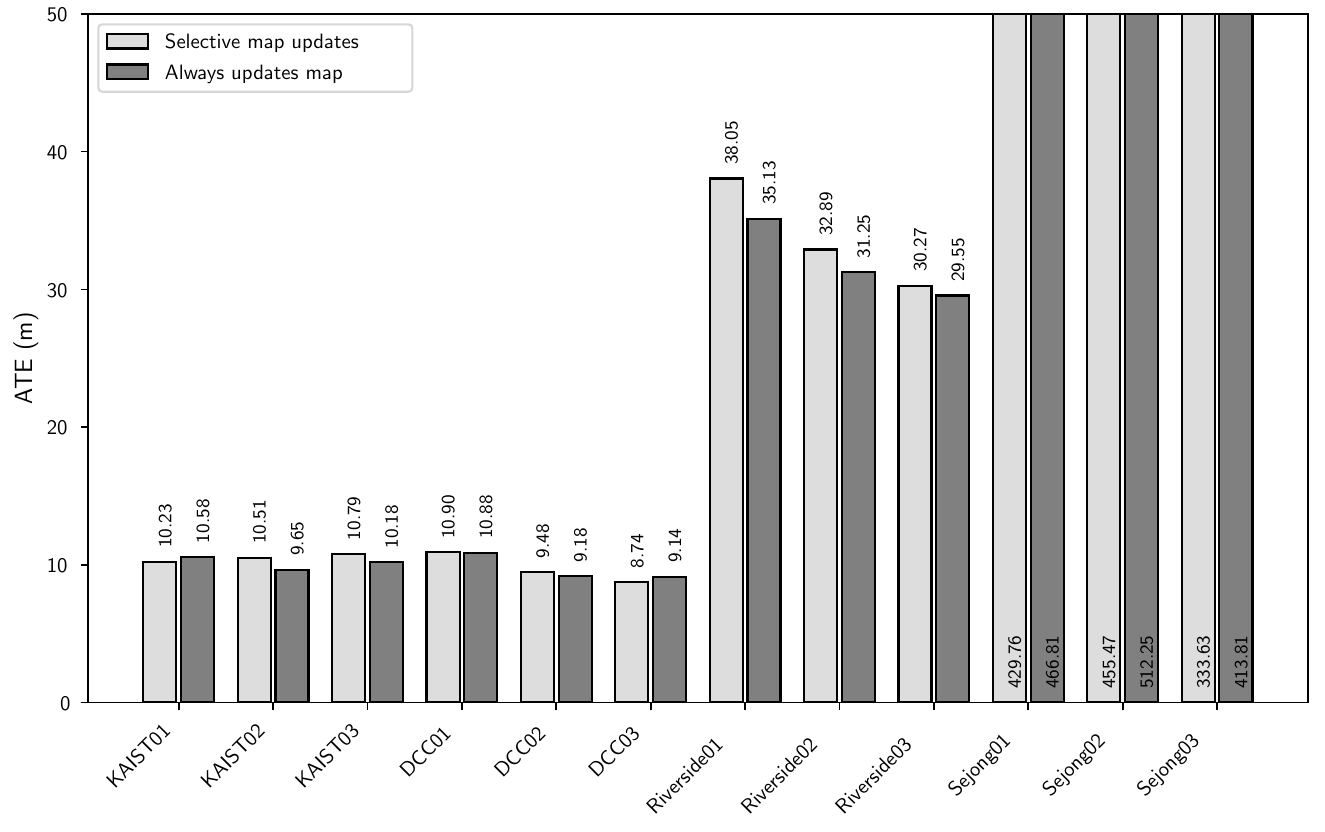} }
\subfigure[Newer College Dataset (handheld)]{\includegraphics[width=0.52\textwidth]{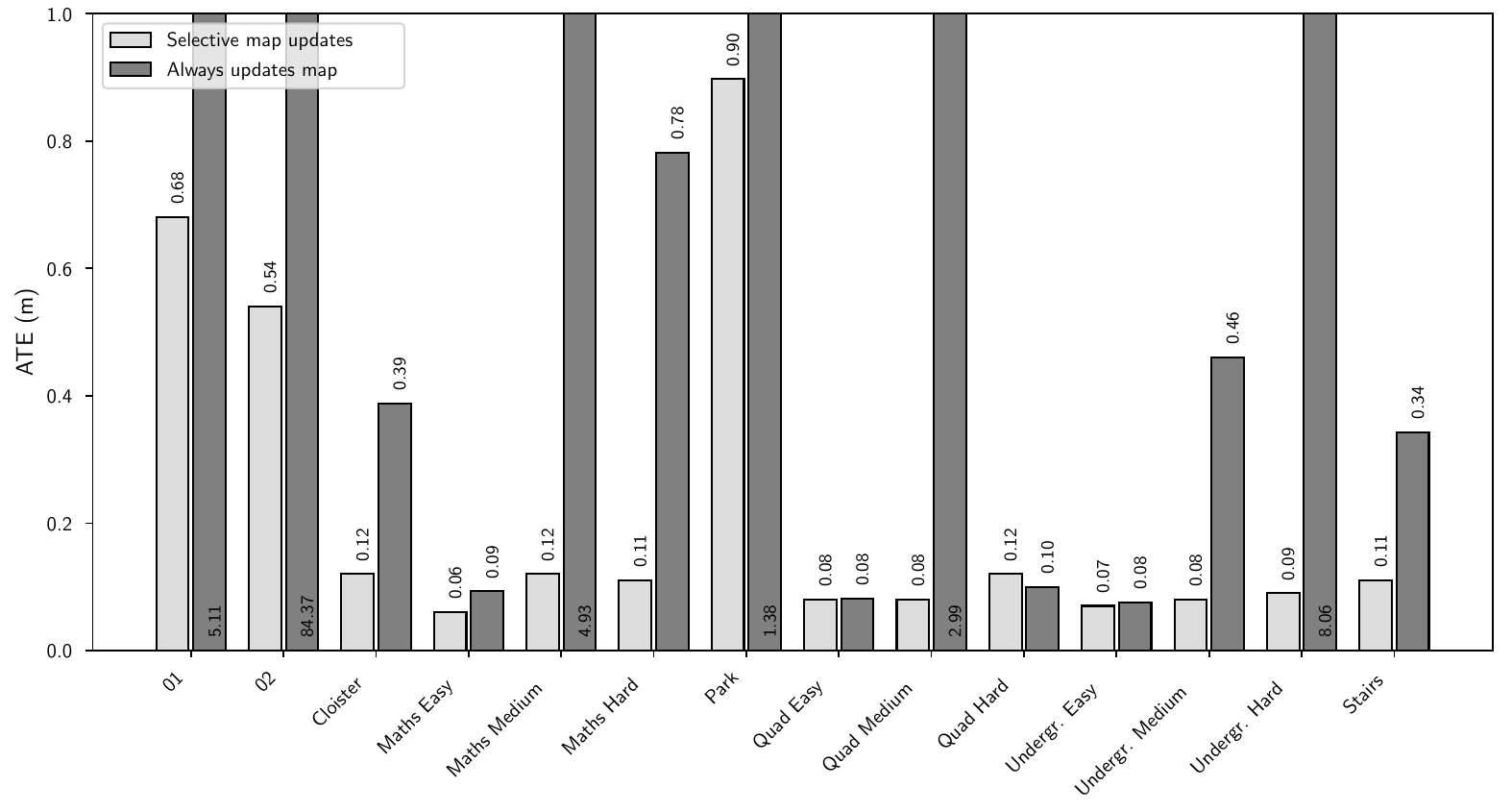} }
\caption{\REVIEW{Results of the ablation experiments to evaluate the proposed selective map update feature, showing how it becomes relevant for datasets with irregular motion profiles (e.g. handheld sensors).
Refer to discussion in Section~\ref{sect:ablation.localmap.updates}.}
}
\label{fig:ablation.map.updates}
\end{figure*}

\subsection{\REVIEW{Scan deskewing}}

\REVIEW{First, we want to
validate that scan deskewing by means of
trajectory interpolation on SE(3) per Eq.~(\ref{eq:deskew})
is good enough and effectively leads to more accurate results.
As a benchmark, we tested our LiDAR odometry system (without loop closure)
on the UAL campus dataset in \cite{blanco2019benchmarking}
with and without deskewing the input scans, obtaining
the results in Table~\ref{tab:ual.vlp16.metrics}.
The obtained ATE values with respect to ground truth 
with and without deskewing are 6.65~m and 7.97~m, respectively.
Therefore, undistorting the scans reduced the ATE in a 16.6\%,
despite the fact that driving speed was relatively slow in this
dataset, an average of $\sim 3~m/s$ (10.8~km/h or 6.7~mph).
For comparison, enabling deskewing in KISS-ICP also reduced
the ATE from 13.21~m to 10.06~m, a 23.8\% improvement.
Therefore, we can conclude that deskewing is absolutely
a must for accurate localization, even at reduced speeds.
}

\subsection{\REVIEW{Local map updates}}
\label{sect:ablation.localmap.updates}

\REVIEW{
As mentioned in Section~\ref{sect:lidar.odometry}, our system differs from others
in that we do not always update the local map for every single time step.
The reason behind this decision is that small localization errors and imperfect scan de-skewing,
specially during abrupt angular accelerations
would lead to map insertion of misplaced points which would then serve as (incorrect)
matches for successive scans, typically leading to unbounded error growth in localization.
We postulate that such misplaced points are more likely in high-dynamics datasets than in those with smoother trajectories; e.g. handheld vs. automotive.
This effect is expected when input scans are subsampled before registration against a local metric maps that do not fuse information from different scans, such as the point clouds used in our default system configuration and in many recent LiDAR odometry approaches.
Probabilistic map representations or the 3D-NDT map described in Section~\ref{sect:metric.maps} are expected to be more stable against frequent updates.}

\REVIEW{To support our hypothesis, 
we performed the following ablation experiment: we compared the trajectory error
with and without the proposed selective map update in an automotive (Mulran)
and a handheld (Newer College) dataset.
The quantitative results were already reported in 
Table~\ref{tab:mulran.metrics} and Table~\ref{tab:ncd.metrics}, respectively, 
but for the sake of a clearer visual comparison we have represented them in 
Figure~\ref{fig:ablation.map.updates}.
As can be observed, updating the map for every single time step is not problematic
for the automotive dataset, with trajectory errors slightly better in some sequences
and worse on others, but not leading to drastic differences. 
In turn, not using our proposed selective map update strategy for the Newer College dataset
leads to a dramatic increase of (at least) $50\%$ for the ATE in 11 out of 14 sequences, 
hence supporting the introduction of our selective update strategy for robustness
against high-dynamic motion profiles.
}


\begin{figure}
\centering
\includegraphics[width=1.0\columnwidth]{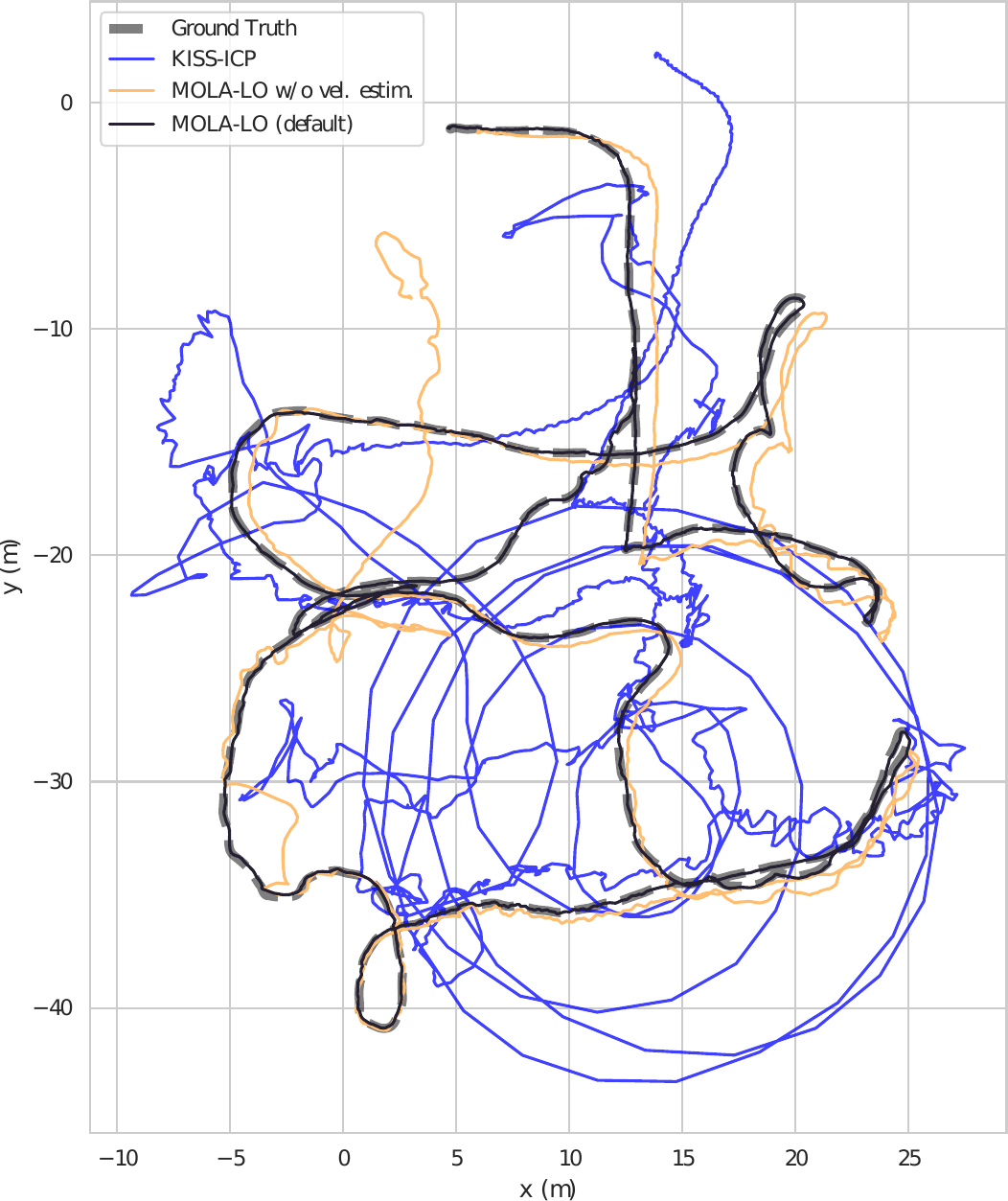}
\caption{\REVIEW{Evaluation of the impact of the tightly-coupled estimation of velocity vectors as part of the ablation study in Section~\ref{sect:ablation.twist} for the ``Underground hard'' sequence of the Newer College Dataset. The picture shows a bird-eye view of the estimated trajectories.
In the legend, ``MOLA-LO'' refers to the default system as introduced in the manuscript, while ``MOLA-LO without velocity estimation'' is a version where tightly-coupled estimation of this vector is disabled.}
}
\label{fig:ablation.twist}
\end{figure}

\subsection{\REVIEW{Tightly-coupled estimation of vehicle velocities}}
\label{sect:ablation.twist}

\REVIEW{Next, we want to evaluate the impact of the tightly-coupled estimation of 
linear and angular velocities within the optimization loop (refer to Section~\ref{sect:icp.local.map.update}).
As baseline, we use the proposal of keeping the velocity vectors as estimated 
from the last time step, as done in the state-of-the-art method KISS-ICP \cite{vizzo2023kiss}.
Comparison will be done using two Newer College Dataset sequences with high dynamics: ``Maths hard'' (MH) and ``Underground hard'' (UH). 
The trajectory ATE for MH increases from 0.11~m using the default system configuration to 0.15~m if the feature at test is disabled.
A more drastic increase of the error occurs for the UH sequence,
where an ATE of 0.086~m becomes 2.79~m. 
By visualizing the estimated paths in detail, shown in 
Figure~\ref{fig:ablation.twist}, 
it can be seen that the error comes from two sources: divergence
of the estimation at multiple points, and jerky trajectories
for the rest.
These experiments demonstrate that this feature alone has a great
potential to increase the stability of LiDAR-only odometry methods.
}

\subsection{\REVIEW{Horn's method}}
\label{sect:ablation.horn}

\REVIEW{As mentioned in Section \ref{sect:icp.solvers},
the closed-form Horn's solution to pointwise pairings
would seem to be an ideal alternative to iterative methods such as Gauss-Newton due to its simplicity and efficiency. 
However, as we demonstrate in this section, there are two problems
that render iterative non-linear methods as better practical solutions.
We have compared the accuracy of our default LiDAR odometry system
with another version where the Gauss-Newton optimizer is replaced
with Horn's closed form solution (a change that only takes adding three lines in a YAML configuration file). 
The dataset on which the study has been done is KITTI, with results
already shown in Table~\ref{tab:kitti.metrics} in the rows ``MOLA-LO (default)'' and ``MOLA-LO (Horn's)''. 
In all sequences, excepting \texttt{01} and \texttt{04}, 
the Horn's method leads to worse results. Overall, the 
average RTE worsens from 0.55\% to 0.65\%.
The reason for these results is the impossibility for 
this method to cope with outliers, something easy to integrate
in non-linear optimizers. Another interesting observation
is that, despite the solver itself runs $\sim 4$ times faster than Gauss-Newton,
the overall time cost is $\sim 2$ times slower due to the need to spend more ICP iterations until convergence.}

\subsection{\REVIEW{Loop-closure with and without GNSS}}
\label{sect:ablation.lc.gnss}

\REVIEW{Finally, the capability of the proposed 
loop closure algorithm to incorporate optional GNSS readings
is put at test with the Mulran dataset.
Results are shown in the two last rows of Table~\ref{tab:mulran.metrics}.
Overall, the conclusion here is that the presence of GNSS: 
(i) improves the global accuracy of the trajectories 
(reduced ATE values in the version with GNSS in 11 out of 12 sequences), 
and (ii) enables identifying potential loop closures 
that would not be found by our metric uncertainty-based 
hypothesis generator in the very long loops ($>20~km$) 
of Sejong sequences.
}



\section{Conclusions}
\label{sect:conclusions}
This paper introduced a whole framework that aims at filling a gap in
the robotics community's needs for flexible map building and editing from
3D LiDAR.
It has been shown how the proposed LO and SLAM systems compare well or outperform
other SOTA methods regarding estimated trajectory accuracy and robustness against
divergence.
\REVIEW{Excepting georeferencing and loop-closure, all other components}
of the framework are available as open-source software.
The present work leaves plenty of research topics open for future works:
(i) benchmarking different metric map data structures to find out which ones suit
best to each problem (e.g. real-time LO vs. localization without mapping), 
(ii) 
designing new pipeline blocks for smarter sampling of point clouds
to achieve both, faster and more accurate LO, (iii) alternative pipeline designs
suitable for efficient depth camera odometry, \REVIEW{or (iv) adding support for tightly-integrated measurements from LiDAR intensity images or inertial sensors}.


\end{document}